\documentclass[manuscript,screen]{acmart}

\usepackage{tikz,tikz-qtree}
\usepackage{bbding} 

\newcommand{\treesec}[1]{\textcolor{black}{\textbf{#1}}}

\usepackage{graphicx}
\usepackage{xcolor}

\usepackage{multicol}

\usepackage{algorithmic}  

\usepackage[ruled,vlined,linesnumbered]{algorithm2e}
\usepackage{adjustbox}
\SetKwComment{Comment}{$\triangleright$\ }{}

\usepackage{nicefrac}

\usepackage[shortlabels]{enumitem}

\usepackage{hyperref}
\usepackage[nameinlink]{cleveref}
\crefname{ineq}{inequality}{inequalities}
\creflabelformat{ineq}{#2{\upshape(#1)}#3} 

\usepackage{longtable}

\usepackage{amsthm}

\usepackage{bbm}
\usepackage{dsfont}

\usepackage{multirow}

\usepackage{bbold}

\usepackage[super]{nth}

\theoremstyle{definition}
\newtheorem{definition}{Definition}
\theoremstyle{plain}

\theoremstyle{plain}
\theoremstyle{plain}

\usepackage{thmtools}
\usepackage{thm-restate}

\theoremstyle{plain}

\theoremstyle{plain}

\newcommand{\eg}[0]{\textit{e.g.}}
\newcommand{\etal}[0]{\textit{et al.}\xspace}

\newcommand{\ie}[0]{\textit{i.e.}}

\newcommand{\eps}[0]{\varepsilon}

\usepackage{accents}

\newcommand{\bb}[1]{\mathbb{#1}}

\newcommand{\cal}[1]{\mathcal{#1}}

\usepackage{xcolor}
\definecolor{macro_red}{HTML}{B31412}
\definecolor{macro_blue}{HTML}{185ABC}
\definecolor{macro_green}{HTML}{137333}
\newcommand{\envdesign}[1]{\textcolor{macro_red}{#1}}
\newcommand{\oladapt}[1]{\textcolor{macro_blue}{#1}}
\newcommand{\syslearn}[1]{\textcolor{macro_green}{#1}}

\usepackage{titlesec}
\titlespacing{\section}{0pt}{6pt}{6pt}
\titlespacing{\subsection}{0pt}{6pt}{6pt}

\usepackage{subcaption}

\usepackage{booktabs}
\usepackage{makecell}

\usepackage{wrapfig}

\makeatletter
\newcommand{\pushright}[1]{\ifmeasuring@#1\else\omit\hfill$\displaystyle#1$\ignorespaces\fi}
\makeatother

\usepackage{rotating}

\makeatletter
\let\save@mathaccent\mathaccent
\newcommand*\if@single[3]{%
  \setbox0\hbox{${\mathaccent"0362{#1}}^H$}%
  \setbox2\hbox{${\mathaccent"0362{\kern0pt#1}}^H$}%
  \ifdim\ht0=\ht2 #3\else #2\fi
  }
\newcommand*\rel@kern[1]{\kern#1\dimexpr\macc@kerna}
\newcommand*\widebar[1]{\@ifnextchar^{{\wide@bar{#1}{0}}}{\wide@bar{#1}{1}}}
\newcommand*\wide@bar[2]{\if@single{#1}{\wide@bar@{#1}{#2}{1}}{\wide@bar@{#1}{#2}{2}}}
\newcommand*\wide@bar@[3]{%
  \begingroup
  \def\mathaccent##1##2{%
    \let\mathaccent\save@mathaccent
    \if#32 \let\macc@nucleus\first@char \fi
    \setbox\z@\hbox{$\macc@style{\macc@nucleus}_{}$}%
    \setbox\tw@\hbox{$\macc@style{\macc@nucleus}{}_{}$}%
    \dimen@\wd\tw@
    \advance\dimen@-\wd\z@
    \divide\dimen@ 3
    \@tempdima\wd\tw@
    \advance\@tempdima-\scriptspace
    \divide\@tempdima 10
    \advance\dimen@-\@tempdima
    \ifdim\dimen@>\z@ \dimen@0pt\fi
    \rel@kern{0.6}\kern-\dimen@
    \if#31
      \overline{\rel@kern{-0.6}\kern\dimen@\macc@nucleus\rel@kern{0.4}\kern\dimen@}%
      \advance\dimen@0.4\dimexpr\macc@kerna
      \let\final@kern#2%
      \ifdim\dimen@<\z@ \let\final@kern1\fi
      \if\final@kern1 \kern-\dimen@\fi
    \else
      \overline{\rel@kern{-0.6}\kern\dimen@#1}%
    \fi
  }%
  \macc@depth\@ne
  \let\math@bgroup\@empty \let\math@egroup\macc@set@skewchar
  \mathsurround\z@ \frozen@everymath{\mathgroup\macc@group\relax}%
  \macc@set@skewchar\relax
  \let\mathaccentV\macc@nested@a
  \if#31
    \macc@nested@a\relax111{#1}%
  \else
    \def\gobble@till@marker##1\endmarker{}%
    \futurelet\first@char\gobble@till@marker#1\endmarker
    \ifcat\noexpand\first@char A\else
      \def\first@char{}%
    \fi
    \macc@nested@a\relax111{\first@char}%
  \fi
  \endgroup
}
\makeatother

\usepackage{amsmath,amsfonts,bm}

\newcommand{\Scal}{\mathcal{S}}

\newcommand{\Acal}{\mathcal{A}}

\def\eqref#1{equation~\ref{#1}}

\def\1{\bm{1}}

\def\eps{{\epsilon}}

\DeclareMathAlphabet{\mathsfit}{\encodingdefault}{\sfdefault}{m}{sl}
\SetMathAlphabet{\mathsfit}{bold}{\encodingdefault}{\sfdefault}{bx}{n}

\usepackage{wrapfig}
\AtBeginDocument{%
  \providecommand\BibTeX{{%
    \normalfont B\kern-0.5em{\scshape i\kern-0.25em b}\kern-0.8em\TeX}}}
\setlist[itemize]{noitemsep, topsep=0pt}

\begin{document}

\title{Trustworthy Reinforcement Learning Against Intrinsic Vulnerabilities: Robustness, Safety, and Generalizability}




\author{Mengdi Xu}
\authornote{Authors contributed equally to this research.}
\email{mengdixu@andrew.cmu.edu}
\orcid{0000-0001-9332-4175}
\author{Zuxin Liu}
\authornotemark[1]
\email{zuxinl@andrew.cmu.edu}
\orcid{0000-0001-7412-5074}
\author{Peide Huang}
\authornotemark[1]
\email{peideh@andrew.cmu.edu}
\orcid{0000-0002-1048-8690}
\affiliation{%
  \institution{Carnegie Mellon University}
  \city{Pittsburgh}
  \state{Pennsylvania}
  \country{USA}
  \postcode{15213}
}




\author{Wenhao Ding}
\email{wenhaod@andrew.cmu.edu}
\orcid{0000-0003-3218-8792}
\affiliation{%
  \institution{Carnegie Mellon University}
  \city{Pittsburgh}
  \state{Pennsylvania}
  \country{USA}
  \postcode{15213}
}
 
\author{Zhepeng Cen}
\email{zcen@andrew.cmu.edu}
\orcid{0000-0001-7127-244X}
\affiliation{%
  \institution{Carnegie Mellon University}
  \city{Pittsburgh}
  \state{Pennsylvania}
  \country{USA}
  \postcode{15213}
}

\author{Bo Li}
\orcid{0000-0003-4883-7267}
\affiliation{%
  \institution{University of Illinois Urbana-Champaign}
  \city{Champaign}
  \state{Illinois}
  \country{USA}}
  \postcode{61801-3633}
\email{lbo@illinois.edu}

\author{Ding Zhao}
\orcid{0000-0002-9400-8446}
\affiliation{%
  \institution{Carnegie Mellon University}
  \city{Pittsburgh}
  \state{Pennsylvania}
  \country{USA}
  \postcode{15213}
}
\email{dingzhao@cmu.edu}





\renewcommand{\shortauthors}{}


\begin{abstract}
A trustworthy reinforcement learning algorithm should be competent in solving challenging real-world problems, including {robustly} handling uncertainties, satisfying {safety} constraints to avoid catastrophic failures, and {generalizing} to unseen scenarios during deployments. This study aims to overview these main perspectives of trustworthy reinforcement learning considering its intrinsic vulnerabilities on robustness, safety, and generalizability. In particular, we give rigorous formulations, categorize corresponding methodologies, and discuss benchmarks for each perspective. Moreover, we provide an outlook section to spur promising future directions with a brief discussion on extrinsic vulnerabilities considering human feedback. We hope this survey could bring together separate threads of studies together in a unified framework and promote the trustworthiness of reinforcement learning.

\end{abstract}




\maketitle


\section{Introduction}


With its vast potential to tackle some of the world’s most pressing problems, reinforcement learning (RL)~\cite{DBLP:books/lib/SuttonB98} is applied to transportation~\cite{9146378}, 
manufacturing~\cite{doi:10.1080/00207543.2021.1973138}, security\cite{ilahi2020challenges},  healthcare~\cite{10.1145/3477600}, and world hunger~\cite{9086620}.
As RL has started to shift towards deployment on real-world problems, its rapid development is coupled with as much risk as reward~\cite{ai2019high, dulac2021challenges,national2019national}. Before consumers embrace RL-empowered services, researchers are tasked with proving the trustworthiness of their innovations.


Trustworthiness is conceived to maximize the benefits of AI systems while at the same time minimizing their risks~\cite{ai2019high}.
It has rich meanings that go beyond its literal sense, and motivates a comprehensive framework that includes multiple principles, requirements, and criteria~\cite{ai2019high}.
Recently, there has been exciting progress in the area of \textit{trustworthy RL}~\cite{moos2022robust, pinto2017robust,nilim2005robust, xu2010distributionally, sinha2017certifying, achiam2017constrained, ray2019benchmarking, stooke2020responsive, liu2022constrained, liu2022robustness, finn2017model, akkaya2019solving, rajeswaran2016epopt, peng2018sim}, which greatly help to advance our understanding of intrinsic vulnerabilities in RL and potential solutions in particular aspects of trustworthy RL. 
It is clear that the next leap toward trustworthy RL will require a holistic and fundamental understanding of the challenges of these problems, the weakness, and advantages of existing trustworthy RL approaches, and a paradigm shift of trustworthy RL based on existing work. 
Compared with the trustworthiness issues in traditional machine learning (ML), the problems in RL are orders of magnitude more complicated, given that RL is a multi-faceted system containing several Markov Decision Processes (MDP) components (observations, environmental dynamics, actions, and rewards)~\cite{DBLP:books/lib/SuttonB98}.
Notably, these components may be subject to diverse robustness, safety, generalization, and security concerns, which have or have not been considered in traditional ML.
The concerns in the more static parts (\eg, observations) can find their root in ML studies, while those more relevant to the interaction property of the system (\eg, actions) are unique to RL and have been less explored. 
In addition, when viewing RL as two stages (model training followed by model deployment), we can study the vulnerabilities of the two stages alone and the connections between them.

\begin{figure}[t]
            \centering
            \includegraphics[width=1.0\textwidth]{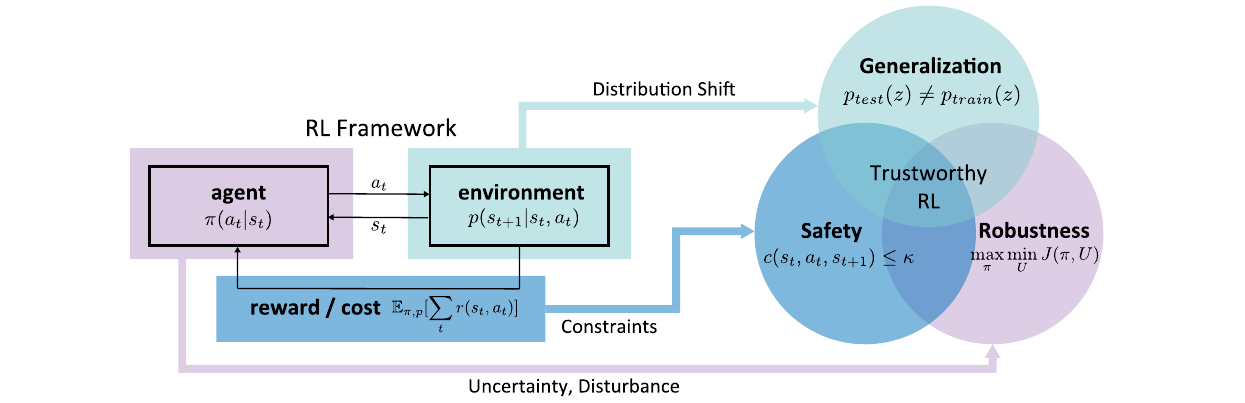}
            \caption{Overview diagram for trustworthy RL against intrinsic vulnerabilities: robustness, safety, generalization }
            \label{fig:overview}
\end{figure}

To promote the advancement of the field, in this paper, we provide a holistic survey of trustworthy RL from three aspects regarding the four MDP elements and two stages including
\begin{itemize}
    \item \textbf{robustness} against perturbations and uncertainties,
    \item \textbf{safety} of RL to constrain devastating costs, and
    \item \textbf{generalization} to in-domain and out-of-domain unseen environments.
\end{itemize}

These requirements are usually simultaneously required for an intelligent agent with strong interconnections. 
Taking autonomous vehicles (AVs) as an example,
we concretely explain the three aspects of trustworthiness. 
First, the observations of the AVs are susceptible to the adversaries perturbing the camera or LiDAR input, the environmental dynamics and rewards may be poisoned when training the AVs, and the actuators can also be manipulated by the adversaries when executing a deployed benign policy.
Regarding the two stages and their connections, during training, safe exploration is expected so that the car would not crash; during deployment, safety constraints are also expected to be adhered to avoid dangerous outcomes. In addition, good generalization is required from the training environment of the AVs to the testing environment. An AV should generalize between cities and different weathers and seasons.

Despite these interconnecting aspects of trustworthiness, existing surveys mainly focus on limited parts. 
García \etal~\cite{garcia2015comprehensive}, Gu \etal~\cite{gu2022review}, and Brunke \etal~\cite{brunke2022safe} work on safe RL, while Moos \etal~\cite{moos2022robust} and Kirk \etal ~\cite{kirk2021survey} focus on robustness and generalization, respectively. They advance the field by providing concrete descriptions for one aspect of trustworthy RL, yet lack a comprehensive characterization of the intrinsic vulnerabilities of RL as we envisioned above. In this survey, we would like to provide a unified framework for the three aspects of trustworthy RL.
For each aspect, we will
1) clarify the terminologies, 2) analyze their intrinsic vulnerabilities, 3) introduce methods that address these vulnerabilities, and 4) summarize popular benchmarks.
As shown in Fig.~\ref{fig:overview}, robustness, safety, and generalizability have strong correlations to agent, environment, reward, and cost, which correspond to the components of the MDP. To be within a scope that is easy for readers,
, we constrain our survey in the MDP setting. We view our survey as the \textit{intrinsic} trustworthy aspects of RL as we assume the human preference and setting of the agent and environment are pre-defined. In the outlook section, we will discuss the links between intrinsic and \textit{extrinsic} trustworthiness. Particularly, we will provide an outlook on four fundamental questions.

\begin{itemize}
    \item How to certify and evaluate for trustworthy reinforcement learning?
    \item What is the relation between the different aspects of trustworthiness?
    \item How to co-design trustworthy RL with the physical agent and environment?
    \item How to achieve the human-centric design for trustworthy RL?
\end{itemize}
We hope this survey will  bring together disparate threads of studies together in a unified framework and spur new research for a holistic view of the intrinsic aspects of trustworthy RL.



The remainder of the survey is organized as follows.
From \Cref{sec:robust} to~\Cref{sec:generalization}, each section addresses one aspect of trustworthiness.
We will explain our outlook on the three aspects of the key challenges in future directions beyond the survey in~\Cref{sec:challenges}. We will conclude the survey with 10 takeaways in ~\Cref{sec:conclusion}.

\section{Robust RL against perturbation}
\label{sec:robust}

\subsection{Overview}
Robust RL aims to improve the worst-case performance of algorithms deterministically or statistically in the face of uncertainties and adversarial attacks. 
The discrepancies between training tasks and testing tasks widely exist. For instance, in continuous control tasks, the real physical parameters may be different from those in simulation, and in autonomous driving scenarios, the surrounding agents may have novel driving behaviors. Such discrepancies motivate the development of robust RL. Furthermore, the safety-critical nature of real-world applications makes robustness an even vital feature to help avoid catastrophic failures. There is a surge of interest in researching effective attacks and defenses in RL. Detailed reviews of both fields can be found in ~\cite{chen2019adversarial, ilahi2021challenges, moos2022robust}.

\begin{figure}
            \centering
            \resizebox{1.0\textwidth}{!}{
\usetikzlibrary{trees}
\begin{tikzpicture}[
    i/.style={
        anchor=north,
        align=center,
        top color=white,
        bottom color=blue!10,
        rectangle,rounded corners,
        minimum height=6mm,
        draw=blue!75,
        align=center,
        color=blue!75,
        text=black,
        text depth = 0pt
    }
]
    \tikzset{execute at begin node=\strut}
    \tikzset{font=\small,
        grow=down,
        level distance=1.5cm,
        every tree node/.style={align=center,anchor=north},
        every leaf node/.style=
            {
                anchor=north,
                align=center,
                fill=blue!5,
                rectangle,
                draw=blue!50,
                align=center,
                distance=1cm,
                text depth = -10pt
            },
        edge from parent/.style=
            {
                draw=blue!50,
                thick,
                edge from parent path={(\tikzparentnode.south)
                -- +(0,-8pt)
                -| (\tikzchildnode)}
            }
    }

    \Tree [.\node[i] {Robust RL};
    [
        .\node[i] {Train with Adversary \\
        \treesec{(\S \ref*{sec:robust_train_adv})} };
        [
            .\node[i]{State \\Observations};
            {
                \cite{behzadan2017whatever}
                \cite{kos2017delving} 
                \cite{pattanaik2018robust}\\ 
                \cite{gleave2019adversarial}
                \cite{zhang2021atla}
                \cite{sun2022who}
            }           
        ]
        [
            .\node[i]{Actions};
            {
                \cite{tessler2019action}
                \cite{klima2019robust}
            }
        ]
        [
            .\node[i] {Transitions \\and Rewards};
            {
                \cite{pinto2017robust}
                \cite{mandlekar2017adversarially}\\
                \cite{huang2022robust}
            }
        ]
    ]
    [
        .\node[i] {Relaxation Methods \\
        \treesec{(\S \ref*{sec:robust_train_relax})} };
        [
            .\node[i]{State \\Observations};
            {
                \cite{oikarinen2021robust}
                \cite{zhang2021robust}\\
                \cite{wu2021crop}
                \cite{wu2022copa}
            }
        ]
        [
            .\node[i] {Transitions \\and Rewards};
            {
                \cite{sinha2017certifying}
                \cite{lecarpentier2019non} \\
                \cite{derman2020distributional, yang2017convex}
            }
        ]
    ]
    [
        .\node[i]{Regularization \\ 
        \treesec{(\S \ref*{sec:robust_train_reg})}
        }; 
        [
            .\node[i]{Agent's or \\ Adversary's Policies};
            {
                \cite{zhang2021robust}
                \cite{shen2020deep}\\
                \cite{smirnova2019distributionally}
            }        
        ]
    ]
    [
        .\node[i]{Constraint Opt.\\
        \treesec{(\S \ref*{sec:robust_train_constraint})}
        };
        [
            .\node[i]{Transitions \\and Rewards};
            {
                \cite{abdullah2019wasserstein}
                \cite{sinha2017certifying}
            }
        ]
    ]
    [
        .\node[i]{Model Estimation\\ 
        \treesec{(\S \ref*{sec:robust_train_other})}};
        [
            .\node[i]{Transitions \\ and Rewards};
            {
                \cite{rajeswaran2016epopt} 
                \cite{wang2020reinforcement}\\
                \cite{romoff2018reward}
                \cite{fu2017learning} 
            }
        ]
        [
            .\node[i]{Multi-agent \\Scenarios};
            {
                \cite{gallego2019reinforcement}
            }
        ]
    ]
    [
        .\node[i]{Robust Testing \\ Methods \treesec{(\S \ref*{sec:robust_test})}};
        [
            .\node[i]{Defensive \\ Actions};
            {
                \cite{everett2021certifiable}
                \cite{gowal2018effectiveness}\\
                \cite{weng2018towards}
                \cite{sinha2020formulazero}
            }
        ]
        [
            .\node[i]{Detect \\ Attacks};
            {
                \cite{lin2017detecting}
                \cite{havens2018online}
            }
        ]
    ]
    ]
\end{tikzpicture}

}
            \caption{Categories of robust RL methods }
            \label{fig:tree-robust-rl-new}
            \vspace{-0.1in}
\end{figure}
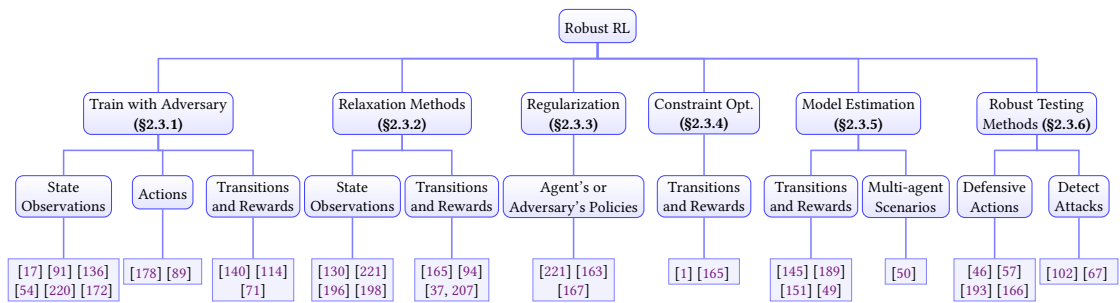

In this section, we first summarize robust RL formulations in Sec.~\ref{sec:robust-formulation}, in terms of robustness against uncertainties in different MDP components, including observations/states, actions, transitions, and rewards.
We then present robust training and testing methods in Sec.~\ref{sec:robust-methods} to improve the robustness against uncertainties or artificially designed attacks on each component. We summarize the robust RL methodologies in Fig.~\ref{fig:tree-robust-rl-new}. 
Finally, we present the applications and benchmarks for testing the robustness of proposed robust RL algorithms in Sec~\ref{sec:robust_benchmarks}.


\subsection{Problem formulation of robust RL}
\label{sec:robust-formulation}
We will build robust RL on the MDP setting as shown in Fig.~\ref{fig:overview}.
A MDP is represented by a tuple $M=(\mathcal S, \mathcal A, R, P, \rho_0, \gamma)$. Here, $\mathcal S$ is the state space, $\mathcal A$ is the action space, $R: \mathcal S \times \mathcal A \mapsto \mathbb{R}$ is the reward function, $P: \mathcal S \times \mathcal A \mapsto \Delta(\mathcal{S})$ is the transition function with $\Delta(\cdot)$ defining the probability simplex, $\rho_0 \in \Delta(\cal S)$ is the distribution over the initial state, and $\gamma \in (0, 1]$ is the discount factor.
At time step $t$, the agent observes its state $s_t \in \cal S$, takes action $a_t \in \cal A$, transitions to the next state $s_{t+1}\sim P(s_t,a_t)$, and receives reward $r_t = R(s_t,a_t)$. 
The goal is to learn a policy $\pi:\cal S\rightarrow \Delta(\cal A)$ that maximizes the expected cumulative reward $\mathbb{E}[\sum\nolimits_t \gamma^t r_t]$ with $s_0\sim \rho_0$ and $a_t\sim \pi(s_t)$.

Moreover, in robust RL settings, there exists an uncertain variable $U$ that lies within an uncertainty set $U \in \mathcal{F}_U$. The uncertain variable could be the state $s \in \mathcal{S}$, action $a \in \cal A$, reward function $R$ or transition $P$. Formally, $U \in \{ s, a, R, P \}$.
Robust RL algorithms aim to find a policy that maximizes the worst-possible performance against the uncertain variable $U$. Formally, they solve the following maximin optimization problem:
\begin{align}
    \max_{\pi} \min_{U \in \mathcal{F}_U}  J_{M}(\pi, U),
\end{align}
where $J_{M}(\pi, U)$ is the expected accumulated return as a function of the policy $\pi$, the MDP $M$, and the uncertain variable $U$. 
In this section, we discuss naturalistic uncertainties that arise in realistic applications and artificially designed adversarial attacks to different choices of $U$, including state observations (Sec.~\ref{sec:robust-ob}), actions (Sec~\ref{sec:robust-act}), transitions, and rewards (Sec~\ref{sec:robust-env}). 
A lot of robust RL methods motivated by naturalistic uncertainties adopt the adversarial training framework and thus assume an adversary agent conducting adversarial attacks to mimic the naturalistic uncertainties.

\vspace{-0.1in}

\subsubsection{Robust RL against uncertain state observations}
\label{sec:robust-ob}~\\
A majority of Robust RL methods aim to robustly handle the mismatches between observed states and actual states.
The state uncertainty may come from partial state observation. 
due to sensor errors or limited sensor capacities.
In general, the mismatch between the actual state and the state observation decreases the policy's performance or may even cause catastrophic failures in safety-critical situations. Robust RL against state observation uncertainties aims to robustly handle such state observation mismatch.
Let the observed states be $\tilde{s} \in \mathcal S$ which serves as the input for action generations $a \sim \pi(\tilde{s})$. Let the actual states be $s \in \mathcal S$ for environment rollouts $s' \sim P(s, a)$. Robust RL aims to find the optimal policy $\pi$ that achieves the best performance under the worst-possible state observations~\cite{zhang2021robust}.
\begin{align}
    \max_{\pi} \min_{ \tilde{s} } \mathbb{E}_{a_t \sim \pi(\tilde{s_t}), P, r } \left[  \sum_{t=1}^{T} \gamma^{t-1} r(s_t, a_t) \right].
\end{align}

The attacker adds perturbations to the state or observations to achieve certain \textit{adversarial goals}.
Let the state observation be $\tilde{s} = s+ \eps$ where $\eps$ represents the perturbation, the target adversarial state be $s^{\rm adv}$, and the adversarial policy be $\pi^{\rm adv}$; the normal reward signal be $r$ and the adversarial reward signal be $r^{\rm adv}$.
For \textit{training time attacks}, the attacker may minimize the cumulative reward (\ie, $\min \bb E\left[\sum \gamma^t r_t\right]$) or maximize an adversarial cumulative reward (\ie, $\max \bb E\left[\sum \gamma^t r^{\rm adv}_t\right]$).
For \textit{testing time attacks}, the attacker corrupts the agent performance by (1) luring the agent to the target adversarial state $s^{\rm adv}$, (2) misleading the action selection, \ie, $\pi(\tilde s)\neq \pi(s)$ or luring the agent to learn the target adversarial policy $\pi^{\rm adv}$. 
Most works on attacking the observation space of the agent assume the threat model of $\ell_p$-bounded attacks, extending the study on adversarial examples in the image domain~\cite{szegedy2013intriguing,goodfellow2014explaining}. This threat model holds in all the following descriptions unless otherwise stated.

Attacks on state observations at \textit{training time} are widely studied in multiple settings. 
ATLA~\cite{zhang2021atla} and PA-AD~\cite{sun2022who} consider an optimal adversary under the SA-MDP framework~\cite{zhang2021robust}, which aims to lead to minimal value functions under bounded state perturbations.
To find this optimal adversary (\ie, the optimal adversarial state perturbation), 
ATLA~\cite{zhang2021atla} proposes to train an adversary whose action space is the perturbation set in the state space,
while PA-AD~\cite{sun2022who} further decouples the problem of finding state perturbations into finding policy perturbations plus finding the state that achieves the lowest value policy, thus addressing the challenge of large state space.
In the \textit{multi-agent environment} specifically, \citet{gleave2019adversarial} consider a different threat model where the adversary can control the agent's opponent.
They propose to craft natural adversarial observations via training the opponent's adversarial policy w.r.t. the agent's fixed trained policy, thus achieving the adversarial goal empirically.
Different from these works which minimize the cumulative reward in the original task, \citet{tretschk2018sequential} try to maximize an adversarial cumulative reward guided by an adversarial reward signal $r^{\rm adv}$. 
Assuming whitebox access to the trained agent, they train an Adversarial Transformer Network (ATN)~\cite{baluja2018learning} along with the trained agent on the adversarial task to generate the perturbations.

At \textit{testing time}, to lure the agent to targeted corrupted action sequences, \citet{lin2017tactics} first generate a preferred action sequence for luring the agent via a planning algorithm, and then craft the adversarial state perturbations that lead to the planned actions. The action sequence is dynamically adjusted after each step to increase the attack's effectiveness.
In addition, non-targeted methods are effective to different degrees by applying \textit{random noise} to simply interfere with the action selection~\cite{kos2017delving}, or adopting \textit{adversarial attacks} (\eg, FGSM~\cite{goodfellow2014explaining} and CW attacks~\cite{carlini2017towards}) to deliberately alter the probability of action selection (either reduce the probability of selecting the optimal action~\cite{huang2017adversarial,kos2017delving,behzadan2017whatever} or increase the probability of selecting the worst action~\cite{lin2017tactics,pattanaik2018robust}).
\citet{behzadan2017vulnerability} propose the policy induction attack, where a surrogate Q-network is trained in tandem with the target agent Q-network, and the adversarial state observations are computed w.r.t. the surrogate model, and then leveraged to poison the target network, exploiting the transferability of adversarial examples. 




\subsubsection{Robust RL against uncertain actions}
\label{sec:robust-act}~\\
Robust RL against action uncertainties focuses on the discrepancy between the actions generated by the RL agent and the conducted actions. Such action uncertainty may come from the actuator noise, limited power range, or actuator failures in the real world.
Let the uncertain actions be $\tilde a = \nu(\pi(s))$, where $\nu$ is the action perturbation function. Robust RL against uncertain actions aims to find the optimal policy $\pi$ that achieves the best performance under the worst-possible actions. More concretely, the optimal policy is a solution to the following optimization problem:
\begin{align}
    \max_{\pi} \min_{ \tilde{a} } \mathbb{E}_{\tilde{a}_t \sim \nu(\pi(s_t)), P, r } \left[  \sum_{t=1}^{T} \gamma^{t-1} r(s_t, \tilde{a}_t) \right].
\end{align}


In the existing literature, the naturalistic uncertainties in action are represented by action attackers manipulating actions directly to minimize the accumulated return.
For instance, PR-MDP and NR-MDP~\cite{tessler2019action} assume that the perturbation function $\nu$ is related to an adversary's policy, and characterize the action uncertainty in the presence of action perturbations.

\subsubsection{Robust RL against uncertain transitions and rewards}
\label{sec:robust-env}~\\
The discrepancies between training and test environments widely exist due to sim-to-real gaps, the non-stationary nature of testing domains, training, and testing task distribution shifts, or partial observable environment transitions and rewards.
The environment discrepancy is generally reflected in different transitions and rewards.
Thus Robust RL against uncertain transitions and rewards aims to handle such environment mismatches robustly. 
In this case, the uncertain variable follows $U=(P, r)$ or $U=r$.

\textbf{Environment discrepancies.} We summarize two mainstream robust formulations related to environment discrepancies: (1) the \textit{robust MDP} formulation that assumes the transition and reward functions are uncertain, and (2) the \textit{distributionally robust MDP} formulation that assumes the latent distribution for generating the transitions and rewards is uncertain.
In comparison, the distributionally robust formulation encodes prior distributional information on possible transitions and rewards, and thus may help balance the performance and robustness. 
It is known that the distributionally robust MDP formulation help generate less conservative policies~\cite{xu2010distributionally, sinha2020formulazero} by balancing the nominal performances and the worst-case performances. The trade-offs between optimizing over nominal and worst-case distribution can be found in~\cite{xu2006robustness}.

\textit{\underline{Robust MDP.}} Robust RL methods and robust MDP formulations assume no distributional information on $(P, r)$. Robust MDP formulations directly put set constraints over environment transition probability $\tilde{p}: \mathcal{S} \times \mathcal{A} \rightarrow \mathcal{S} \subseteq \mathbb{R}^{d_{\mathcal{S}}}$ and reward distribution $\tilde{r}: \mathcal{S}\times \mathcal{A} \rightarrow \mathcal{R} \subseteq \mathbb{R}$ by assuming $({\tilde{p}, \tilde{r}})$ takes values in the uncertainty set $ \mathcal{F}$. It aims to derive a policy $\pi$ by solving the following max-min problem
\begin{align}
    \max_{\pi} \min_{({\tilde{P}, \tilde{r}}) \in \mathcal{F}} \mathbb{E}_{\pi,  ({\tilde{P}, \tilde{r}})} \left[  \sum_{t=1}^{T} \gamma^{t-1} \tilde{r}(s_t, a_t) \right]. \label{eq:dro_set_over_p_r}
\end{align}
$T$ is the planning horizon.
$\mathcal{F}$ here is a general ambiguity set without special structures. 

\textit{\underline{Distributionally robust MDP.}} In contrast, distributionally Robust MDPs (DR-MDP) assume additional distributional information on $(P, r)$. DR-MDP puts set constraints on $\mu$, where $(P, r) \sim \mu$. Formally, the transition probability $\tilde{p}$ and the reward $\tilde{r}$ are unknown parameters and obey a joint probability distribution $\mathbb{\mu}$ within an ambiguity set $\mathcal{F}$. Similar to the robust MDP settings, let $T$ be the planning horizon. More concretely, the DR-MDP objective is 
\begin{align}
    \max_{\pi} \min_{\mu \in \mathcal{F_S}} \mathbb{E}_{\mu} \left[ \mathbb{E}_{\pi, ({\tilde{p}, \tilde{r}}) \sim \mathbb{\mu}} \sum_{t=1}^{T} \gamma^{t-1} \tilde{r}(s_t, a_t) \right]. \label{eq:dro_set_over_mu}
\end{align}

It is known that solving the maximin problem in Eq.~(\ref{eq:dro_set_over_p_r}) or Eq.~(\ref{eq:dro_set_over_mu}) with general ambiguity set $\mathcal{F}$ is NP-hard~\cite{wiesemann2013robust}. Hence, several choices of the ambiguity set that facilitate solving robust MDPs are proposed. 
$\mathcal{F}_{s,a}$ is a state-action-wise independent ambiguity set following "$(s,a)$"-rectangularity~\cite{nilim2005robust}, where $\mathcal{F} = \bigotimes_{(s,a) \in \cal S \times \cal A}F_{s,a}$.
$\mathcal{F}_s$ is a state-wise independent ambiguity set~\cite{wiesemann2013robust}, where $\mathcal{F} = \bigotimes_{(s) \in \cal S }F_{s}$.
The k-rectangular ambiguity set proposed in~\cite{mannor2016robust} is a set such that its projection onto $\cal S' \subset \cal S$ is one of at most k different possible sets. 
The r-rectangular ambiguity set proposed in~\cite{goyal2022robust} assumes the transition is a linear function of a factor matrix within a factor matrix uncertainty set. 
Both k-rectangular and r-rectangular ambiguity sets are coupled sets that model the correlation of transitions between different states.
$\mathcal{F}_{s,a}$ assumes that the transition probability corresponding to each $(s,a)$ pair are unrelated to any other state-action pairs. One can use robust value iteration to get a stationary and deterministic optimal policy with $\mathcal{F}_{s,a}$.
$\mathcal{F}_{s}$ generalizes $\mathcal{F}_{s,a}$ by assuming that the transitions corresponding to each state are independent of other states. The optimal robust policy under $\mathcal{F}_{s}$ could be stationary but not necessarily deterministic.
The k-rectangular ambiguity set~\cite{mannor2016robust} is proposed to address the conservative issue of $\mathcal{F}_{s,a}$ and $\mathcal{F}_{s,a}$ when modeling uncertainty by modeling the correlations between transitions corresponding to different states. 


In addition to the rectangularity assumption, the discrepancy measure is another vital setting to define the ambiguity set $\mathcal{F}$.
Possible choices include Wasserstein distance~\cite{abdullah2019wasserstein, sinha2017certifying, lecarpentier2019non, derman2020distributional}, KL divergence~\cite{smirnova2019distributionally, mankowitz2019robust} or moments~\cite{xu2010distributionally, yu2015distributionally, chen2019distributionally, yang2017convex}. A comprehensive review of ambiguity sets is in~\cite{rahimian2019distributionally}. Choosing a proper structure of the ambiguity set and dynamically updating the ambiguity set with streaming data are still open problems.


Beyond the MDP formulations, the robust and distributionally robust setting in~\cite{nakao2021distributionally} can be further extended to settings with partial observations, such as robust POMDP formulations in~\cite{osogami2015robust, rasouli2018robust, saghafian2018ambiguous} and distributionally robust POMDP formulation in~\cite{nakao2021distributionally}.

\textbf{Adversarial attacks on rewards.}\quad
Adversarial attacks on rewards distort the rewards at each time step to lure the agent into learning a bad policy. Let the corrupted rewards be $\tilde r = \nu(r)$, where $\nu$ manifests a reward Poisoning strategy~\cite{ma2019policy, huang2019deceptive, zhang2020adaptive} or equivalently a reward corruption function in the corrupted reward MDP formulation~\cite{everitt2017reinforcement, zhang2008value}. 
\citet{huang2019deceptive} analyze the impact of falsified reward (cost) signal on the convergence of Q-learning algorithm.
The authors propose a robust region regarding the falsified cost within which the adversary will consistently fail and provide conditions that the agent will learn the target adversarial policy.
Similarly, \citet{zhang2020adaptive} categorize the reward-poisoning attacks into three categories depending on the attack budget. A threshold has to be reached such that the attacks become feasible. Further increasing the threshold afterward gives rise to another two categories: non-adaptive attacks and adaptive attacks \cite{zhang2020adaptive}.

\subsection{Methodologies towards robust RL}
\label{sec:robust-methods}

In this section, we will summarize commonly-used training methods once the robust goals are defined in the previous section. Particularly, people would like to address the following issues in adversarial training: (1) How to set up the goal of the adversaries? (2) How to make the adversarial training computationally feasible with relaxation? (3) How to regulate the capabilities of adversaries such that the policy will not be over-conservative? We will introduce corresponding methods in the sections below.

\subsubsection{Robust training with adversaries}~\\
\label{sec:robust_train_adv}
\textit{Adversarial training}~\cite{goodfellow2014explaining,madry2018towards} has been recognized as one of the most effective approaches in traditional supervised learning tasks in \textit{training} time defenses. In this subsection, we will discuss one type of adversarial training strategy which is training RL agents with a lower bound of the loss (upper bound of the accumulated reward) obtained with local adversarial attacks.
The local attacks can be either unparameterized attacks or attacks from a parameterized adversary agent. Training with an adversary agent can naturally be formulated as a zero-sum game between the adversary agent and the RL agent~\cite{pinto2017robust}.




\textbf{State observations} Robust RL against state observations uncertainties directly adds perturbations to state observations or gradient updates of the RL agent. Such methods are based on a Q function to quantify the optimality of a state-action pair and search for the worst-possible perturbations.
In the Atari Pong environment with discrete actions and image-based observations, the adversarial training on noisy observations and FGSM attacks improves the resilience of the DQN~\cite{behzadan2017whatever} and A3C algorithm~\cite{kos2017delving}. 
Instead of storing perturbed images based on FGSM attacks, \citet{pattanaik2018robust} propose to store states perturbed by stronger gradient-based attacks, which helps improve DQN and DDPG's robustness over dynamics uncertainties. 
In contrast to the FSGM attacks in~\cite{kos2017delving} that decrease the probability of selecting the optimal actions given the current state, the gradient-based attacks in~\cite{pattanaik2018robust} encourage the probability of selecting the worst possible action. 
In addition to adding un-parameterized perturbations, it is possible to have a parameterized adversary agent that perturbs the state observations and minimizes the RL agent's cumulative reward.
The adversary agent could be either trained in parallel~\cite{zhang2021atla,sun2022who} with the RL agent or be a pre-trained adversary opponent~\cite{gleave2019adversarial}.

\textbf{Actions} Existing adversarial training strategies against action perturbations mainly focus on local adversarial attacks~\cite{tessler2019action, klima2019robust}. They adopt the game-theoretic formulation of robust RL, treating the RL agent as the protagonist agent trained against an adversary agent. For instance, \citet{tessler2019action} propose two criteria of robustness to action uncertainty: the Probabilistic Action Robust MDP (PR-MDP) criterion and the Noisy Action Robust MDP (NR-MDP) criterion which consider the occasional adversarial action, and the constant adversarial perturbations, respectively.
They formulate the robust learning problem by finding the optimal solution to the proposed MDP formulations and develop policy iteration methods for both criteria in the tabular setting.
In a deep RL setting, \citet{tessler2019action} proposes a robust deep deterministic policy gradient algorithm to train an adversary agent along with the protagonist agent. 
Similar to the PR-MDP setting, \citet{klima2019robust} introduce an adversary agent to shrink the action space of the protagonist agent in critical domains. They build on TD-learning with a modified TD error, which is a convex combination of Q values for the protagonist and the adversary agents. 




\textbf{Transitions and rewards} Robust adversarial RL (RARL)~\cite{pinto2017robust} is one of the most popular frameworks to handle environment uncertainties.  RARL models the environment discrepancy as extra forces/disturbances exerted by an adversary player.
It formulates the policy learning as a zero-sum, minimax objective, and proposes to alternatively optimize the adversary's and the protagonist's policy until convergence.
This can be viewed as optimizing the conditional value at risk (CVaR)~\cite{tamar2015optimizing} for the worst possible cases.
A contemporary work ARPL~\cite{mandlekar2017adversarially} leverages active computation of adversarial examples in terms of dynamics noise, process noise, or observation noise during training to enable training robust policies that transfer across domains.
The method gains robustness by learning from the naturally occurring adversarial scenarios that are generated on the fly, which constantly expose the flaws in the agent's decision. 
Recently, \citet{huang2022robust} proposes the RRL-stack formulation which treats robust RL as a general-sum Stackelberg game with the protagonist agent as the leader and the adversary agent as the follower. RRL-stack naturally encodes the sequential nature and provides a general formulation for robust training. With a proposed Stackelberg Policy Gradient algorithm, the RRL-stack formulation has better training stability compared with RARL in robotics and autonomous driving scenarios. 

\vspace{-0.1in}

\subsubsection{Robust training with relaxation}~\\
\label{sec:robust_train_relax}
In contrast to methods in Sec.~\ref{sec:robust_train_adv} that assume an adversary agent, another line of work trains RL agents with an upper bound of the loss (lower bound of the accumulated reward) which can be obtained with relaxation methods.

\textbf{State observations}
It is possible to derive a lower bound of the accumulated reward under adversarial perturbations and directly maximize the lower bound instead of the standard RL objective. 
For instance, \citet{oikarinen2021robust} constructs the loss with adversarial examples that lead to the maximized training loss.
\citet{zhang2021robust} propose State-Adversarial MDP where the perturbation is on state observations. Given a policy $\pi$, they derive the upper bound on performance drop between the value for a non-adversarial MDP and the value against the optimal adversary, leveraging convex relaxations of neural networks. 
The performance drop upper bound is in general equivalent to robustness guarantee and is closely related to the certifiable robust RL literature~\cite{wu2021crop, wu2022copa}.

\textbf{Transitions and rewards} It is possible to derive the policy performance gap under environment transition discrepancies by leveraging the nature of the ambiguity set.
\citet{lecarpentier2019non} define a novel non-stationary MDPs (NS-MDPs) which assumes the transition dynamics and reward functions are changing slowly over time. Mathematically, They bound the evolution rate with Lipschitz Continuity and use the Wasserstein distance as the metric
\citet{mankowitz2019robust} aim to improve the robustness of state-of-the-art continuous control RL algorithms and propose a robust version of maximum a-posteriori Policy Optimization (R-MPO). They utilize domain knowledge to select parameters to perturb and define the parameter-wise range of uncertainty set in training and testing.



With a Wasserstein-based ambiguity set and finite state action spaces, \citet{yang2017convex} constructs the lower bound of the inner value minimization problem using Kantorovich duality to solve the Bellman equations. 
The Wasserstein distributionally robust MDPs (DRWMDP) setting proposed in~\cite{derman2020distributional} bears some resemblances to~\cite{yang2017convex}. \citet{derman2020distributional} establish the connection between DRWMDP and regularizations and provides a regularized value function lower bound that guarantee robustness w.r.t. the Wasserstein ball in linear approximation case. 
In addition to Wasserstein-based ambiguity sets, the ambiguity set in~\cite{xu2010distributionally} is represented by a sequence of nested sets, and each set is associated with a confidence level.
\cite{yu2015distributionally} later extends~\cite{xu2010distributionally} to a more general uncertainty set formulation that can model both nested and disjoint sets.~\cite{chen2019distributionally} generalizes the results in~\cite{yu2015distributionally} with a new class of uncertainty set with the Wasserstein-based ambiguity set and the general-moment-based ambiguity set as two special cases.

\subsubsection{Robust training with regularization}~\\
\label{sec:robust_train_reg}
In Robust RL, a regularizer can be applied to the policy to enhance robustness. 
The regularizer can help encourage the smoothness of the learned policy, \eg, $\min_\theta\max_s\max_{\tilde s} D(\pi(\cdot | s), \pi(\cdot | \tilde s))$. Thus it prevents the drastic change of the policy under perturbations on state observations~\cite{zhang2021robust,shen2020deep}. 
Regularization-based methods~\cite{gu2014towards,hein2017formal,yan2018deep} have also been investigated in standard neural network training to improve the robustness of the trained models. 
Beyond policy smoothness, \citet{smirnova2019distributionally} propose to put a set constraint on the RL agent's policy to ensure safe exploration during the learning process.
They propose a distributionally robust policy iteration scheme and employ an adaptive KL-based uncertainty set with the uncertainty level related to the sample size. 
\citet{huang2022robust} propose to adaptively-regularize the adversary's policy using the highest possible return of the protagonist given the current environment generated by the adversary.
The regularized adversary generates challenging yet solvable environments to improve the robustness of the protagonist agents.

\subsubsection{Robust training with constrained optimization}~\\
\label{sec:robust_train_constraint}
With an explicitly defined ambiguity set, it is possible to solve the inner minimization problem in Eq.(~\ref{eq:dro_set_over_p_r}) using techniques in constrained optimization~\cite{sinha2017certifying,abdullah2019wasserstein}.
With a Wasserstein ambiguity set, \citet{sinha2017certifying} present a case study considering distributional robustness of deep Q-learning with Q value modeled by an overparameterized feed-forward neural net. With a Lagrangian penalty formulation, it can solve the worst perturbation over the transition dynamics with gradient descent when the penalty is sufficiently large.
\citet{abdullah2019wasserstein} propose Wasserstein Robust RL (WR$^2$L), which considers an RL setting that assumes
an environment adversary adding perturbations within an average Wasserstein ball $\mathcal{P}_0$. 
WR$^2$L conducts gradient ascents of the dynamics parameters until (almost) convergence. 
To guarantee that the updated dynamics parameters satisfy the ambiguity set constraint, WR$^2$L calculates the Wasserstein distance based on Monte-Carlo estimation which scales to high-dimensional non-convex or non-concave settings. 

\subsubsection{Robust training with model estimation}~\\
\label{sec:robust_train_other}
\textbf{Transitions.}
Ensemble learning methods aim to improve predictive performance by leveraging on a set of learning algorithms or models. Ensemble learning is combined with adversarial training in EPOpt~\cite{rajeswaran2016epopt}.
EPOpt iterates between two procedures: (1) modeling the source domain with an ensemble of models and finding a robust policy for the source domain, and (2) refining the ensemble of models given the data collected from the target domain.  

\textbf{Rewards.}\quad 
\citet{wang2020reinforcement} consider the setting of a finite reward set and model the reward corruption via a confusion matrix. Through dynamically refining the estimation of the confusion matrices with aggregated rewards, they manage to approximate the true reward signal.
\citet{romoff2018reward} propose to use an estimator for reward estimation in the scenario of corrupted stochastic reward. Basically, they learn the value functions and the reward functions simultaneously. When updating the value estimation, they leverage the estimated reward rather than the sampled reward, reducing the variance propagated to the value function.
AIRL~\cite{fu2017learning} is an adversarial learning based inverse adversarial learning algorithm that aims to learn the disentangled reward function that is invariant to changing dynamics by using a state-action level discriminator restricted to a reward approximator plus a shaping term.


\textbf{Multi-agent Scenarios.}\quad
\citet{gallego2019reinforcement} augment the MDPs with interfering adversaries and introduce the Threatened MDPs (TMDPs), explicitly modeling the beliefs that the agent has about the adversary's strategy.
The level-$k$ scheme~\cite{stahl1994experimental} is adopted such that the agent and adversary alternatively model each other's belief and thus become a player of a higher level.
A level-$2$ learner has been demonstrated to be significantly better than an independent learner (that does not consider the adversary) when playing against the adversary.





\vspace{-0.1in}

\subsubsection{Robust testing methods}~\\
\label{sec:robust_test}
With a pre-trained policy, robust testing methods aim to enhance the robustness of the agent during deployment. We identify two types of testing time robust methods, including (1) passively selecting defensive actions and (2) actively detecting potential attacks. 

\textbf{Selecting defense actions.}
At test-time, one natural way to improve robustness is selecting actions under the worst possible perturbations. Considering potential perturbations on state observations, CARRL~\cite{everett2021certifiable} proposes to select the action under the worst observation perturbations, by passing the worst observation to a trained DNN model. It computes a lower bound for Q-value based on the neural net verification methods which are related to model architecture and weights. Other methods derive lower bounds of the Q values under the worst-case perturbations based on neural network verification~\cite{gowal2018effectiveness,weng2018towards}.
Beyond the uncertainty in states, \citet{sinha2020formulazero} consider the uncertainty in the beliefs about the opponents' behaviors at test-time. They propose a distributionally robust bandit optimization method and dynamically adjust the risk aversion level. Therefore, the RL agent selects the optimal policy assuming the worst-possible beliefs within the ambiguity set.

\textbf{Detecting potential attacks}.
In contrast to the above methods which are specifically designed to achieve action sub-optimality, directly \textit{detecting} adversarial examples and predicting action based on the recovered state or reward information is rather generic.
This type of method first detects the adversarial attacks and/or predicts the benign states/reward functions, and then provides suggestions on actions based on the already trained model.
With \textit{state observation} uncertainties, the prediction can be based on a frame prediction model (visual foresight module~\cite{lin2017detecting}) or a master policy in the meta-learning hierarchy MLAH~\cite{havens2018online}.

\subsection{Application benchmarks and resources}\quad
\label{sec:robust_benchmarks}
Deep robust RL algorithms are widely tested in MuJoCo control tasks~\cite{todorov2012mujoco} with physical parameter discrepancies between training and testing tasks. For instance, the 2D Hopper benchmark~\cite{erez2012infinite}, and the HalfCheetah benchmark~\cite{wawrzynski2009real} vary the agent's mass, ground friction, joint damping, and armature according to Gaussian distributions. Both benchmarks are later adopted in~\cite{rajeswaran2016epopt}.
\citet{pinto2017robust} later propose to test in a wider range of tasks including InvertedPendulum, HalfCheetah, Swimmer, Hopper, Walker2d, and Ant with different mass and frictions.
In addition to the MuJoCo-based benchmark, there are also robust benchmarks based on the OpenAI Gym environment~\cite{brockman2016openai}. For instance, \citet{huang2022robust} propose to test in a LunarLander benchmark with different action delays.
Mankowitz et al.~\cite{mankowitz2019robust} present experiments on dexterous hand tasks which aim to rotate a cube into a target orientation using the Shadow hand domain.
Robust RL methods are also tested in Atari games with noise added to the image inputs such as~\cite{behzadan2017whatever, kos2017delving}. 

Beyond robotics continuous control tasks and simulated games, robust RL is also tested in mobile robot tasks and autonomous driving scenarios.
\citet{liu2022robustness} propose a safe and robust benchmark containing mobile robot tasks based on Bullet safety gym~\cite{gronauer2022bullet} environments.
\citet{jaafra2019robust} propose to test in CARLA simulator~\cite{dosovitskiy2017carla} with different conditions, including the traffic density, such as the number of dynamic objects, and visual effects such as weather and lightening conditions.
\citet{huang2022robust} propose a highway merge benchmark based on the highway-env environment 
\cite{highway-env} with different surrounding vehicles' initial positions.




\section{Safe learning to avoid devastating costs}
\label{sec:safe}

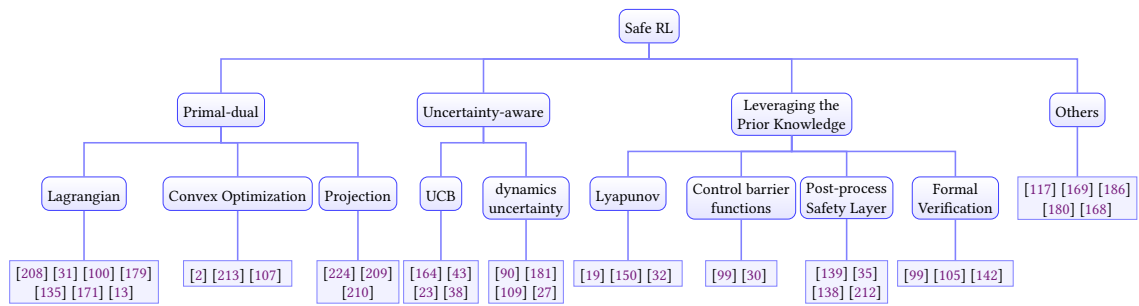
\begin{figure}
            \centering
            \resizebox{1.0\textwidth}{!}{
\usetikzlibrary{trees}
\begin{tikzpicture}[
    i/.style={
        anchor=north,
        align=center,
        top color=white,
        bottom color=blue!10,
        rectangle,rounded corners,
        minimum height=6mm,
        draw=blue!75,
        align=center,
        color=blue!75,
        text=black,
        text depth = 0pt
    }
]
    \tikzset{execute at begin node=\strut}
    \tikzset{font=\small,
        grow=down,
        level distance=1.5cm,
        every tree node/.style={align=center,anchor=north},
        every leaf node/.style=
            {
                anchor=north,
                align=center,
                fill=blue!5,
                rectangle,
                draw=blue!50,
                align=center,
                distance=1cm,
                text depth = -10pt
            },
        edge from parent/.style=
            {
                draw=blue!50,
                thick,
                edge from parent path={(\tikzparentnode.south)
                -- +(0,-8pt)
                -| (\tikzchildnode)}
            }
    }

    \Tree [.\node[i] {Safe RL};
    [
        .\node[i] {Primal-dual};
        [
            .\node[i]{Lagrangian};
            {
                \cite{yang2021wcsac}
                \cite{chow2017risk}
                \cite{liang2018accelerated}
                \cite{tessler2018reward}
                \\
                \cite{paternain2019constrained}
                \cite{stooke2020responsive}
                \cite{as2022constrained}
            }
        ]
        [
            .\node[i]{Convex Optimization};
            {
                \cite{achiam2017constrained}
                \cite{yu2019convergent}
                \cite{liu2022constrained}
            }
        ]
        [
            .\node[i] {Projection};
            {
                \cite{zhang2020first}
                \cite{yang2020projection}
                \\
                \cite{yang2021accelerating}
            }
        ]
    ]
    [
        .\node[i] {Uncertainty-aware};
        [
            .\node[i] {UCB};
            {
                \cite{singh2020learning}
                \cite{efroni2020exploration}
                \\
                \cite{brantley2020constrained}
                \cite{ding2021provably}
            }
        ]
        [
            .\node[i] {dynamics \\ uncertainty};
            {
                \cite{koller2018learning}
                \cite{thananjeyan2020safety}
                \\
                \cite{liu2020constrained}
                \cite{chen2021context}
            }
        ]
    ]
    [
        .\node[i] {Leveraging the\\Prior Knowledge};
        [
            .\node[i] {Lyapunov};
            {
                \citep{berkenkamp2017safe}
                \citep{richards2018lyapunov}
                \citep{chow2019lyapunov}
            }
        ]
        [
            .\node[i] {Control barrier \\ functions};
            {
                \citep{li2019temporal}
                \citep{cheng2019end}
            }
        ]
        [
            .\node[i] {Post-process\\Safety Layer};
            {
                \cite{pham2018optlayer}
                \cite{dalal2018safe}
                \\
                \cite{peng2022safe}
                \cite{yu2022towards}
            }
        ]
        [
            .\node[i] {Formal \\ Verification};
            {
                \cite{li2019temporal}
                \cite{liu2021recurrent}
                \cite{puranic2021learning}
            }
        ]
    ]
    [
        .\node[i] {Others};
            {
                \cite{mehta2020curriculum}
                \cite{srinivasan2020learning}
                \cite{turchetta2020safe}
               \\ \cite{thananjeyan2021recovery}
               \cite{sootla2022saute}
            }
    ]
    ]
\end{tikzpicture}}
            \caption{Categories of safe RL methods}
            \label{fig:tree-safe-rl}
            \vspace{-0.1in}
\end{figure}

\subsection{Overview}
Safety is also another major concern when deploying them to real-world safety-critical applications, such as self-driving cars. Traditional reinforcement learning only aims to maximize the task reward received from the environment, which lacks the guarantees of satisfying safety constraints. A safe policy should explicitly consider the safety constraints during training and prevent an RL agent from causing devastating costs or being in dangerous states. 
For instance, if an RL algorithm is deployed on a real robot arm, the safety constraints would be avoiding hitting fragile objects and surrounding people that may break valuable property or cause injury. Similarly, an RL agent in self-driving applications should obey traffic rules as well as avoid collisions with surrounding obstacles.
Therefore, it is important to develop safe reinforcement learning algorithms for real-world applications, which allow them to complete tasks while satisfying certain safety constraints.


Safe reinforcement learning, which is also named constrained reinforcement learning, aims to learn policies that maximize the expected task reward while satisfying safety constraints.
Depending on the safety requirements and the training requirements, safe RL has different formulations.
In terms of the safety requirement, there are \textit{trajectory-wise safety constraint} and \textit{state-wise safety constraint}.
From the training requirement perspective, we can categorize safe RL into two groups: \textit{safe deployment} and \textit{safe exploration}, where the former aims to act safely after the model training but may violate constraints during training, and the latter one refers to providing safety guarantees during training so that the whole training procedure could be finished without causing catastrophic unsafe behaviors.

Note in the setting of RL, agents need to learn from failures, similarly in the safe RL setting. One may argue that we should always avoid any safety-critical failures with domain knowledge rather than active learning. In this survey, however, we find in many cases, a precise description of the environment risk is unavailable. This is also the reason why we should pay attention to both safe deployment and safe exploration to manage the risks. Mathematically, the consequence is described as safety constraints. In the following section, we extend the traditional MDP to model them.



\subsection{Problem formulation of safe reinforcement learning}

Safe RL is usually modeled under the constrained Markov decision processes (CMDPs)~\citep{altman1998constrained} framework, where the agents are enforced with restrictions on auxiliary safety constraint violation costs.
A CMDP is defined by a tuple $(\Scal, \Acal, R, P, \gamma, \rho_0, C)$, where $\Scal$ is the state space, $\Acal$ is the action space, $R:\cal S\times \cal A \times \cal S\mapsto \bb R$ is the reward function, $P: \cal S\times \cal A \mapsto \Delta(\cal S)$ is the transition function,  $\gamma$ is the discount factor, and $\rho_0: \cal S \mapsto  \Delta(\cal S)$ is the distribution over the initial state. The first 6 elements are the same as the original MDP.
The last element $C = \{c^i: \cal S \times \cal A \times \cal S \mapsto \bb R_{\geq0}, i=1,..., m\}$ is unique to CMDPs, which is a set of costs for violating $m$ safety constraints. 
For simplicity, we consider only one constraint $c$ in the following definitions, though they could be extended to multiple constraints as well.


We denote $V_r^\pi(\rho_0) = \mathbb{E}_\pi[ \sum_t \gamma^tr(s_t, a_t, s_{t+1}) ]$ as the reward value function, which is the expected cumulative rewards under the policy $\pi \in \Pi : \cal S \mapsto \Delta(\cal A)$ and the initial state distribution $\rho_0$. Similarly, we denote $V_c^\pi(\rho_0)=\mathbb{E}_\pi[ \sum_t c(s_t, a_t, s_{t+1}) ]$ as the cost value function.
There are multiple formulations and definitions of the safe RL problem, depending on the level of safety requirements, though they share the same principle: maximizing the task performance while satisfying the constraint requirement.
We begin by introducing the most commonly used \textbf{trajectory-wise safety constraint}.

\textbf{Trajectory-wise safety constraint.} The constraint aims to ensure that the episodic cost under the policy $\Tilde{\pi}$ from the initial states to the terminal states is under the threshold $\kappa$.
\begin{equation}
    \pi^* = \arg\max_{\pi}  V_r^\pi(\mu_0), \quad s.t. V_c^{\Tilde{\pi}}(\mu_0) \leq \kappa
\end{equation}

where $\Tilde{\pi}$ depends on the safety requirement of different tasks. For instance, $\Tilde{\pi} = \pi^*$ for \textbf{safe deployment} applications, since we only require the optimized policy satisfying safety constraints. In contrast, for \textbf{safe exploration} applications, we require $\Tilde{\pi} = \{\pi_1, \pi_2,...,\pi_K\}$, where $\pi_k, k\in\{1,2,...,K\}$ is the policy for the $k$-th optimization iteration and $K$ is the total number of policy updates. 
Another similar formulation of safe RL studies the \textbf{state-wise safety constraint}, which aims to ensure that the cost of each state along the policy trajectory is under the threshold $\kappa$.

\begin{equation}
    \pi^* = \arg\max_{\pi}  V_r^\pi(\mu_0), \quad s.t. \forall (s_t, a_t, s_{t+1}) \sim \pi, s_0\sim \mu_0, c(s_t, a_t, s_{t+1}) \leq \kappa
\end{equation}

We could observe that the \textbf{safe exploration} formulation and the \textbf{state-wise} formulation have stricter safety requirements, since the constraint should be satisfied either all the time during the training or for every state. 
As a result, these formulations are more challenging and have less literature. 
Unless otherwise stated, we will by default assume the \textbf{safe deployment} formulation with \textbf{trajectory-wise} safety constraint in the following subsections.

\begin{wrapfigure}{R}{0.65\textwidth}
\vspace*{-0.03in}
\centering
\begin{subfigure}[h]{0.2\textwidth}
 \centering
 \includegraphics[width=\textwidth]{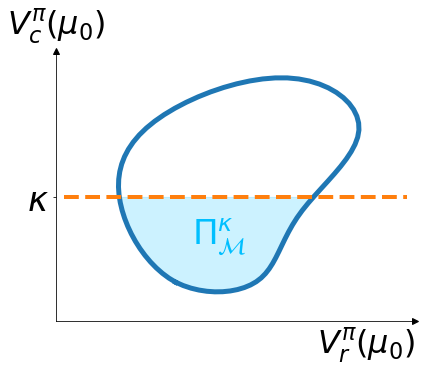}
 \caption{Feasibility}
 \label{fig:Feasibility}
\end{subfigure}
\hfill
\begin{subfigure}[h]{0.2\textwidth}
 \centering
 \includegraphics[width=\textwidth]{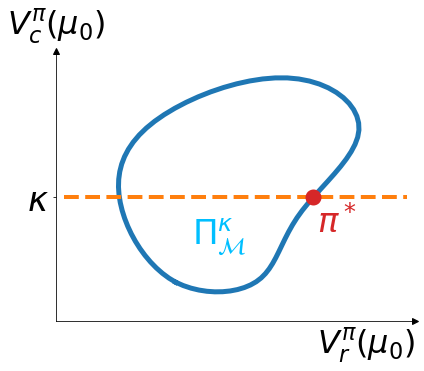}
 \caption{Optimality}
 \label{fig:Optimality}
\end{subfigure}
\hfill
\begin{subfigure}[h]{0.2\textwidth}
 \centering
 \includegraphics[width=\textwidth]{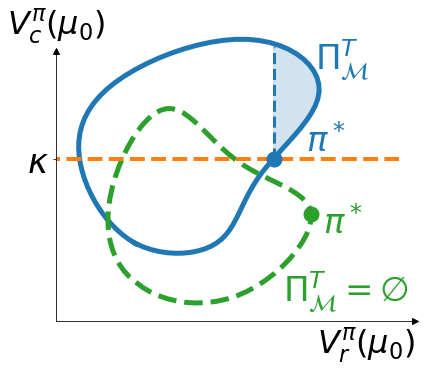}
 \caption{Temptation}
 \label{fig:Temptation}
\end{subfigure}
\vspace*{-2mm}
\caption{\small Illustration of definitions via a mapping from the policy space to the metric plane $\Pi \xrightarrow[]{} \mathbb{R}^2$, where the x-axis is the reward return and the y-axis is the cost return. A point on the metric plane denotes corresponding policies, i.e., the point $(v_r, v_c)$ represents the policies $\{\pi \in \Pi|V_r^{\pi}(\mu_0)=v_r, V_c^{\pi}(\mu_0)=v_c\}$. The blue and green circles denote the policy space of two safe RL problems.}
\label{fig:definition}
\vspace{-1mm}
\end{wrapfigure}


To better describe the unique properties of a safe RL problem, we provide the feasibility, optimality, and temptation definitions following the previous work~\citep{liu2022robustness}.
Their figure illustrations for one CMDP are presented in Fig. \ref{fig:definition}. 

\begin{definition} \textbf{Feasibility}.
The feasible policy class is the set of policies that satisfies the constraint with threshold $\kappa$: $\Pi^\kappa \coloneqq \{ \pi(a|s) : V_c^\pi(\mu_0)\leq \kappa, \pi \in \Pi \}$. A feasible policy should satisfy $\pi \in \Pi^\kappa$.
\end{definition}

\begin{definition} \textbf{Optimality}.
A policy $\pi^*$ is optimal in the safe RL context if 1) it is feasible: $\pi^* \in \Pi^\kappa$; 2) no other feasible policy has higher reward return than it: $\forall \pi \in \Pi^\kappa, V_r^{\pi^*}(\mu_0) \geq V_r^{\pi}(\mu_0)$.
\end{definition}
\vspace{-1mm}

We denote $\pi^*$ as the optimal policy.
Note that the optimality is defined w.r.t. the reward return within the feasible policy class $\Pi^\kappa$ rather than the full policy class space $\Pi$, which means that policies that have a higher reward return than $\pi^*$ may exist in a safe RL problem due to the constraint, and we formally define them as tempting policies because they are rewarding but unsafe:

\begin{definition} \textbf{Temptation}.
We define the tempting policy class as the set of policies that have a higher reward return than the optimal policy: 
$\Pi^T \coloneqq \{ \pi(a|s) : V_r^\pi(\mu_0) > V_r^{\pi^*}(\mu_0), \pi \in \Pi \}$.
\textbf{A tempting safe RL problem} has a  non-empty tempting policy class: $\Pi^T \neq \emptyset$.
\end{definition}

The existence of tempting policies is a unique feature and one of the major challenges of safe RL, since the agent needs to update the policy carefully to prevent being tempted when maximizing the reward. 
Suppose that a safe RL problem has no tempting policies, then one can simply maximize the reward to obtain the optimal policy, which is equivalent to solving a standard RL problem without considering the safety constraints.

\subsection{Methodologies towards safe reinforcement learning}

In this section, we describe how to solve the safe RL problem from different perspectives: optimization, planning, and control theory.

\subsubsection{Primal-dual-based method}~\\
The primal-dual framework is commonly used to solve constrained optimization problems, which introduces an additional Lagrange multiplier $\lambda$ to penalize constraint violations and gives rise to the weighted optimization objective:
\begin{equation}
    L(\pi, \lambda)=V_r^\pi(\mu_0)+\lambda(\kappa-V_c^\pi(\mu_0)).
\end{equation}
Manually selecting $\lambda$ is a straightforward solution to the Lagrangian dual problem and is applier in early researches~\cite{borkar2005actor, bhatnagar2012online}. However, hand-turning $\lambda$ is challenging and can lead to suboptimal solutions, as the constraint satisfaction performance is sensitive to $\lambda$. Recent works focus on 
adaptively tuning the multiplier such that the penalty weight for constraint violation could be dynamically adjusted based on the policy performance and the reward scale. We will first introduce the Lagrangian-based approaches, which is the most commonly used framework to learn the Lagrange multiplier, and then present other types of approaches that utilize additional constraints, approximations, or variational inference to overcome the limitations of the Lagrangian-based methods.


\textbf{Lagrangian-based method}. As a simple and effective framework, Lagrangian-based methods update the primal variable (i.e., policy parameter) and dual variable (i.e., Lagrange multiplier) in turn by solving the following min-max optimization problem:
\begin{equation}
    \min_{\lambda\geq 0}\max_{\theta} V_r^{\pi_\theta}(\mu_0) +\lambda(\kappa-V_c^{\pi_\theta}(\mu_0)),
\end{equation}
where we denote $\theta$ as the policy parameters. Alternating between the maximizing over $\theta$ via any unconstrained reinforcement learning algorithms and minimizing over the Lagrange multiplier $\lambda$ yields a series of Lagrangian-based methods to solve the safe deployment problem~\cite{yang2021wcsac}.
\citet{chow2017risk} propose PDO to update both primal parameters and dual variables by performing gradient descent based on on-policy estimations of the reward and cost value functions $V_r^{\pi_\theta}(\mu_0)$ and $J_c(\pi_\theta)$. \citet{liang2018accelerated} accelerate the dual variable learning process by incorporating an off-policy evaluation of $V_c^{\pi_\theta}(\mu_0)$ to learn the Lagrange multiplier in the dual problem, while keeping on-policy training for the primal update of the policy parameters. \citet{tessler2018reward} reformulate the Lagrangian dual problem and merged the cost into reward directly in each transition step to guide the policy, which can also handle mean value constraints. \citet{paternain2019constrained} prove strong duality holds in the Lagrangian dual problem of CMDP under mild assumptions and primal-dual algorithms can converge to a saddle point without a duality gap. \citet{stooke2020responsive} observe that the large phase shift between the constraint function and the Lagrangian multiplier could lead to unstable training and oscillation behaviors of the training curves, so they apply the idea of PID control to the dual variable learning phase such that the oscillation and overshoot behaviors could be mitigated.
The augmented Lagrangian multiplier is also adopted in recent works, which can achieve superior safety performance~\cite{as2022constrained}.
The Lagrangian-based method is a type of generic algorithm to learn a constraint satisfaction policy and can be easily incorporated into the classic RL method. 
However, approximately solving the minimization (dual problem) can lead to suboptimal dual variables for each iteration, and the non-stationary cost penalty term involving $\lambda$ will make the policy gradient step in the primal problem hard to optimize.

\textbf{Convex optimization for the dual variable.}
To deal with the performance instability during optimization, CPO~\cite{achiam2017constrained} extends the idea in TRPO~\cite{schulman2015trust} that adds additional trust region constraint during policy updating, and proposes to solve the dual variable via convex optimization after approximating the policy parameter space via second-order Taylor expansion. Similar to TRPO, it also reconciles the updating policy to the safe set via line search when violating the constraint. \citet{yu2019convergent} also propose to replace the non-convex constrained objective with an approximated convex quadratic function, and thus transform the original problem into a sequence of surrogate convex constrained optimization problems. Though the dual variables could be solved efficiently via convex optimization, the approximation errors can lead to poor constraint satisfaction performance in practice. 
\citet{liu2022constrained} further propose a variational inference-based approach -- CVPO -- to decompose the policy learning to a convex optimization phase with a non-parametric variational distribution and a supervised learning phase to fit the parameterized policy. 
CVPO solves the safe RL problem via a novel Expectation-Maximization style to naturally incorporate constraints during policy optimization, which provides both optimality and robustness guarantees of the learned policy.

\textbf{Projection-based}. Another type of method adds a projection step based on primal-dual algorithms to facilitate policy learning. \citet{zhang2020first} propose a first-order gradient descent method with trust region constraint which first calculate the optimal updated policy by closed-form solution and then projects it back into parametric policy space, which is more computationally efficient compared to other trust region method. PCPO~\cite{yang2020projection} first ignores cost and solves unconstrained RL and then projects the immediate policy into the safe set, which is obtained by approximation on cost constraint. \citet{yang2021accelerating} add a new step which projection immediate policy into the neighbor set of a baseline policy to accelerate learning. Although these methods improve safety during training effectively, constraint violation is still almost inevitable because of the extrapolation error of value function parameterized by neural network and approximation error in safe set estimation. 

\subsubsection{Uncertainty-aware method}~\\
We have introduced a series of Lagrangian-based methods that solve the safe RL problem from the optimization perspective, now we present approaches that consider from the data collection or action sampling perspective.
Instead of directly interacting with the environment based on the policy output, some approaches explicitly take uncertainty estimation into account to minimize the constraint violations. We generally categorize them into two groups: upper confidence bound-based approaches and dynamics uncertainty-based methods, where the former ones are usually model-free while the latter ones are usually model-based approaches.

\textbf{Upper confidence bound (UCB).} UCB-based methods leverage the uncertainty estimation to balance the exploration and exploitation during training and are common to explore tabular environment efficiently with bounded constraint violation. \citet{singh2020learning} and \citet{efroni2020exploration} apply UCB on reward, cost and transition function to accelerate sampling on tabular CMDP; their methods bound both reward regret and cost constraint violation via linear program or other primal-dual optimization approaches. Different from linear constraint, \citet{brantley2020constrained} propose UCB-based planning to deal with concave-convex and hard constraints setting for episodic tabular CMDP. \citet{ding2021provably} extended tabular constrained RL to CMDP with linear transition kernel and provide a provably efficient method in function approximation setting.

\textbf{Dynamics uncertainty}. Safe MPC~\citep{koller2018learning} utilizes Gaussian Processes to estimate the dynamics uncertainty and explore the environment safely to learn the dynamics. SAVED~\citep{thananjeyan2020safety} and RCE~\citep{liu2020constrained} both use an ensemble of neural networks as the dynamics model to estimate the dynamics prediction uncertainty and solve the constrained optimization problem via a model-predictive-control fashion, where the former formulates chance constraints to ensure safety from a probabilistic perspective, while the latter one considers the worst case unsafe scenario. CASRL~\citep{chen2021context} further extends previous approaches from stationary environments to non-stationary environments by modeling the non-stationary disturbances as probabilistic latent variables.

\subsubsection{Leveraging the prior knowledge for safe exploration}~\\
Different from previous approaches that focus on safety after the policy training, some researchers also care about safety during the training process, i.e., providing certain safety guarantees during the exploration. Apparently, it is hard to guarantee the policy always safety certain safety constraints during the training without any prior knowledge of the system, safety constraints, and the environment. Therefore, some studies leverage prior or domain knowledge of specific tasks to achieve safe exploration under certain assumptions. Most approaches among them borrow ideas from the control community, such as the Lyapunov theory and control barrier functions, to leverage the prior knowledge of the system dynamics. Another type of approach utilizes the structure of safety constraints to achieve safe exploration.

\textbf{Lyapunov-based method}:
Lyapunov theory can provide useful insights into the stability and safety of control systems. In control theory, safety is usually defined through the region of attraction that could be computed via Lyapunov functions. The key idea of using Lyapunov functions is to show the stability of the system, which is similar to showing the stability of gradient descent on strictly quasiconvex functions. If one can
show that applying the policy and dynamics to the state could result in strictly smaller values
on the Lyapunov function, then the state eventually converges to the equilibrium point at the origin. \citet{berkenkamp2017safe} utilize this principle to achieve safe exploration during the learning time with Lipschitz continuity assumption of the dynamics and policy classes. \citet{richards2018lyapunov} further extend the previous work by constructing a neural network Lyapunov function. The proposed safe training algorithm could adapt the safety certificates to the shape of the largest safe region in the state space, while relying only on knowledge
of inputs and outputs of the dynamics, rather than on any specific model structure. In contrast to restricting the exploration space, \citet{chow2019lyapunov} propose a Lyapunov-based safe policy gradient algorithm to solve the constrained optimization problem with unknown dynamics and constraints.

\textbf{Control barrier functions-based method}: Control barrier functions are another useful tool in control theory, which provides a continuous function for penalizing unsafe states or actions to infinite cost. \citet{li2019temporal} propose a temporal logic-guided approach with Lyapunov functions and control barrier functions to safeguard the policy exploration and deployment, which allows users to specify task objectives and constraints in different forms and at various levels. \citet{cheng2019end} propose a controller architecture that combines (1) a model-free RL-based controller with (2) model-based controllers utilizing control barrier functions (CBFs) and (3) online learning of the unknown system dynamics, to ensure safety during learning. The proposed framework leverages the success of RL algorithms to learn high-performance controllers, while the CBF-based controllers both guarantee safety and guide the learning process by constraining the set of explorable policies. 

\textbf{Post-process safety layer}: 
Post-process safety layer is another type of method to guarantee safe exploration. It corrects all unsafe actions generated from policy and ensures state-wise safety, which exceeds other safe RL methods only with expected or cumulative constraints. \citet{pham2018optlayer} propose an optimization layer (OptLayer) to calculate the safe action closest to the original unsafe action. OptLayer consists of a full quadratic solver, and it will run the iterative interior point method by a forward propagation for every unsafe action. Safety layer~\cite{dalal2018safe} provides a closed-form solution to a similar optimization objective instead and simplifies computation effectively by assuming that at most one constraint is active at each time. Although these methods obtain zero constraint violations in certain environments, their safety guarantee relies on the precise prediction of single-step state transition and thus is only applicable to linear systems or others with relatively simple dynamics.
\citet{peng2022safe} propose to use a safe expert policy to guide the learning policy when the action is away from the expert's. 
\citet{yu2022towards} propose to perform action corrections by learning a safety editor policy simultaneously from the replay buffer, which can achieve efficient off-policy training.

\textbf{Formal methods}: Formal verification provides a rigorous way to provide safety for control systems. Temporal logic is a common formal method to evaluate a temporal behavior of the system, such as the multi-modal sensing information of a robot. Users could provide their domain knowledge to the temporal logic specifications that are described mathematically, precisely, and unambiguously, so that the behavior of a system could be formally defined. \citet{li2019temporal,liu2021recurrent} propose a temporal logic guided reinforcement learning algorithm to achieve safe exploration. The proposed framework allows users to specify task objectives and constraints in different mathematical forms and at various levels. \citet{puranic2021learning} also utilize signal temporal logic to learn control policies while satisfying pre-defined safety requirements.

\subsubsection{Other approaches}~\\
There are also some research solving the safe RL problem from other perspectives. \citet{mehta2020curriculum} introduce a curriculum learning structure to learn a safe policy under the supervision of a safety teacher that could guide the agent back to safe zones when it violates constraints. \citet{srinivasan2020learning} pretrain a safety critic to estimate the future failure probability of a safety-constrained policy, and later fine-tunes the first-stage policy to the target tasks using the learned safety precautions. 
\citet{turchetta2020safe} propose an adaptive supervisor to prevent agents from taking dangerous policies by choosing and activating reset controllers with a learnable strategy. \citet{thananjeyan2021recovery} propose to learn the constraint violating zones from offline data before the policy training, and recover the agent from unsafe states to safe zones by using a separate recovery policy.
\citet{sootla2022saute} wrap the environment by augmenting the state space such that the modified MDP could be solved by any RL algorithms and can achieve almost surely safety performance.



\subsection{Benchmarks and resources}

\textbf{Safe RL tasks and environments.}
The AI Safety Gridworlds environment~\citep{leike2017ai} provides a testing ground for tabular safe RL algorithms.
Safety Gym~\cite{ray2019benchmarking} is a Mujoco-based safe RL environment, which provides several high-dimensional continuous control tasks with different difficulty levels. 
Safety Gym attracts much attention in the safe RL community and has become one of the most popular public benchmarks for safe RL.
The Bullet Safety Gym~\citep{gronauer2022bullet} is a PyBullet simulation engine-based variant of safety gym, which provides more robots and safe RL tasks, while the Safe Control Gym~\citep{yuan2021safe} provides several classical control-based environments.


\textbf{Safe RL algorithm resources.}
Safety Gym~\cite{ray2019benchmarking} provides several constrained deep RL algorithms to establish baselines that future work can build on, as well as the Bullet Safety Gym~\citep{gronauer2022bullet}.
SafePO~\citep{jibenchmarking} is a public safe policy optimization algorithm benchmark that provides a good number of on-policy model-free safe RL method implementations.

\section{Generalization in Reinforcement Learning}
\label{sec:generalization}

\subsection{Overview}

Generalization in RL focuses on designing algorithms to produce policies that can transfer or adapt to a variety of environments, without being overfitting to the training environments. This capability is crucial to the real-life deployment of RL agents since the environments at the testing time are often different from the training environments or inherently dynamic. There are several existing surveys that categorize RL generalization studies by their methodologies~\cite{kirk2021survey,zhao2020sim,vithayathil2020survey}. However, in this survey, we take a different approach to categorization by their evaluation variation.

The structure of this section is as follows: we first discuss the formalization of the contextual MDP on which the definition of generalizable RL is based in Section \ref{sec:def_generalizable_rl}. Then we describe the two dimensions of evaluation variation in Section \ref{sec:train_test_dist} and \ref{sec:components_mdp} respectively. Finally, we discuss the methodology categorization of generalizable RL in Section \ref{sec:approaches_generalizable_rl}

\subsection{Problem formulation for generalizable RL}\label{sec:problem_formulation_generalizable_rl}

\subsubsection{Definition of generalizable RL}\label{sec:def_generalizable_rl}~\\
To discuss generalization in a unified framework, we need to first formalize the concepts of a collection of environments. The formalization we adopt here is the Contextual Markov Decision Process (Contextual MDP), first proposed in \citet{hallak2015contextual}, and used by a recent survey in the generalization of RL~\cite{kirk2021survey}. 
A Contextual Markov Decision Process (MDP) extends the standard single-task MDP to a multi-task setting. In this work, we consider discounted infinite-horizon CMDPs, represented as a tuple $M=(\mathcal S, \mathcal Z, \mathcal A, R, P, p_0, \rho, \gamma)$. Here, $\mathcal S$ is the state space, $\mathcal Z$ is the context space, $\mathcal A$ is the action space, $R: \mathcal S \times \mathcal A \times \mathcal Z \mapsto \mathbb{R}$ is the context-dependent reward function, $P: \mathcal S \times \mathcal A \times \mathcal Z \mapsto \Delta(\mathcal{S})$ is the context-dependent transition function, $p_0: \mathcal{Z} \mapsto \Delta(\mathcal{S})$ is the context-dependent initial state distribution, $\rho \in \Delta(\mathcal{Z}) $ is the context distribution and $\gamma \in (0, 1)$ is the discount factor. Note that $\rho$ could be a time-dependent variable in the case of non-stationary environment distribution.

To sample a trajectory $\boldsymbol{\tau} :=\{(s_t, a_t, r_t)\}_{t=0}^\infty$ in Contextual MDPs, the context $z \sim \rho(\cdot)$ is randomly generated by the environment at the beginning of each episode. Here each $z$ defines an environment (i.e. task). With the initial state $s_0 \sim p_0(\cdot \;|\; z)$, at each time step $t$, the agent follows a policy $\pi$ to select an action $a_t \sim \pi(s_t)$ and receives a reward $ R(s_t, a_t, z)$. Then the environment transits to the next state $s_{t+1} \sim P(\cdot \;|\; s_t, a_t, z)$. In some special cases, $z$ is sampled at every time step so it changes throughout the episode.


Note that for most of the works studied in this survey, we do not emphasize the difference between the state and observation. However, we attempt to introduce the context-dependent emission function $o = \mathcal O(s | z)$ whenever the evaluation is focused on the generalization against apparent variation on the observation but not the underlying state of the environment. Examples would be only changing the color of objects in robotic manipulation tasks~\cite{hansen2021generalization} or the background distractors in control tasks with pixel observation~\cite{zhang2020learning}. With the formalization of contextual MDP, we are ready to define the generalization tasks.

\textbf{Definition of Generalizable RL:} assuming the agent is trained in the environment distribution $\rho_{\text{train}}(z)$ for $N_{\text{train}}$ environment steps, and the objective is to maximize the performance metric $P$ measured over a target task distribution $\rho_{\text{test}}(z)$ after $N_{\text{test}}$ environment steps:
\begin{alignat*}{2}
    &\max_{\pi^{(N_\text{test})}_{\text{test}}}~&&\mathbb{E}_{\pi, z \sim \rho_{\text{test}}(z)} P^{N_{\text{test}}}_{\text{test}}(\pi, z)\\ &s.t.~ &&\pi^{(N_{\text{train}})}_{\text{train}}=\arg \max_\pi \mathbb{E}_{\pi, z \sim \rho_{\text{train}}(z)} P^{N_{\text{train}}}_{\text{train}}(\pi, z),\\ &  &&\pi^{(0)}_{\text{test}} = \pi^{(N_{\text{train}})}_{\text{train}}
\end{alignat*}
where $P^N(\pi, z)$ represents the performance metric (that can be different for training and testing) of the policy $\pi$ on the task context $z$ after $N$ environment steps, $\pi^{(N)}$ represents the agent's policy after $N$ environment steps. Note that the word ``testing'' here does not necessarily mean the agent cannot be updated, rather refers to the ``target'' evaluation environment. Only in the zero-shot setting, the agent cannot be updated during testing.



\subsubsection{Relationship between the training and testing distribution}\label{sec:train_test_dist}~\\
In this section, we will describe the categorization of the relationship between the training distribution $\rho_{\text{train}}(z)$ and testing distribution $\rho_{\text{test}}(z)$. Here we have three categories: IID, OOD, and non-stationary.

\textbf{Independent-and-Identical Distribution (IID): } The training and testing environments are drawn from the same distribution, i.e., $\rho_{\text{train}}(z) = \rho_{\text{test}}(z)$. Note that although we are describing the relationship between the distributions, the RL agent is usually trained or tested on a set of sample MDPs drawn from the distribution. Therefore, even though the training and testing are drawn from the same distribution, it is still a non-trivial problem since the agent may never experience the exact same environment during training. Qualitatively, with a smaller number of distinct contexts seen during training, the IID generalization becomes more difficult since the training environments are sparsely sampled and cannot represent the true distribution well. 

\textbf{Out-of-Distribution (OOD) (stationary): } The training and testing environments are drawn from the different distributions, i.e., $\rho_{\text{train}}(z) \neq \rho_{\text{test}}(z)$. This mismatch could be due to many reasons. For example, the exact testing distribution is unknown or difficult to model. One of the motivations for studying OOD generalization is to enable Sim-to-Real transfer. Due to the low data efficiency of RL, it is a common practice to train agents in simulators and then transfer the agents to the physical world. However, even the state-of-the-art simulators cannot capture the real world perfectly, and mismatches between the simulated and real environments can cause catastrophic failures during deployment. Another motivation is combinatorial generalization, where the testing contexts take values seen during training \textit{independently}, but in unseen \textit{combinations}. For example, during training, the agent has seen context combinations $(a_0, b_0), (a_0, b_1), (a_1, b_0)$, while it is tested on $(a_1, b_1)$.

\textbf{Non-stationary: } The testing environments are drawn from time-variant, non-stationary distributions. Although there is some overlap between the OOD and non-stationary, studies categorized into non-stationary focus on dealing with the non-stationarity explicitly, i.e., life-long or continuous RL~\cite{thrun1995lifelong}. In this setting, there exists a (often infinite and unknown) sequence of testing distributions, $\{\rho_{\text{test}, 1}(z), \rho_{\text{test}, 2}(z), \ldots\}$. The agents may have to learn how to leverage past experience, identify new distribution, and avoid catastrophic forgetting. To evaluate these properties, we can examine, for example, whether the agent experiences a significant performance drop when encountering the transitions of testing distribution.

\subsubsection{MDP components variation of the generalizable RL}\label{sec:components_mdp}~\\
With the relationship between training and testing distribution described above, we now examine what component(s) of MDP the context controls, which constitutes another dimension of evaluation variation. There are four context-dependent components of MDP: initial state distribution, dynamics, reward function, and observation emission function.

\begin{table}[]
\centering
\caption{Categorization of generalizable RL. Each row represents the MDP components variation during evaluation, and each column represents the distributional variation during evaluation. We use colors to represent different methodologies of generalizable RL. \envdesign{red}: generalize by environment design; \oladapt{blue}: generalize by online adaptation; \syslearn{green}: generalize by system learning. If a method uses two or more approaches, it will be colored with corresponding colors. 
}
\label{tab:generalization}
\begin{tabular}{ |p{2.0cm}||p{3.5cm}|p{3.5cm}|p{3.5cm}|}
 \hline
 \textbf{Evaluation Variation} & \textbf{IID ($p_{\text{train}}(z) = p_{\text{test}}(z)$)} & \textbf{OOD ($p_{\text{train}}(z) \neq p_{\text{test}}(z)$)} & \textbf{Non-Stationary ($\{p_{\text{test}, 1}(z), p_{\text{test}, 2}(z), \ldots\}$)}\\
 \hline
 \textbf{Observation emission function $o = \mathcal O(s | z)$}   & \syslearn{Homomorphic MLP~\cite{van2020mdp}}, \syslearn{DBC~\cite{zhang2020learning}} & \envdesign{SODA~\cite{hansen2021generalization}}, \envdesign{PLR~\cite{jiang2021prioritized}}, \envdesign{DR-Sim2Real~\cite{tobin2017domain}}, \envdesign{ViewInvariantServo~\cite{sadeghi2018sim2real}} & \oladapt{Benna-Fusi-RL~\cite{kaplanis2018continual}} \\
 \hline
 \textbf{Initial state distribution $\rho_0(s | z)$} &  \envdesign{ALP-GMM~\cite{portelas2020teacher}} & \envdesign{SODA~\cite{hansen2021generalization}}, \envdesign{DR-Sim2Real~\cite{tobin2017domain}}, \envdesign{DR-}\syslearn{RDPG~\cite{peng2018sim}}  & \oladapt{Benna-Fusi-RL~\cite{kaplanis2018continual}}\\
 \hline
 \textbf{Dynamics $P(s, a | z)$}   & \envdesign{RARL~\cite{pinto2017robust}}, \envdesign{RRL-Stack~\cite{huang2022robust}},
 \envdesign{ActiveDR~\cite{mehta2020active}}, \envdesign{ALP-GMM~\cite{portelas2020teacher}},  \syslearn{ICIL~\cite{bica2021invariant}},   \syslearn{HB-MTRL~\cite{wilson2007multi}}, \syslearn{ML-GP~\cite{saemundsson2018meta}}, 
 \oladapt{UP-OSI~\cite{yu2017preparing}}, \oladapt{OSC-RL~\cite{kaspar2020sim2real}}, \oladapt{Agile-}\envdesign{Quad~\cite{tan2018sim}}, \oladapt{VAE-MAML~\cite{arndt2020meta}}, \oladapt{GrBAL~\cite{nagabandi2018learning}}, \oladapt{EPOpt~\cite{rajeswaran2016epopt}} & \envdesign{ActiveDR~\cite{mehta2020active}}, \envdesign{AutoDR~\cite{akkaya2019solving}}, \envdesign{BayRn~\cite{muratore2020bayesian}},
 \envdesign{PAIRED~\cite{dennis2020emergent}}, \envdesign{DR-}\syslearn{RDPG~\cite{peng2018sim}}, \oladapt{UP-OSI~\cite{yu2017preparing}}, \oladapt{GrBAL~\cite{nagabandi2018learning}}, \oladapt{EPOpt~\cite{rajeswaran2016epopt}}, \oladapt{T-MCL~\cite{seo2020trajectory}} & \syslearn{DPGP-RL~\cite{xu2020task}}, \oladapt{UP-OSI~\cite{yu2017preparing}}, \oladapt{GrBAL~\cite{nagabandi2018learning}}, \oladapt{PG-ELLA~\cite{ammar2014online}}, \oladapt{Benna-Fusi-RL~\cite{kaplanis2018continual}} \\
 \hline
 \textbf{Reward function $R(s, a | z)$}   & \oladapt{MAML~\cite{finn2017model}}, \oladapt{meta-}\envdesign{ADR~\cite{mehta2020curriculum}},  \syslearn{HB-MTRL~\cite{wilson2007multi}} \syslearn{MISA~\cite{zhang2020invariant}} & \syslearn{GRADER~\cite{ding2022generalizing}} & \oladapt{Benna-Fusi-RL~\cite{kaplanis2018continual}}\\
 \hline
\end{tabular}
\end{table}

\textbf{Observation emission function $o = \mathcal O(s | z)$:} The context-dependent emission function outputs the observation based on the state and the context. Since most studies do not explicitly differentiate between state and observation, we only introduce this term when the context creates a variation on the observation but not the underlying state of the environment. Examples: changing the color of the object or the view angle of the camera when using the pixel images as inputs to the policy.

\textbf{Initial state distribution $\rho_0(s | z)$:} The context-dependent initial state distribution outputs the initial state distribution based on the context. When the context changes the underlying state of the system, for example, the initial positions of objects for the robotic manipulation tasks, the shape, and the location of obstacles for robotic locomotion tasks.

\textbf{Dynamics $P(s | s, a, z)$:} The context-dependent dynamics outputs the distribution over the next state based on the previous state, the action, and the context. Examples: physical parameters such as gravity constant, friction coefficient, or perturbation forces applied to the robot. 

\textbf{Reward function $R(s, a | z)$:} The context-dependent reward function outputs a scalar reward based on the state, the action, and the context. The goal-conditioned MDP~\cite{Kaelbling93learningto,pmlr-v37-schaul15,andrychowicz2017hindsight} can be considered as a MDP with a context-dependent reward function. Here, the goal is equivalent to the context.

With the two dimensions of the evaluation variation described in Section \ref{sec:train_test_dist} and \ref{sec:components_mdp}, we classify a list of papers into Table~\ref{tab:generalization}. Note that classifying one paper into one category does not necessarily imply the method itself can only be applied to that evaluation variation. Instead, we consider the evaluation actually appearing in the paper. From Table~\ref{tab:generalization}, we observe that most of the papers focus on the dynamics variation for IID, OOD, and non-stationary. This phenomenon might result from the popular applications in robotic Sim-to-Real transfer, which is one of the main motivations of generalizable RL.



\subsection{Methodologies toward generalizable reinforcement learning} \label{sec:approaches_generalizable_rl}

To improve the generalization, one can approach from three different perspectives: designing environments that guide training (\textbf{Environment Design}), learning the features of environments (\textbf{System Learning}), or designing a learning algorithm that can adapt fast to diverse tasks during testing (\textbf{Online Adaptation}). 
Next, we will introduce the three main approaches.
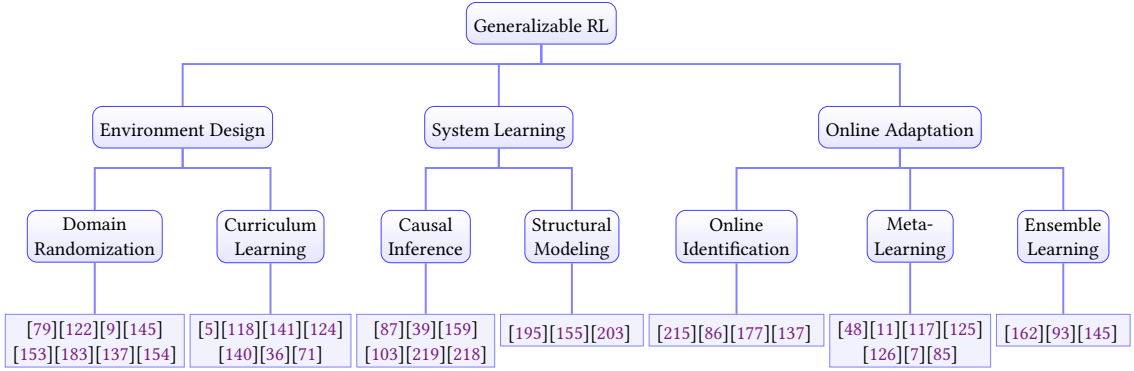
\begin{figure}[t]
    \centering
    \resizebox{1.0\textwidth}{!}{
\usetikzlibrary{trees}
\begin{tikzpicture}[
    i/.style={
        anchor=north,
        align=center,
        top color=white,
        bottom color=blue!10,
        rectangle,rounded corners,
        minimum height=6mm,
        draw=blue!75,
        align=center,
        color=blue!75,
        text=black,
        text depth = 0pt
    }
]
    \tikzset{execute at begin node=\strut}
    \tikzset{font=\small,
        grow=down,
        level distance=1.5cm,
        every tree node/.style={align=center,anchor=north},
        every leaf node/.style=
            {
                anchor=north,
                align=center,
                fill=blue!5,
                rectangle,
                draw=blue!50,
                align=center,
                distance=1cm,
                text depth = -10pt
            },
        edge from parent/.style=
            {
                draw=blue!50,
                thick,
                edge from parent path={(\tikzparentnode.south)
                -- +(0,-8pt)
                -| (\tikzchildnode)}
            }
    }

    \Tree [.\node[i] {Generalizable RL};
    [
        .\node[i] {Environment Design};
        [
            .\node[i]{Domain\\ Randomization};
            {
            \cite{jakobi1997evolutionary}\cite{mordatch2015ensemble}\cite{antonova2017reinforcement}\cite{rajeswaran2016epopt}\\\cite{sadeghi2016cad2rl}\cite{tobin2017domain}\cite{peng2018sim}\cite{sadeghi2018sim2real}
            }
        ]
        [
            .\node[i]{Curriculum\\Learning};
            {
            \cite{akkaya2019solving}\cite{mehta2020active}\cite{portelas2020teacher}\cite{muratore2020bayesian}\\\cite{pinto2017robust}\cite{dennis2020emergent}\cite{huang2022robust}
            }
        ]
    ]
    [
        .\node[i] {System Learning};
        [
            .\node[i] {Causal\\Inference};
            {
            \cite{ke2019learning}\cite{ding2022generalizing}\cite{scherrer2021learning}\\\cite{lindgren2018experimental}\cite{zhang2020learning}\cite{zhang2020invariant}
            }
        ]
        [
            .\node[i] {Structural\\ Modeling};
            {
            \cite{wilson2007multi}\cite{saemundsson2018meta}\cite{xu2020task}
            }
        ]
    ]
    [
        .\node[i] {Online Adaptation};
        [
            .\node[i] {Online\\ Identification};
            {
            \cite{yu2017preparing}\cite{kaspar2020sim2real}\cite{tan2018sim}\cite{peng2018sim}
            }
        ]
        [
            .\node[i] {Meta-\\Learning};
            {
            \cite{finn2017model}\cite{arndt2020meta}\cite{mehta2020curriculum}\cite{nagabandi2018learning}\\\cite{nagabandi2018deep}\cite{ammar2014online}\cite{kaplanis2018continual}
            }
        ]
        [
            .\node[i] {Ensemble\\Learning};
            {
            \cite{seo2020trajectory}\cite{kumar2020one}\cite{rajeswaran2016epopt}
            }
        ]
    ]
    ]
\end{tikzpicture}}
    \caption{Categorization of generalizable RL methods}
    \label{fig:tree-generalization-rl}
    \vspace{-0.1in}
\end{figure}

\subsubsection{Environment design}~\\
Since the challenges to generalizable RL often come from the discrepancy between the training and testing environments, it is natural to ask such a question: how to design/generate training environments in order to improve the testing performance?

\textbf{Domain Randomization.} Domain randomization is a technique to diversify the variability of simulation environments during the training time so that the policy is able to generalize to the real-world environment during testing time. Early works use randomized model parameters that are not critical to the robot's performance~\cite{jakobi1997evolutionary}, training on perturbed dynamic models~\cite{mordatch2015ensemble, antonova2017reinforcement, rajeswaran2016epopt} and randomized vision features~\cite{sadeghi2016cad2rl}, before the terminology \textit{domain randomization} was first coined in~\cite{tobin2017domain}. The simulator features that could be randomized include but are not limited to:
\begin{itemize}
    \item Controller properties such as noise of torque, control frequency and delay, controller gain, and damping~\cite{peng2018sim}.
    \item Physics parameters such as mass, gravity, and friction~\cite{peng2018sim}.
    \item Visual inputs such as lighting, textures of objects, position, orientation, and field of view of the cameras~\cite{tobin2017domain, sadeghi2018sim2real}.
\end{itemize}

\textbf{Curriculum Learning.} The aforementioned works rely on handpicking randomized features and values, which inevitably requires some degrees of parameter tuning to find the proper values or distributions of the randomized features. In order to minimize human labor, there is an increasing interest recently in developing methods for automatic domain randomization and curriculum learning. Active/Automatic domain randomization uses the performance of models as a feedback signal to update the distribution of training contexts. \citet{akkaya2019solving} proposes Automatic Domain randomization which gradually increases the difficulty of training environments only when a minimum level of performance is achieved. \citet{mehta2020active} propose Active Domain Randomization to learn a sampling strategy that produces the most informative environment by leveraging the performance difference between the randomized and reference environments. \citet{portelas2020teacher} learns a difficulty level and an order of environments to set as training environments by iteratively sampling and fitting a Gaussian mixture model so that it maximizes the efficiency of learning. \citet{muratore2020bayesian} uses Bayesian optimization to search the space of source domain distribution parameters for learning a robust policy that maximizes real-world performance.

Adversarial training is another common formulation of curriculum learning. RARL~\cite{pinto2017robust} formulates the curriculum learning as a two-player zero-sum game between the agent and an adversary applying perturbation forces. Empirical studies find this kind of adversarial training can improve the generalization of the RL agent. Differently from the aforementioned zero-sum simultaneous game setting, PAIRED~\cite{dennis2020emergent} assumes the adversary is maximizing the regret, which is defined as the performance gap between the optimal agent and the current agent. RRL-Stack~\cite{huang2022robust} proposes to use a general-sum Stackelberg game formulation to address the potential problems of unsolvable environments and unstable training.

One important assumption of curriculum learning is the access and control of the environment context since the algorithms need to actively change the distributions of training environments. Compared with other methods discussed in later sections of our survey, curriculum RL does require more information and control over the environment. However, in most cases, this is not considered restrictive since RL agents are usually trained in simulators therefore access is not a major constraint. As an emerging area, the empirical evaluations of curriculum RL have attracted a lot of attention recently, but there is very little theoretical understanding of how or why it works in practice. We believe there will be more research in this area soon.

\subsubsection{System learning}~\\

\textbf{Causality and Invariant Feature Discovery.} 
There usually exist underlying causality behind the dynamic system and tasks, which determines the mechanism of how things happen or how one object influence another object. Discovering such causality is a popular topic in RL, where the methods can be generally categorized into explicit learning and implicit learning. 
Explicit learning methods estimate the graph structure by either differentiable optimization~\cite{ke2019learning}, which learns a soft adjacent matrix with interventional data, or by statistic independent test~\cite{ding2022generalizing}, which iteratively updates the world causal model and planning policy.
Active intervention methods are also explored in ~\citet{scherrer2021learning} to increase the efficiency of data collection and decrease the cost of conducting intervention~\cite{lindgren2018experimental}.
Implicit learning focus on extracting invariant feature from multiple environments that share the same underlying causality~\cite{zhang2020learning, zhang2020invariant}. This kind of feature is assumed to be the only important factor that influences the dynamics. Therefore, different systems can be represented by assigning different values to this variable.

\textbf{Structural Modeling}.
In addition to learning the causality hidden in the environments, structural modeling aims to establish an explicit relationship between the environments. A common choice for such a relationship is to maintain a hierarchical probabilistic model. In \citet{wilson2007multi}, the authors propose Hierarchical Bayesian Multi-Task RL (HB MTRL), which uses hierarchical Bayesian model classes of MDPs. Each class of MDPs shares some common structure and the shared knowledge can be transferred among MDPs of the same class. \citet{saemundsson2018meta} uses a hierarchical latent variable model with Gaussian Process to model dynamics and infers the relationship between tasks automatically from the collected data. Similarly, \citet{xu2020task} uses Dirichlet-Process-Gaussian-Process (DPGP) to model the MDPs distribution and maintains a mixture of experts to handle the non-stationarity. 
\subsubsection{Online adaptation}~\\
\textbf{Online Identification. }
Online Identification takes a different approach from a reversed perspective of most generalizable RL. Instead of learning a policy that performs well over a wide range of environments, online identification identifies parameters of a dynamic model in the real world and then use them as information for the controller. \citet{yu2017preparing} trains an Online System Identification (OIS) to predict the dynamics model parameters and feeds them to the controller along with system states. \citet{kaspar2020sim2real} executes predefined actions on the real robot to collect trajectories and optimizes the simulator parameters to align the simulated trajectories with real trajectories. \citet{tan2018sim} also uses collected real trajectories to fine-tune the simulator and match trajectories. 
Online Identification can also be incorporated into the policy by using a recurrent model conditioned on a latent variable representing the history of states and actions~\cite{peng2018sim}.

\textbf{Meta-Learning. } Meta-Learning, or learning-to-learn, aims to learn from past experience so that it can adapt fast to the new unseen testing tasks. Model-Agnostic Meta-Learning (MAML) is first introduced in \citet{finn2017model} to train a model on a variety of learning tasks using gradient-based methods, such that it can adapt to new learning tasks using only a few training samples. \citet{arndt2020meta} applies MAML with a task-specific trajectory generation model to facilitate exploration and deploy trained policy on a real robot. \citet{mehta2020curriculum} considers the importance of task distributions in gradient-based Meta-learning and proposes meta-ADR to find a curriculum of tasks to avoid various problems associated with meta-learning.

\citet{nagabandi2018learning, nagabandi2018deep} uses meta-learning to train a dynamics model prior such that this prior can be rapidly adapted to the testing environments. \citet{ammar2014online} proposes a multi-task policy gradient method to learn and transfer knowledge between tasks to accelerate learning with theoretical guarantees.  \citet{kaplanis2018continual} provides another perspective to equip RL agents with a synaptic model that can mitigate catastrophic forgetting.

\textbf{Ensemble Learning. } Ensemble learning aims to learn a collection of models and combine them to improve performance during testing. Different from Structural Modelling, Ensemble learning does not focus on maintaining an explicit (mostly hierarchical) relationship between models. Instead, it often assumes a flat structure of models and focuses on policy adaptation during testing time. From a model-based perspective, \citet{seo2020trajectory} learns an ensemble of dynamic models by updating the most accurate model during training and incorporating an online adaptive mechanism by extracting contextual information using the sampled trajectory. From a model-free perspective, SMERL~\cite{kumar2020one} aims to learn a diverse set of policies so that it can generalize to the unseen environment by adopting the best solutions. \citet{rajeswaran2016epopt} learns a robust policy based on an ensemble of the training environments and keeps adapting the distribution over the training environments using the data from the testing environments.

\subsection{Application benchmarks and resources}

There exist many benchmark environments for generalizable RL. In general, the benchmark environments for generalization can be differentiated by their application domains: control/robotics (e.g. RoboSuite~\cite{zhu2020robosuite} and DMC~\cite{grigsby2020measuring}), or game-like arcade (e.g. Arcade Learning Environment game modes~\cite{bellemare2013arcade} and Phy-Q~\cite{xue2021phy}). Note that it is challenging to assess the difficulty of the environment just by its application domain. A grid-world-like environment may impose greater challenges in generalization than continuous control tasks, depending on the underlying experimental settings. 

Most of the benchmark environments provide the users with great flexibility to control the training and testing contexts. Therefore, many of them are not tied to a specific evaluation variation, such as IID, OOD, or non-stationary. It also highlights the difficulty of creating a standard training/testing protocol for generalizable RL, in contrast to the supervised learning domain. 















\section{Outlook and future research}
\label{sec:challenges}

In this section, we will provide our outlook for promising research directions. 
The first upfront task is that we should be able to evaluate and certify RL regarding trustworthiness requirements. Second, we should understand whether there would be some fundamental trade-offs between the trustworthy aspects, as well as the trade-off between the  requirements of trustworthiness and RL's nominal performance. Third, besides the intrinsic ''digital trustworthy RL'', how ''physical trustworthy RL'' would be influenced by the agent's physical design and operational environment setting? Finally, we will discuss the ''societal trustworthy RL'' comprising human modeling, privacy, trust, ethics, and fairness. 

\subsection{How to certify and evaluate trustworthy reinforcement learning?}

\subsubsection{How to design trustworthy RL with certification?}~\\
\indent It is challenging and important to provide certification for the robustness, safety, or even generalization of reinforcement learning models and systems. Although there are some existing works providing certified \textit{robustness} for RL under different settings~\cite{wu2022crop,wu2022copa,kumar2021policy,lutjens2020certified,wang2019verification}, the constraints for the attacks are limited (e.g., $\ell_p$ norm bounded input state perturbation) and the certified goal such as action consistency is not practical enough for safe deployment.
In particular, the dynamic nature of RL and complicated environment constraints and uncertainties that cannot be modeled accurately contribute to the high complexity in this direction.
Thus, more realistic certification goals such as robust action dynamics which take the temporal property RL into account are promising and more practical.

From the \textit{safety} perspective, which focuses on a set of specified safety constraints, the certification goal is more specific considering these constraints. However, additional challenges are also introduced given the fact that 1) these constraints are usually hard to characterize in an end-to-end fashion, 2) new certification techniques are required since most existing certification methods are only based on \textit{prediction consistency} without explicitly taking specific (safety) goals into account, 3) these constraints are usually correlated with others or even the environments which makes the certification process of high complexity.
As a result, how to  certify the safety of RL by categorizing the safe constraints into different levels, and characterizing the sufficient conditions for certifying safety is of great importance.

Finally, certifying the \textit{generalization} for RL has a lot of real-world implications. For instance, if an autonomous vehicle is trained in a rural environment, it is important to know its performance certification when driving in an urban area. In this case, how to measure the difference between different environments or observations, how to define the certification goal (e.g., the lower bound of reward given shifted observations), and how to train a certifiably generalizable RL would lead to interesting work with both theoretical and practical impacts. 
In addition, the generalization and robustness certifications are naturally connected. For instance, the robustness certification is on the individual input level while generalization certification is on the input distribution level; certification for robustness can be viewed as the "worst-case" shift while that for generalization can be viewed as certifying under "nature distribution shift".
In this way, some certification techniques for robustness can be potentially leveraged for certifying generalization under different conditions. How to analyze their fundamental connections or indications is also an interesting direction to be explored. 

\vspace{-0.1in}

\subsubsection{How to design evaluation platforms for trustworthy RL?}~\\
\indent It is vital to design proper platforms and evaluation metrics to assess the trustworthiness of a proposed RL algorithm. 
For a continuous environment, 
the most common platform is the MuJoCo environment \cite{todorov2012mujoco}, where one can vary the agent's mass, ground friction, joint damping, and armature to test robustness. Atari Game is another popular platform for observation-based attacks.
Safety evaluation environments, such as the AI Safety Gridworlds environment \citep{leike2017ai}, Safety Gym~\cite{ray2019benchmarking}, and Bullet Safety Gym \citep{gronauer2022bullet}, design control tasks with safety zones. 
The evaluation of generalization focus on changing irrelevant factors, e.g., the background of the environment~\cite{cobbe2020leveraging}. In addition, compositional generalization is evaluated by combining different pre-defined sub-tasks~\cite{mendez2022composuite}.

These platforms greatly facilitate the development of trustworthy RL.
However, there are three critical limitations of these existing benchmarks.
The most important one is still a large sim-to-real gap between the benchmark and the real world. Evaluation of robustness, safety, and generalization only in simple tasks may not generalize well in complex real-world applications.
The second limitation is that most of the benchmarks use pre-defined tasks and parameters set by the creators, which lacks diversity and may be subject to human biases.
Last but not least, although tasks are usually randomly sampled, the distribution rarely triggers critical events with catastrophic consequences, resulting in an underestimation of risks and very slow convergence of the results~\cite{ding2022survey}. Some recent benchmarks~\cite{xu2022safebench} use realistic 3D simulators to construct real-world scenarios and use accelerated evaluation methods~\cite{xu2021accelerated, arief2021deep} to emphasize the rare safety-critical cases. However, there is a trade-off between the modeling error and evaluation error~\cite{huang2019evaluation}.


These attempts are pioneers toward the efficient and accurate evaluation of certain aspects. However, we are still far away from attaining satisfying benchmarks for evaluating robustness, safety, and generalization. 
Comprehensive and standard benchmarks are urgently required to make RL algorithms deploy in the real world.


\subsection{What is the relation between the different aspects of trustworthiness?}

We have discussed the trustworthiness of RL from three different aspects.
As a result, we have at least four dimensions to evaluate a trustworthy RL system: the safety, robustness, and generalization aspects of the system as well as the nominal task performance of an RL system i.e. the original functional goal of the RL.
However, how to design a trustworthy RL system that is safe, robust, generalizable, and has high task performance is still an open problem and is rarely studied in the literature.
To better understand this problem, we need to answer one question first: what are the relations between the four aforementioned dimensions?
While most existing papers only focus on one or two of them, we believe that they are not orthogonal to each other.
We provide several hypotheses and thoughts regarding the relations between them, which might be helpful for future trustworthy RL studies.


\textbf{The trade-off between trustworthiness and task performance.} We could observe that improving any aspects of trustworthiness might potentially induce a drop in the best possible task performance.
For instance, improving the safety of an RL agent may lead the agent to be conservative in exploring high-rewarding regions, and thus has relatively lower task performance than the unsafe one~\cite{liu2022robustness}; increasing the robustness against adversarial perturbations may over-smooth the policy, and thus decrease the task performance; training a generalizable policy on multiple tasks may decrease the performance on a single task due to the limitation of model capacity.
As a result, we can see that improving the trustworthiness may be at the cost of sacrificing the optimal task performance, which is also a reflection of the no-free-lunch theorem.
How to determine the sweet point to balance them is an interesting problem, and how to better understand the trade-offs could help us adjust the trustworthiness as needed based on different applications.

\textbf{The interconnections between the trustworthiness aspects.} We believe that the conceptions of safety, robustness, and generalization have non-negligible overlaps and are not orthogonal to each other, though they are mostly discussed independently in the RL literature.
For example, we could view a robust RL policy in terms of observation noises as a generalization capability to the states around the training samples. Similarly, a generalizable policy for unseen scenarios can be regarded as the robustness property to dynamics uncertainties. 
Therefore, they are inseparable in a certain context, and we can also see similar discussions and thoughts in the general machine learning domain \cite{xu2012robustness}.
On the other hand, safety is also closely related to generalization and robustness, because an agent cannot be regarded as safe if it is not generalizable to novel environments or robust against adversarial perturbations.
A recent work unveils the connections between robustness and safety by showing that a safe policy in a noise-free environment may not be safe under adversarial attacks \cite{liu2022robustness}, and we hope to see more interdisciplinary discussions in this direction.











\subsection{How to co-design trustworthy RL with the physical agent and environment?}
While there are tremendous developments on the computational side of trustworthy RL, the progress of designing and optimizing the physical properties of the agents are generally lagged behind~\cite{miriyev2020skills}. Similar to organisms, the RL agents in the real world have both the "brain" which provides cognition capability and "body" which serves the sensing and actuating organs after decisions are made. Some prior works have explored this direction in simulated environments. To find a robot's morphology that is nearly optimal for a specific task, \citet{Gupta2021-kq} uses genetic algorithms to mutate and evolve a population of learning agents. Similarly, \citet{Ha2019-tr} uses sampling-based algorithms to update the physical parameters of the agent. More recently, \citet{yuan2021transform2act} proposes to use Graph Neural Networks to model both the control policy and morphing strategy in order to jointly optimize both of them. Most of these methods show advantages compared with non-co-optimized agents. 

Some key challenges of the "physical trustworthy RL" still remain. For example, how to model the physical structures, sensing, and actuation systems so that we could use gradient methods to efficiently train the whole agent. How to build an easy-to-use simulation platform to optimize a structural and modular system? How to design step-wise rather than episodic training to accelerate the training?
We believe co-design the software and hardware of RL agents would provide a path to  the future RL agents could be more adaptive to the changing test environments and more resilient to damages or adversarial environments, resulting in a new generation of trustworthy intelligent agents.



\subsection{How to achieve the human-centric design for trustworthy RL?}

In addition to the intrinsic perspectives of trustworthy RL discussed in this paper (i.e., robustness, safety, generalization), the extrinsic perspectives which mainly involve human feedback have also raised great attention and need to be further explored for the ''societal trustworthy RL''.
In particular, such extrinsic perspectives broadly include the \textit{explainability}, \textit{privacy} protection for sensitive individual information, \textit{ethics}, and general human trust on trained RL agents or models. 
Different trustworthy RL algorithms have also been studied in the literature by considering such human-centric design~\cite{akers2015social,nam2017predicting,tabrez2019improving,gao2019fast} to bridge the \textit{trust} with human.

\textbf{Human behavior modeling and human-machine interaction.}
First, it is important and challenging to model human behaviors and interactions between human and RL agents.
    For instance, \textit{behavior analysis} of an RL agent would be important to understand if a trained RL model is \textit{egoistic} or \textit{altruistic}, which is beneficial to build human trust
   In particular, human's sub-optimal behaviors cast a great challenge on human behavior modeling, which raises a key open problem in trustworthy RL.
Inferring an agent's intent given history observations or opponent policies is widely studied in inverse reinforcement learning (IRL) (see \cite{arora2021survey} and references therein). However, IRL assumes the demonstrated behaviors are near-optimal, which is frequently violated by human.
Human's sub-optimal behavior depends on various factors.
For instance, \citet{ziebart2008maximum} assumes human's sub-optimal behavior as an effect of \textit{random noise} inspired by the Boltzmann noisily-rational decision model in cognitive science \cite{baker2007goal} and economic behavior \cite{morgenstern1953theory}. 
Human's \textit{risk sensitivity} (risk-seeking or risk-averse) is also widely known to cause sub-optimal behaviors. 
\cite{ratliff2019inverse} uses the Prospect theory \cite{kahneman2013prospect} to account for decision makers' different altitudes towards gains and losses. 
\citet{reddy2018you} and \citet{golub2013learning} argue that human's sub-optimal is due to \textit{model missepecification}: humans's sub-optimal behavior w.r.t. true environment dynamics model may be near-optimal w.r.t. the user's internal beliefs of the dynamics model.

\textbf{Explainable RL for human trust.}
In addition to the formal modeling of human-machine interaction, making machine decision-making interpretable to human is a key step to building human trust.
Existing efforts have been made to construct structured RL to explore the semantics of the decision-making process~\cite{jaques2020human,radulescu2019holistic}, provide visualization for RL algorithms~\cite{wells2021explainable,rupprecht2019finding}, and explore low-dimensional interpretations~\cite{gjaerum2021explaining}.
However, the dynamic nature of RL and uncertainties in the environments make it challenging to fully provide a quantitative and verifiable explanation. Thus, it is important and interesting to explore different structures and architectures of RL, such as compositional structures and integration with domain knowledge to provide a semantic and logical explanation for RL.

\textbf{Privacy-preserving RL to protect sensitive individual information.}
As the training of RL algorithms requires a large amount of data, such as collected traffic scenarios/driving behaviors for autonomous driving and human medical statistics for intelligent diagnosis, the privacy issues of trained RL models have raised great concerns~\cite{sakuma2008privacy,liu2019privacy}.
Clearly, such data usually contains lots of sensitive information about individuals such as personal behaviors and health status; differentially private RL algorithms~\cite{ma2019differentially}, game theoretic approaches~\cite{cui2019improving}, and locally private RL~\cite{ono2020locally} have been proposed to protect the private information in RL. 
However, from the human perspective, it is not intuitive to understand the privacy protection levels. For instance, a patient may not know what level of $\epsilon$ is suitable for his/her personal data under the $(\epsilon,\delta)$-differentially private RL training.
As a result, new privacy measurement metrics, privacy protection criteria, and data valuation approaches need to be further studied to gain human trust. 

\textbf{Ethical and fair RL with human trust.}
Given the fact that RL agents are usually adopted in human daily life, such as treatment assistant robots and cleaning robots, it is of great importance to ensure the ethics of RL agents.
For instance, a recent large-scale benchmark has been provided to study the ethical issues in RL video games and try to remove toxic or violent dialogues in  games~\cite{hendrycks2021would}.
Similarly, different studies have been conducted to identify the fairness problems in RL as well as mitigate them~\cite{liu2020balancing,jabbari2017fairness}.
On the other hand, the large training observation space and non-linear decision space of RL make it more challenging to train ethical and fair RL agents with human feedback. Thus, not only from the technical perspectives such as different ethics and fairness issues identification and mitigation methods, but also from the policy-making perspectives such as ethics and fairness regulations for RL need to be better understood and studied. 

Overall, the extrinsic \textit{human trust} is widely studied in various contexts leading to different definitions, such as an attitude, an intention, or behavior \cite{lewis2018role}. Trustworthiness in automation \cite{lee2004trust} is defined as human's attitude that an automated agent will help achieve the individual's goal in uncertain situations. 
Researchers already show that trust affects human's reliance on intelligent agents, which may lead to under- or over-reliance and thus influence the overall system performance \cite{parasuraman1997humans}.
It is still an open problem about enhancing human trust in RL agents by endowing RL agents the capability to generate explainable behaviors, understand human preferences and even affect human's behaviors, respect human privacy, and facilitate fairness in teamwork.

\section{Conclusion}
\label{sec:conclusion}
In this survey, we aim to clarify the terminology of robustness, safety, and generalization of RL, analyze their intrinsic vulnerabilities, introduce work tackling these problems, and summarize popular benchmarks, thus bringing together disparate threads of studies together in a unified framework. We hope that this survey will serve as a touch-point and reference for scientists, engineers, and policymakers in the trustworthy RL domain and spur further research.

We summarise the key takeaways of this survey below.

\begin{itemize}
    \item Distributionally Robust RL algorithms provide a natural formulation to encode prior knowledge by choosing a proper ambiguity set. However, it is still unclear how to choose an appropriate ambiguity set.
    Hence principles of choosing ambiguity set should be focused on in the future.
    \item Robust training with adversaries could result in over-conservative policies and training instabilities. Better adversary attack formulation along with robust training should be proposed, such as attacks constrained by known rules or regulated by the RL agents' performance.
    \item Safe RL has multiple definitions in ML, control, and robotic communities, each with its corresponding practical applications. More studies are needed to understand the pros and cons of each direction for different use cases.
    \item Ensuring safety for a complex system is still a challenging problem, especially when the risk of constraint violations is hard to predict or the domain knowledge is limited. Ensuring safety during the RL training phase or in a nonstationary environment are still wide open areas.
    \item Although there exist many environments designed for evaluating RL, few are designed to evaluate their trustworthiness. The research community will largely benefit from a standard evaluation environment and pipeline that can provide diverse options for different evaluation protocols and variations.
    \item Compared to robust RL and safe RL, generalizable RL is a relatively new field. A unified paradigm with rigorous and verifiable theoretical analysis and standard benchmarks is urgently needed to advance this area.  
    \item Sim2Real transfer still poses great challenges to the ML  community, which is critical to the real-world deployment of RL agents. On the one hand, we need more publicly available training simulators with high fidelity in simulating complex domains such as soft objects, contacting forces, and realistic terrains, as well as effective evaluation and certification paradigms for rare but catastrophic scenarios. On the other hand, we need to develop more practical algorithms that are able to adapt online to unseen real-world environments.
    \item More research is needed to understand the interconnection between the different aspects of trustworthiness as well as the nominal functional performance of an RL system. For example, in settings with environment discrepancies, robust RL and generalizable RL are found to be closely related. The connections between safety and robustness or generalization are also observed yet have not been thoroughly established and discussed.
    \item While we mainly focus on the intrinsic vulnerabilities of RL for "digital trustworthy intelligence", more work is needed for "physical trustworthy intelligence" that considers both the RL algorithm design also the physical agent and environment design in the whole life cycle for trustworthy autonomy. This is still a wide-open area.
    \item Finally, intelligent autonomy requires the human-centric consideration for "societal trustworthy" RL, which includes human modeling and interaction, explainability, privacy, trust, ethics, and fairness.
    
\end{itemize}

\noindent \textbf{Acknowledgement}

\noindent We thank (in alphabetical order)
Mansur Arief, Hanjiang Hu, Haohong Lin, Linyi Li, Jielin Qiu, Fan Wu, Chulin Xie, and Jiacheng Zhu for discussion and comments on drafts of this work.\\

\noindent \textbf{Author Contributions}

\noindent Mengdi Xu, Zuxin Liu, and Peide Huang wrote most parts of Sections 2, 3, and 4 respectively. Wenhao Ding and Zhepeng Cen wrote parts of section 4 and 3, respectively. Bo Li and Ding Zhao led the formalism of the survey, guided the students, wrote Sections 1 and 6 and parts of Section 5, and revised the whole manuscript. 


\bibliography{refs,acmart}


\begin{thebibliography}{227}


\ifx \showCODEN    \undefined \def \showCODEN     #1{\unskip}     \fi
\ifx \showDOI      \undefined \def \showDOI       #1{#1}\fi
\ifx \showISBNx    \undefined \def \showISBNx     #1{\unskip}     \fi
\ifx \showISBNxiii \undefined \def \showISBNxiii  #1{\unskip}     \fi
\ifx \showISSN     \undefined \def \showISSN      #1{\unskip}     \fi
\ifx \showLCCN     \undefined \def \showLCCN      #1{\unskip}     \fi
\ifx \shownote     \undefined \def \shownote      #1{#1}          \fi
\ifx \showarticletitle \undefined \def \showarticletitle #1{#1}   \fi
\ifx \showURL      \undefined \def \showURL       {\relax}        \fi
\providecommand\bibfield[2]{#2}
\providecommand\bibinfo[2]{#2}
\providecommand\natexlab[1]{#1}
\providecommand\showeprint[2][]{arXiv:#2}

\bibitem[\protect\citeauthoryear{Abdullah, Ren, Ammar, Milenkovic, Luo, Zhang,
  and Wang}{Abdullah et~al\mbox{.}}{2019}]%
        {abdullah2019wasserstein}
\bibfield{author}{\bibinfo{person}{Mohammed~Amin Abdullah},
  \bibinfo{person}{Hang Ren}, \bibinfo{person}{Haitham~Bou Ammar},
  \bibinfo{person}{Vladimir Milenkovic}, \bibinfo{person}{Rui Luo},
  \bibinfo{person}{Mingtian Zhang}, {and} \bibinfo{person}{Jun Wang}.}
  \bibinfo{year}{2019}\natexlab{}.
\newblock \showarticletitle{Wasserstein robust reinforcement learning}.
\newblock \bibinfo{journal}{\emph{arXiv preprint arXiv:1907.13196}}
  (\bibinfo{year}{2019}).
\newblock


\bibitem[\protect\citeauthoryear{Achiam, Held, Tamar, and Abbeel}{Achiam
  et~al\mbox{.}}{2017}]%
        {achiam2017constrained}
\bibfield{author}{\bibinfo{person}{Joshua Achiam}, \bibinfo{person}{David
  Held}, \bibinfo{person}{Aviv Tamar}, {and} \bibinfo{person}{Pieter Abbeel}.}
  \bibinfo{year}{2017}\natexlab{}.
\newblock \showarticletitle{Constrained policy optimization}. In
  \bibinfo{booktitle}{\emph{International Conference on Machine Learning}}.
  PMLR, \bibinfo{pages}{22--31}.
\newblock


\bibitem[\protect\citeauthoryear{AI}{AI}{2019}]%
        {ai2019high}
\bibfield{author}{\bibinfo{person}{HLEG AI}.} \bibinfo{year}{2019}\natexlab{}.
\newblock \bibinfo{title}{High-level expert group on artificial intelligence}.
\newblock , \bibinfo{numpages}{6}~pages.
\newblock


\bibitem[\protect\citeauthoryear{Akers and Jennings}{Akers and
  Jennings}{2015}]%
        {akers2015social}
\bibfield{author}{\bibinfo{person}{Ronald~L Akers} {and}
  \bibinfo{person}{Wesley~G Jennings}.} \bibinfo{year}{2015}\natexlab{}.
\newblock \showarticletitle{Social learning theory}.
\newblock \bibinfo{journal}{\emph{The handbook of criminological theory}}
  \bibinfo{volume}{4} (\bibinfo{year}{2015}), \bibinfo{pages}{230--240}.
\newblock


\bibitem[\protect\citeauthoryear{Akkaya, Andrychowicz, Chociej, Litwin, McGrew,
  Petron, Paino, Plappert, Powell, Ribas, et~al\mbox{.}}{Akkaya
  et~al\mbox{.}}{2019}]%
        {akkaya2019solving}
\bibfield{author}{\bibinfo{person}{Ilge Akkaya}, \bibinfo{person}{Marcin
  Andrychowicz}, \bibinfo{person}{Maciek Chociej}, \bibinfo{person}{Mateusz
  Litwin}, \bibinfo{person}{Bob McGrew}, \bibinfo{person}{Arthur Petron},
  \bibinfo{person}{Alex Paino}, \bibinfo{person}{Matthias Plappert},
  \bibinfo{person}{Glenn Powell}, \bibinfo{person}{Raphael Ribas},
  {et~al\mbox{.}}} \bibinfo{year}{2019}\natexlab{}.
\newblock \showarticletitle{Solving rubik's cube with a robot hand}.
\newblock \bibinfo{journal}{\emph{arXiv preprint arXiv:1910.07113}}
  (\bibinfo{year}{2019}).
\newblock


\bibitem[\protect\citeauthoryear{Altman}{Altman}{1998}]%
        {altman1998constrained}
\bibfield{author}{\bibinfo{person}{Eitan Altman}.}
  \bibinfo{year}{1998}\natexlab{}.
\newblock \showarticletitle{Constrained Markov decision processes with total
  cost criteria: Lagrangian approach and dual linear program}.
\newblock \bibinfo{journal}{\emph{Mathematical methods of operations research}}
  \bibinfo{volume}{48}, \bibinfo{number}{3} (\bibinfo{year}{1998}),
  \bibinfo{pages}{387--417}.
\newblock


\bibitem[\protect\citeauthoryear{Ammar, Eaton, Ruvolo, and Taylor}{Ammar
  et~al\mbox{.}}{2014}]%
        {ammar2014online}
\bibfield{author}{\bibinfo{person}{Haitham~Bou Ammar}, \bibinfo{person}{Eric
  Eaton}, \bibinfo{person}{Paul Ruvolo}, {and} \bibinfo{person}{Matthew
  Taylor}.} \bibinfo{year}{2014}\natexlab{}.
\newblock \showarticletitle{Online multi-task learning for policy gradient
  methods}. In \bibinfo{booktitle}{\emph{International conference on machine
  learning}}. PMLR, \bibinfo{pages}{1206--1214}.
\newblock


\bibitem[\protect\citeauthoryear{Andrychowicz, Wolski, Ray, Schneider, Fong,
  Welinder, McGrew, Tobin, Pieter~Abbeel, and Zaremba}{Andrychowicz
  et~al\mbox{.}}{2017}]%
        {andrychowicz2017hindsight}
\bibfield{author}{\bibinfo{person}{Marcin Andrychowicz}, \bibinfo{person}{Filip
  Wolski}, \bibinfo{person}{Alex Ray}, \bibinfo{person}{Jonas Schneider},
  \bibinfo{person}{Rachel Fong}, \bibinfo{person}{Peter Welinder},
  \bibinfo{person}{Bob McGrew}, \bibinfo{person}{Josh Tobin},
  \bibinfo{person}{OpenAI Pieter~Abbeel}, {and} \bibinfo{person}{Wojciech
  Zaremba}.} \bibinfo{year}{2017}\natexlab{}.
\newblock \showarticletitle{Hindsight experience replay}.
\newblock \bibinfo{journal}{\emph{Advances in neural information processing
  systems}}  \bibinfo{volume}{30} (\bibinfo{year}{2017}).
\newblock


\bibitem[\protect\citeauthoryear{Antonova, Cruciani, Smith, and
  Kragic}{Antonova et~al\mbox{.}}{2017}]%
        {antonova2017reinforcement}
\bibfield{author}{\bibinfo{person}{Rika Antonova}, \bibinfo{person}{Silvia
  Cruciani}, \bibinfo{person}{Christian Smith}, {and} \bibinfo{person}{Danica
  Kragic}.} \bibinfo{year}{2017}\natexlab{}.
\newblock \showarticletitle{Reinforcement learning for pivoting task}.
\newblock \bibinfo{journal}{\emph{arXiv preprint arXiv:1703.00472}}
  (\bibinfo{year}{2017}).
\newblock


\bibitem[\protect\citeauthoryear{Arief, Huang, Kumar, Bai, He, Ding, Lam, and
  Zhao}{Arief et~al\mbox{.}}{2021}]%
        {arief2021deep}
\bibfield{author}{\bibinfo{person}{Mansur Arief}, \bibinfo{person}{Zhiyuan
  Huang}, \bibinfo{person}{Guru Koushik~Senthil Kumar}, \bibinfo{person}{Yuanlu
  Bai}, \bibinfo{person}{Shengyi He}, \bibinfo{person}{Wenhao Ding},
  \bibinfo{person}{Henry Lam}, {and} \bibinfo{person}{Ding Zhao}.}
  \bibinfo{year}{2021}\natexlab{}.
\newblock \showarticletitle{Deep probabilistic accelerated evaluation: A robust
  certifiable rare-event simulation methodology for black-box safety-critical
  systems}. In \bibinfo{booktitle}{\emph{International Conference on Artificial
  Intelligence and Statistics}}. PMLR, \bibinfo{pages}{595--603}.
\newblock


\bibitem[\protect\citeauthoryear{Arndt, Hazara, Ghadirzadeh, and Kyrki}{Arndt
  et~al\mbox{.}}{2020}]%
        {arndt2020meta}
\bibfield{author}{\bibinfo{person}{Karol Arndt}, \bibinfo{person}{Murtaza
  Hazara}, \bibinfo{person}{Ali Ghadirzadeh}, {and} \bibinfo{person}{Ville
  Kyrki}.} \bibinfo{year}{2020}\natexlab{}.
\newblock \showarticletitle{Meta reinforcement learning for sim-to-real domain
  adaptation}. In \bibinfo{booktitle}{\emph{2020 IEEE International Conference
  on Robotics and Automation (ICRA)}}. IEEE, \bibinfo{pages}{2725--2731}.
\newblock


\bibitem[\protect\citeauthoryear{Arora and Doshi}{Arora and Doshi}{2021}]%
        {arora2021survey}
\bibfield{author}{\bibinfo{person}{Saurabh Arora} {and}
  \bibinfo{person}{Prashant Doshi}.} \bibinfo{year}{2021}\natexlab{}.
\newblock \showarticletitle{A survey of inverse reinforcement learning:
  Challenges, methods and progress}.
\newblock \bibinfo{journal}{\emph{Artificial Intelligence}}
  (\bibinfo{year}{2021}), \bibinfo{pages}{103500}.
\newblock


\bibitem[\protect\citeauthoryear{As, Usmanova, Curi, and Krause}{As
  et~al\mbox{.}}{2022}]%
        {as2022constrained}
\bibfield{author}{\bibinfo{person}{Yarden As}, \bibinfo{person}{Ilnura
  Usmanova}, \bibinfo{person}{Sebastian Curi}, {and} \bibinfo{person}{Andreas
  Krause}.} \bibinfo{year}{2022}\natexlab{}.
\newblock \showarticletitle{Constrained Policy Optimization via Bayesian World
  Models}.
\newblock \bibinfo{journal}{\emph{arXiv preprint arXiv:2201.09802}}
  (\bibinfo{year}{2022}).
\newblock


\bibitem[\protect\citeauthoryear{Baker, Tenenbaum, and Saxe}{Baker
  et~al\mbox{.}}{2007}]%
        {baker2007goal}
\bibfield{author}{\bibinfo{person}{Chris~L Baker}, \bibinfo{person}{Joshua~B
  Tenenbaum}, {and} \bibinfo{person}{Rebecca~R Saxe}.}
  \bibinfo{year}{2007}\natexlab{}.
\newblock \showarticletitle{Goal inference as inverse planning}. In
  \bibinfo{booktitle}{\emph{Proceedings of the Annual Meeting of the Cognitive
  Science Society}}, Vol.~\bibinfo{volume}{29}.
\newblock


\bibitem[\protect\citeauthoryear{Baluja and Fischer}{Baluja and
  Fischer}{2018}]%
        {baluja2018learning}
\bibfield{author}{\bibinfo{person}{Shumeet Baluja} {and} \bibinfo{person}{Ian
  Fischer}.} \bibinfo{year}{2018}\natexlab{}.
\newblock \showarticletitle{Learning to attack: Adversarial transformation
  networks}. In \bibinfo{booktitle}{\emph{Thirty-second aaai conference on
  artificial intelligence}}.
\newblock


\bibitem[\protect\citeauthoryear{Behzadan and Munir}{Behzadan and
  Munir}{2017a}]%
        {behzadan2017vulnerability}
\bibfield{author}{\bibinfo{person}{Vahid Behzadan} {and}
  \bibinfo{person}{Arslan Munir}.} \bibinfo{year}{2017}\natexlab{a}.
\newblock \showarticletitle{Vulnerability of deep reinforcement learning to
  policy induction attacks}. In \bibinfo{booktitle}{\emph{International
  Conference on Machine Learning and Data Mining in Pattern Recognition}}.
  Springer, \bibinfo{pages}{262--275}.
\newblock


\bibitem[\protect\citeauthoryear{Behzadan and Munir}{Behzadan and
  Munir}{2017b}]%
        {behzadan2017whatever}
\bibfield{author}{\bibinfo{person}{Vahid Behzadan} {and}
  \bibinfo{person}{Arslan Munir}.} \bibinfo{year}{2017}\natexlab{b}.
\newblock \showarticletitle{Whatever does not kill deep reinforcement learning,
  makes it stronger}.
\newblock \bibinfo{journal}{\emph{arXiv preprint arXiv:1712.09344}}
  (\bibinfo{year}{2017}).
\newblock


\bibitem[\protect\citeauthoryear{Bellemare, Naddaf, Veness, and
  Bowling}{Bellemare et~al\mbox{.}}{2013}]%
        {bellemare2013arcade}
\bibfield{author}{\bibinfo{person}{Marc~G Bellemare}, \bibinfo{person}{Yavar
  Naddaf}, \bibinfo{person}{Joel Veness}, {and} \bibinfo{person}{Michael
  Bowling}.} \bibinfo{year}{2013}\natexlab{}.
\newblock \showarticletitle{The arcade learning environment: An evaluation
  platform for general agents}.
\newblock \bibinfo{journal}{\emph{Journal of Artificial Intelligence Research}}
   \bibinfo{volume}{47} (\bibinfo{year}{2013}), \bibinfo{pages}{253--279}.
\newblock


\bibitem[\protect\citeauthoryear{Berkenkamp, Turchetta, Schoellig, and
  Krause}{Berkenkamp et~al\mbox{.}}{2017}]%
        {berkenkamp2017safe}
\bibfield{author}{\bibinfo{person}{Felix Berkenkamp}, \bibinfo{person}{Matteo
  Turchetta}, \bibinfo{person}{Angela~P Schoellig}, {and}
  \bibinfo{person}{Andreas Krause}.} \bibinfo{year}{2017}\natexlab{}.
\newblock \showarticletitle{Safe model-based reinforcement learning with
  stability guarantees}.
\newblock \bibinfo{journal}{\emph{arXiv preprint arXiv:1705.08551}}
  (\bibinfo{year}{2017}).
\newblock


\bibitem[\protect\citeauthoryear{Bhatnagar and Lakshmanan}{Bhatnagar and
  Lakshmanan}{2012}]%
        {bhatnagar2012online}
\bibfield{author}{\bibinfo{person}{Shalabh Bhatnagar} {and} \bibinfo{person}{K
  Lakshmanan}.} \bibinfo{year}{2012}\natexlab{}.
\newblock \showarticletitle{An online actor--critic algorithm with function
  approximation for constrained markov decision processes}.
\newblock \bibinfo{journal}{\emph{Journal of Optimization Theory and
  Applications}} \bibinfo{volume}{153}, \bibinfo{number}{3}
  (\bibinfo{year}{2012}), \bibinfo{pages}{688--708}.
\newblock


\bibitem[\protect\citeauthoryear{Bica, Jarrett, and van~der Schaar}{Bica
  et~al\mbox{.}}{2021}]%
        {bica2021invariant}
\bibfield{author}{\bibinfo{person}{Ioana Bica}, \bibinfo{person}{Daniel
  Jarrett}, {and} \bibinfo{person}{Mihaela van~der Schaar}.}
  \bibinfo{year}{2021}\natexlab{}.
\newblock \showarticletitle{Invariant Causal Imitation Learning for
  Generalizable Policies}.
\newblock \bibinfo{journal}{\emph{Advances in Neural Information Processing
  Systems}}  \bibinfo{volume}{34} (\bibinfo{year}{2021}),
  \bibinfo{pages}{3952--3964}.
\newblock


\bibitem[\protect\citeauthoryear{Borkar}{Borkar}{2005}]%
        {borkar2005actor}
\bibfield{author}{\bibinfo{person}{Vivek~S Borkar}.}
  \bibinfo{year}{2005}\natexlab{}.
\newblock \showarticletitle{An actor-critic algorithm for constrained Markov
  decision processes}.
\newblock \bibinfo{journal}{\emph{Systems \& control letters}}
  \bibinfo{volume}{54}, \bibinfo{number}{3} (\bibinfo{year}{2005}),
  \bibinfo{pages}{207--213}.
\newblock


\bibitem[\protect\citeauthoryear{Brantley, Dudik, Lykouris, Miryoosefi,
  Simchowitz, Slivkins, and Sun}{Brantley et~al\mbox{.}}{2020}]%
        {brantley2020constrained}
\bibfield{author}{\bibinfo{person}{Kiant{\'e} Brantley},
  \bibinfo{person}{Miroslav Dudik}, \bibinfo{person}{Thodoris Lykouris},
  \bibinfo{person}{Sobhan Miryoosefi}, \bibinfo{person}{Max Simchowitz},
  \bibinfo{person}{Aleksandrs Slivkins}, {and} \bibinfo{person}{Wen Sun}.}
  \bibinfo{year}{2020}\natexlab{}.
\newblock \showarticletitle{Constrained episodic reinforcement learning in
  concave-convex and knapsack settings}.
\newblock \bibinfo{journal}{\emph{arXiv preprint arXiv:2006.05051}}
  (\bibinfo{year}{2020}).
\newblock


\bibitem[\protect\citeauthoryear{Brockman, Cheung, Pettersson, Schneider,
  Schulman, Tang, and Zaremba}{Brockman et~al\mbox{.}}{2016}]%
        {brockman2016openai}
\bibfield{author}{\bibinfo{person}{Greg Brockman}, \bibinfo{person}{Vicki
  Cheung}, \bibinfo{person}{Ludwig Pettersson}, \bibinfo{person}{Jonas
  Schneider}, \bibinfo{person}{John Schulman}, \bibinfo{person}{Jie Tang},
  {and} \bibinfo{person}{Wojciech Zaremba}.} \bibinfo{year}{2016}\natexlab{}.
\newblock \showarticletitle{Openai gym}.
\newblock \bibinfo{journal}{\emph{arXiv preprint arXiv:1606.01540}}
  (\bibinfo{year}{2016}).
\newblock


\bibitem[\protect\citeauthoryear{Brunke, Greeff, Hall, Yuan, Zhou, Panerati,
  and Schoellig}{Brunke et~al\mbox{.}}{2022}]%
        {brunke2022safe}
\bibfield{author}{\bibinfo{person}{Lukas Brunke}, \bibinfo{person}{Melissa
  Greeff}, \bibinfo{person}{Adam~W Hall}, \bibinfo{person}{Zhaocong Yuan},
  \bibinfo{person}{Siqi Zhou}, \bibinfo{person}{Jacopo Panerati}, {and}
  \bibinfo{person}{Angela~P Schoellig}.} \bibinfo{year}{2022}\natexlab{}.
\newblock \showarticletitle{Safe learning in robotics: From learning-based
  control to safe reinforcement learning}.
\newblock \bibinfo{journal}{\emph{Annual Review of Control, Robotics, and
  Autonomous Systems}}  \bibinfo{volume}{5} (\bibinfo{year}{2022}),
  \bibinfo{pages}{411--444}.
\newblock


\bibitem[\protect\citeauthoryear{Carlini and Wagner}{Carlini and
  Wagner}{2017}]%
        {carlini2017towards}
\bibfield{author}{\bibinfo{person}{Nicholas Carlini} {and}
  \bibinfo{person}{David Wagner}.} \bibinfo{year}{2017}\natexlab{}.
\newblock \showarticletitle{Towards evaluating the robustness of neural
  networks}. In \bibinfo{booktitle}{\emph{2017 ieee symposium on security and
  privacy (sp)}}. IEEE, \bibinfo{pages}{39--57}.
\newblock


\bibitem[\protect\citeauthoryear{Chen, Liu, Zhu, Xu, Ding, and Zhao}{Chen
  et~al\mbox{.}}{2021}]%
        {chen2021context}
\bibfield{author}{\bibinfo{person}{Baiming Chen}, \bibinfo{person}{Zuxin Liu},
  \bibinfo{person}{Jiacheng Zhu}, \bibinfo{person}{Mengdi Xu},
  \bibinfo{person}{Wenhao Ding}, {and} \bibinfo{person}{Ding Zhao}.}
  \bibinfo{year}{2021}\natexlab{}.
\newblock \showarticletitle{Context-Aware Safe Reinforcement Learning for
  Non-Stationary Environments}.
\newblock \bibinfo{journal}{\emph{arXiv preprint arXiv:2101.00531}}
  (\bibinfo{year}{2021}).
\newblock


\bibitem[\protect\citeauthoryear{Chen, Liu, Xiang, Niu, Tong, and Han}{Chen
  et~al\mbox{.}}{2019a}]%
        {chen2019adversarial}
\bibfield{author}{\bibinfo{person}{Tong Chen}, \bibinfo{person}{Jiqiang Liu},
  \bibinfo{person}{Yingxiao Xiang}, \bibinfo{person}{Wenjia Niu},
  \bibinfo{person}{Endong Tong}, {and} \bibinfo{person}{Zhen Han}.}
  \bibinfo{year}{2019}\natexlab{a}.
\newblock \showarticletitle{Adversarial attack and defense in reinforcement
  learning-from AI security view}.
\newblock \bibinfo{journal}{\emph{Cybersecurity}} \bibinfo{volume}{2},
  \bibinfo{number}{1} (\bibinfo{year}{2019}), \bibinfo{pages}{1--22}.
\newblock


\bibitem[\protect\citeauthoryear{Chen, Yu, and Haskell}{Chen
  et~al\mbox{.}}{2019b}]%
        {chen2019distributionally}
\bibfield{author}{\bibinfo{person}{Zhi Chen}, \bibinfo{person}{Pengqian Yu},
  {and} \bibinfo{person}{William~B Haskell}.} \bibinfo{year}{2019}\natexlab{b}.
\newblock \showarticletitle{Distributionally robust optimization for sequential
  decision-making}.
\newblock \bibinfo{journal}{\emph{Optimization}} \bibinfo{volume}{68},
  \bibinfo{number}{12} (\bibinfo{year}{2019}), \bibinfo{pages}{2397--2426}.
\newblock


\bibitem[\protect\citeauthoryear{Cheng, Orosz, Murray, and Burdick}{Cheng
  et~al\mbox{.}}{2019}]%
        {cheng2019end}
\bibfield{author}{\bibinfo{person}{Richard Cheng}, \bibinfo{person}{G{\'a}bor
  Orosz}, \bibinfo{person}{Richard~M Murray}, {and} \bibinfo{person}{Joel~W
  Burdick}.} \bibinfo{year}{2019}\natexlab{}.
\newblock \showarticletitle{End-to-end safe reinforcement learning through
  barrier functions for safety-critical continuous control tasks}. In
  \bibinfo{booktitle}{\emph{Proceedings of the AAAI Conference on Artificial
  Intelligence}}, Vol.~\bibinfo{volume}{33}. \bibinfo{pages}{3387--3395}.
\newblock


\bibitem[\protect\citeauthoryear{Chow, Ghavamzadeh, Janson, and Pavone}{Chow
  et~al\mbox{.}}{2017}]%
        {chow2017risk}
\bibfield{author}{\bibinfo{person}{Yinlam Chow}, \bibinfo{person}{Mohammad
  Ghavamzadeh}, \bibinfo{person}{Lucas Janson}, {and} \bibinfo{person}{Marco
  Pavone}.} \bibinfo{year}{2017}\natexlab{}.
\newblock \showarticletitle{Risk-constrained reinforcement learning with
  percentile risk criteria}.
\newblock \bibinfo{journal}{\emph{The Journal of Machine Learning Research}}
  \bibinfo{volume}{18}, \bibinfo{number}{1} (\bibinfo{year}{2017}),
  \bibinfo{pages}{6070--6120}.
\newblock


\bibitem[\protect\citeauthoryear{Chow, Nachum, Faust, Duenez-Guzman, and
  Ghavamzadeh}{Chow et~al\mbox{.}}{2019}]%
        {chow2019lyapunov}
\bibfield{author}{\bibinfo{person}{Yinlam Chow}, \bibinfo{person}{Ofir Nachum},
  \bibinfo{person}{Aleksandra Faust}, \bibinfo{person}{Edgar Duenez-Guzman},
  {and} \bibinfo{person}{Mohammad Ghavamzadeh}.}
  \bibinfo{year}{2019}\natexlab{}.
\newblock \showarticletitle{Lyapunov-based safe policy optimization for
  continuous control}.
\newblock \bibinfo{journal}{\emph{arXiv preprint arXiv:1901.10031}}
  (\bibinfo{year}{2019}).
\newblock


\bibitem[\protect\citeauthoryear{Cobbe, Hesse, Hilton, and Schulman}{Cobbe
  et~al\mbox{.}}{2020}]%
        {cobbe2020leveraging}
\bibfield{author}{\bibinfo{person}{Karl Cobbe}, \bibinfo{person}{Chris Hesse},
  \bibinfo{person}{Jacob Hilton}, {and} \bibinfo{person}{John Schulman}.}
  \bibinfo{year}{2020}\natexlab{}.
\newblock \showarticletitle{Leveraging procedural generation to benchmark
  reinforcement learning}. In \bibinfo{booktitle}{\emph{International
  conference on machine learning}}. PMLR, \bibinfo{pages}{2048--2056}.
\newblock


\bibitem[\protect\citeauthoryear{Cui, Qu, Nosouhi, Yu, Niu, and Xie}{Cui
  et~al\mbox{.}}{2019}]%
        {cui2019improving}
\bibfield{author}{\bibinfo{person}{Lei Cui}, \bibinfo{person}{Youyang Qu},
  \bibinfo{person}{Mohammad~Reza Nosouhi}, \bibinfo{person}{Shui Yu},
  \bibinfo{person}{Jian-Wei Niu}, {and} \bibinfo{person}{Gang Xie}.}
  \bibinfo{year}{2019}\natexlab{}.
\newblock \showarticletitle{Improving data utility through game theory in
  personalized differential privacy}.
\newblock \bibinfo{journal}{\emph{Journal of Computer Science and Technology}}
  \bibinfo{volume}{34}, \bibinfo{number}{2} (\bibinfo{year}{2019}),
  \bibinfo{pages}{272--286}.
\newblock


\bibitem[\protect\citeauthoryear{Dalal, Dvijotham, Vecerik, Hester, Paduraru,
  and Tassa}{Dalal et~al\mbox{.}}{2018}]%
        {dalal2018safe}
\bibfield{author}{\bibinfo{person}{Gal Dalal}, \bibinfo{person}{Krishnamurthy
  Dvijotham}, \bibinfo{person}{Matej Vecerik}, \bibinfo{person}{Todd Hester},
  \bibinfo{person}{Cosmin Paduraru}, {and} \bibinfo{person}{Yuval Tassa}.}
  \bibinfo{year}{2018}\natexlab{}.
\newblock \showarticletitle{Safe exploration in continuous action spaces}.
\newblock \bibinfo{journal}{\emph{arXiv preprint arXiv:1801.08757}}
  (\bibinfo{year}{2018}).
\newblock


\bibitem[\protect\citeauthoryear{Dennis, Jaques, Vinitsky, Bayen, Russell,
  Critch, and Levine}{Dennis et~al\mbox{.}}{2020}]%
        {dennis2020emergent}
\bibfield{author}{\bibinfo{person}{Michael Dennis}, \bibinfo{person}{Natasha
  Jaques}, \bibinfo{person}{Eugene Vinitsky}, \bibinfo{person}{Alexandre
  Bayen}, \bibinfo{person}{Stuart Russell}, \bibinfo{person}{Andrew Critch},
  {and} \bibinfo{person}{Sergey Levine}.} \bibinfo{year}{2020}\natexlab{}.
\newblock \showarticletitle{Emergent complexity and zero-shot transfer via
  unsupervised environment design}.
\newblock \bibinfo{journal}{\emph{Advances in neural information processing
  systems}}  \bibinfo{volume}{33} (\bibinfo{year}{2020}),
  \bibinfo{pages}{13049--13061}.
\newblock


\bibitem[\protect\citeauthoryear{Derman and Mannor}{Derman and Mannor}{2020}]%
        {derman2020distributional}
\bibfield{author}{\bibinfo{person}{Esther Derman} {and} \bibinfo{person}{Shie
  Mannor}.} \bibinfo{year}{2020}\natexlab{}.
\newblock \showarticletitle{Distributional robustness and regularization in
  reinforcement learning}.
\newblock \bibinfo{journal}{\emph{arXiv preprint arXiv:2003.02894}}
  (\bibinfo{year}{2020}).
\newblock


\bibitem[\protect\citeauthoryear{Ding, Wei, Yang, Wang, and Jovanovic}{Ding
  et~al\mbox{.}}{2021}]%
        {ding2021provably}
\bibfield{author}{\bibinfo{person}{Dongsheng Ding}, \bibinfo{person}{Xiaohan
  Wei}, \bibinfo{person}{Zhuoran Yang}, \bibinfo{person}{Zhaoran Wang}, {and}
  \bibinfo{person}{Mihailo Jovanovic}.} \bibinfo{year}{2021}\natexlab{}.
\newblock \showarticletitle{Provably efficient safe exploration via primal-dual
  policy optimization}. In \bibinfo{booktitle}{\emph{International Conference
  on Artificial Intelligence and Statistics}}. PMLR,
  \bibinfo{pages}{3304--3312}.
\newblock


\bibitem[\protect\citeauthoryear{Ding, Lin, Li, and Zhao}{Ding
  et~al\mbox{.}}{2022a}]%
        {ding2022generalizing}
\bibfield{author}{\bibinfo{person}{Wenhao Ding}, \bibinfo{person}{Haohong Lin},
  \bibinfo{person}{Bo Li}, {and} \bibinfo{person}{Ding Zhao}.}
  \bibinfo{year}{2022}\natexlab{a}.
\newblock \showarticletitle{Generalizing Goal-Conditioned Reinforcement
  Learning with Variational Causal Reasoning}.
\newblock \bibinfo{journal}{\emph{arXiv preprint arXiv:2207.09081}}
  (\bibinfo{year}{2022}).
\newblock


\bibitem[\protect\citeauthoryear{Ding, Xu, Lin, Li, and Zhao}{Ding
  et~al\mbox{.}}{2022b}]%
        {ding2022survey}
\bibfield{author}{\bibinfo{person}{Wenhao Ding}, \bibinfo{person}{Chejian Xu},
  \bibinfo{person}{Haohong Lin}, \bibinfo{person}{Bo Li}, {and}
  \bibinfo{person}{Ding Zhao}.} \bibinfo{year}{2022}\natexlab{b}.
\newblock \showarticletitle{A Survey on Safety-critical Scenario Generation
  from Methodological Perspective}.
\newblock \bibinfo{journal}{\emph{arXiv preprint arXiv:2202.02215}}
  (\bibinfo{year}{2022}).
\newblock


\bibitem[\protect\citeauthoryear{Dosovitskiy, Ros, Codevilla, Lopez, and
  Koltun}{Dosovitskiy et~al\mbox{.}}{2017}]%
        {dosovitskiy2017carla}
\bibfield{author}{\bibinfo{person}{Alexey Dosovitskiy}, \bibinfo{person}{German
  Ros}, \bibinfo{person}{Felipe Codevilla}, \bibinfo{person}{Antonio Lopez},
  {and} \bibinfo{person}{Vladlen Koltun}.} \bibinfo{year}{2017}\natexlab{}.
\newblock \showarticletitle{CARLA: An open urban driving simulator}. In
  \bibinfo{booktitle}{\emph{Conference on robot learning}}. PMLR,
  \bibinfo{pages}{1--16}.
\newblock


\bibitem[\protect\citeauthoryear{Dulac-Arnold, Levine, Mankowitz, Li, Paduraru,
  Gowal, and Hester}{Dulac-Arnold et~al\mbox{.}}{2021}]%
        {dulac2021challenges}
\bibfield{author}{\bibinfo{person}{Gabriel Dulac-Arnold}, \bibinfo{person}{Nir
  Levine}, \bibinfo{person}{Daniel~J Mankowitz}, \bibinfo{person}{Jerry Li},
  \bibinfo{person}{Cosmin Paduraru}, \bibinfo{person}{Sven Gowal}, {and}
  \bibinfo{person}{Todd Hester}.} \bibinfo{year}{2021}\natexlab{}.
\newblock \showarticletitle{Challenges of real-world reinforcement learning:
  definitions, benchmarks and analysis}.
\newblock \bibinfo{journal}{\emph{Machine Learning}} \bibinfo{volume}{110},
  \bibinfo{number}{9} (\bibinfo{year}{2021}), \bibinfo{pages}{2419--2468}.
\newblock


\bibitem[\protect\citeauthoryear{Efroni, Mannor, and Pirotta}{Efroni
  et~al\mbox{.}}{2020}]%
        {efroni2020exploration}
\bibfield{author}{\bibinfo{person}{Yonathan Efroni}, \bibinfo{person}{Shie
  Mannor}, {and} \bibinfo{person}{Matteo Pirotta}.}
  \bibinfo{year}{2020}\natexlab{}.
\newblock \showarticletitle{Exploration-exploitation in constrained mdps}.
\newblock \bibinfo{journal}{\emph{arXiv preprint arXiv:2003.02189}}
  (\bibinfo{year}{2020}).
\newblock


\bibitem[\protect\citeauthoryear{Elavarasan and Vincent}{Elavarasan and
  Vincent}{2020}]%
        {9086620}
\bibfield{author}{\bibinfo{person}{Dhivya Elavarasan} {and}
  \bibinfo{person}{P.~M.~Durairaj Vincent}.} \bibinfo{year}{2020}\natexlab{}.
\newblock \showarticletitle{Crop Yield Prediction Using Deep Reinforcement
  Learning Model for Sustainable Agrarian Applications}.
\newblock \bibinfo{journal}{\emph{IEEE Access}}  \bibinfo{volume}{8}
  (\bibinfo{year}{2020}), \bibinfo{pages}{86886--86901}.
\newblock
\urldef\tempurl%
\url{https://doi.org/10.1109/ACCESS.2020.2992480}
\showDOI{\tempurl}


\bibitem[\protect\citeauthoryear{Erez, Tassa, and Todorov}{Erez
  et~al\mbox{.}}{2012}]%
        {erez2012infinite}
\bibfield{author}{\bibinfo{person}{Tom Erez}, \bibinfo{person}{Yuval Tassa},
  {and} \bibinfo{person}{Emanuel Todorov}.} \bibinfo{year}{2012}\natexlab{}.
\newblock \showarticletitle{Infinite-horizon model predictive control for
  periodic tasks with contacts}.
\newblock \bibinfo{journal}{\emph{Robotics: Science and systems VII}}
  \bibinfo{volume}{73} (\bibinfo{year}{2012}).
\newblock


\bibitem[\protect\citeauthoryear{Everett, L{\"u}tjens, and How}{Everett
  et~al\mbox{.}}{2021}]%
        {everett2021certifiable}
\bibfield{author}{\bibinfo{person}{Michael Everett}, \bibinfo{person}{Bj{\"o}rn
  L{\"u}tjens}, {and} \bibinfo{person}{Jonathan~P How}.}
  \bibinfo{year}{2021}\natexlab{}.
\newblock \showarticletitle{Certifiable Robustness to Adversarial State
  Uncertainty in Deep Reinforcement Learning}.
\newblock \bibinfo{journal}{\emph{IEEE Transactions on Neural Networks and
  Learning Systems}} (\bibinfo{year}{2021}).
\newblock


\bibitem[\protect\citeauthoryear{Everitt, Krakovna, Orseau, Hutter, and
  Legg}{Everitt et~al\mbox{.}}{2017}]%
        {everitt2017reinforcement}
\bibfield{author}{\bibinfo{person}{Tom Everitt}, \bibinfo{person}{Victoria
  Krakovna}, \bibinfo{person}{Laurent Orseau}, \bibinfo{person}{Marcus Hutter},
  {and} \bibinfo{person}{Shane Legg}.} \bibinfo{year}{2017}\natexlab{}.
\newblock \showarticletitle{Reinforcement learning with a corrupted reward
  channel}.
\newblock \bibinfo{journal}{\emph{arXiv preprint arXiv:1705.08417}}
  (\bibinfo{year}{2017}).
\newblock


\bibitem[\protect\citeauthoryear{Finn, Abbeel, and Levine}{Finn
  et~al\mbox{.}}{2017}]%
        {finn2017model}
\bibfield{author}{\bibinfo{person}{Chelsea Finn}, \bibinfo{person}{Pieter
  Abbeel}, {and} \bibinfo{person}{Sergey Levine}.}
  \bibinfo{year}{2017}\natexlab{}.
\newblock \showarticletitle{Model-agnostic meta-learning for fast adaptation of
  deep networks}. In \bibinfo{booktitle}{\emph{International Conference on
  Machine Learning}}. PMLR, \bibinfo{pages}{1126--1135}.
\newblock


\bibitem[\protect\citeauthoryear{Fu, Luo, and Levine}{Fu et~al\mbox{.}}{2017}]%
        {fu2017learning}
\bibfield{author}{\bibinfo{person}{Justin Fu}, \bibinfo{person}{Katie Luo},
  {and} \bibinfo{person}{Sergey Levine}.} \bibinfo{year}{2017}\natexlab{}.
\newblock \showarticletitle{Learning robust rewards with adversarial inverse
  reinforcement learning}.
\newblock \bibinfo{journal}{\emph{arXiv preprint arXiv:1710.11248}}
  (\bibinfo{year}{2017}).
\newblock


\bibitem[\protect\citeauthoryear{Gallego, Naveiro, and Insua}{Gallego
  et~al\mbox{.}}{2019}]%
        {gallego2019reinforcement}
\bibfield{author}{\bibinfo{person}{Victor Gallego}, \bibinfo{person}{Roi
  Naveiro}, {and} \bibinfo{person}{David~Rios Insua}.}
  \bibinfo{year}{2019}\natexlab{}.
\newblock \showarticletitle{Reinforcement learning under threats}. In
  \bibinfo{booktitle}{\emph{Proceedings of the AAAI Conference on Artificial
  Intelligence}}, Vol.~\bibinfo{volume}{33}. \bibinfo{pages}{9939--9940}.
\newblock


\bibitem[\protect\citeauthoryear{Gao, Sibirtseva, Castellano, and Kragic}{Gao
  et~al\mbox{.}}{2019}]%
        {gao2019fast}
\bibfield{author}{\bibinfo{person}{Yuan Gao}, \bibinfo{person}{Elena
  Sibirtseva}, \bibinfo{person}{Ginevra Castellano}, {and}
  \bibinfo{person}{Danica Kragic}.} \bibinfo{year}{2019}\natexlab{}.
\newblock \showarticletitle{Fast adaptation with meta-reinforcement learning
  for trust modelling in human-robot interaction}. In
  \bibinfo{booktitle}{\emph{2019 IEEE/RSJ International Conference on
  Intelligent Robots and Systems (IROS)}}. IEEE, \bibinfo{pages}{305--312}.
\newblock


\bibitem[\protect\citeauthoryear{Garc{\i}a and Fern{\'a}ndez}{Garc{\i}a and
  Fern{\'a}ndez}{2015}]%
        {garcia2015comprehensive}
\bibfield{author}{\bibinfo{person}{Javier Garc{\i}a} {and}
  \bibinfo{person}{Fernando Fern{\'a}ndez}.} \bibinfo{year}{2015}\natexlab{}.
\newblock \showarticletitle{A comprehensive survey on safe reinforcement
  learning}.
\newblock \bibinfo{journal}{\emph{Journal of Machine Learning Research}}
  \bibinfo{volume}{16}, \bibinfo{number}{1} (\bibinfo{year}{2015}),
  \bibinfo{pages}{1437--1480}.
\newblock


\bibitem[\protect\citeauthoryear{Gj{\ae}rum, Str{\"u}mke, Alsos, and
  Lekkas}{Gj{\ae}rum et~al\mbox{.}}{2021}]%
        {gjaerum2021explaining}
\bibfield{author}{\bibinfo{person}{Vilde~B Gj{\ae}rum}, \bibinfo{person}{Inga
  Str{\"u}mke}, \bibinfo{person}{Ole~Andreas Alsos}, {and}
  \bibinfo{person}{Anastasios~M Lekkas}.} \bibinfo{year}{2021}\natexlab{}.
\newblock \showarticletitle{Explaining a Deep Reinforcement Learning Docking
  Agent Using Linear Model Trees with User Adapted Visualization}.
\newblock \bibinfo{journal}{\emph{Journal of Marine Science and Engineering}}
  \bibinfo{volume}{9}, \bibinfo{number}{11} (\bibinfo{year}{2021}),
  \bibinfo{pages}{1178}.
\newblock


\bibitem[\protect\citeauthoryear{Gleave, Dennis, Wild, Kant, Levine, and
  Russell}{Gleave et~al\mbox{.}}{2019}]%
        {gleave2019adversarial}
\bibfield{author}{\bibinfo{person}{Adam Gleave}, \bibinfo{person}{Michael
  Dennis}, \bibinfo{person}{Cody Wild}, \bibinfo{person}{Neel Kant},
  \bibinfo{person}{Sergey Levine}, {and} \bibinfo{person}{Stuart Russell}.}
  \bibinfo{year}{2019}\natexlab{}.
\newblock \showarticletitle{Adversarial policies: Attacking deep reinforcement
  learning}.
\newblock \bibinfo{journal}{\emph{arXiv preprint arXiv:1905.10615}}
  (\bibinfo{year}{2019}).
\newblock


\bibitem[\protect\citeauthoryear{Golub, Chase, and Yu}{Golub
  et~al\mbox{.}}{2013}]%
        {golub2013learning}
\bibfield{author}{\bibinfo{person}{Matthew Golub}, \bibinfo{person}{Steven
  Chase}, {and} \bibinfo{person}{Byron Yu}.} \bibinfo{year}{2013}\natexlab{}.
\newblock \showarticletitle{Learning an internal dynamics model from control
  demonstration}. In \bibinfo{booktitle}{\emph{International Conference on
  Machine Learning}}. PMLR, \bibinfo{pages}{606--614}.
\newblock


\bibitem[\protect\citeauthoryear{Goodfellow, Shlens, and Szegedy}{Goodfellow
  et~al\mbox{.}}{2014}]%
        {goodfellow2014explaining}
\bibfield{author}{\bibinfo{person}{Ian~J Goodfellow}, \bibinfo{person}{Jonathon
  Shlens}, {and} \bibinfo{person}{Christian Szegedy}.}
  \bibinfo{year}{2014}\natexlab{}.
\newblock \showarticletitle{Explaining and harnessing adversarial examples}.
\newblock \bibinfo{journal}{\emph{arXiv preprint arXiv:1412.6572}}
  (\bibinfo{year}{2014}).
\newblock


\bibitem[\protect\citeauthoryear{Gowal, Dvijotham, Stanforth, Bunel, Qin,
  Uesato, Arandjelovic, Mann, and Kohli}{Gowal et~al\mbox{.}}{2018}]%
        {gowal2018effectiveness}
\bibfield{author}{\bibinfo{person}{Sven Gowal}, \bibinfo{person}{Krishnamurthy
  Dvijotham}, \bibinfo{person}{Robert Stanforth}, \bibinfo{person}{Rudy Bunel},
  \bibinfo{person}{Chongli Qin}, \bibinfo{person}{Jonathan Uesato},
  \bibinfo{person}{Relja Arandjelovic}, \bibinfo{person}{Timothy Mann}, {and}
  \bibinfo{person}{Pushmeet Kohli}.} \bibinfo{year}{2018}\natexlab{}.
\newblock \showarticletitle{On the effectiveness of interval bound propagation
  for training verifiably robust models}.
\newblock \bibinfo{journal}{\emph{arXiv preprint arXiv:1810.12715}}
  (\bibinfo{year}{2018}).
\newblock


\bibitem[\protect\citeauthoryear{Goyal and Grand-Clement}{Goyal and
  Grand-Clement}{2022}]%
        {goyal2022robust}
\bibfield{author}{\bibinfo{person}{Vineet Goyal} {and} \bibinfo{person}{Julien
  Grand-Clement}.} \bibinfo{year}{2022}\natexlab{}.
\newblock \showarticletitle{Robust Markov Decision Processes: Beyond
  Rectangularity}.
\newblock \bibinfo{journal}{\emph{Mathematics of Operations Research}}
  (\bibinfo{year}{2022}).
\newblock


\bibitem[\protect\citeauthoryear{Grigsby and Qi}{Grigsby and Qi}{2020}]%
        {grigsby2020measuring}
\bibfield{author}{\bibinfo{person}{Jake Grigsby} {and} \bibinfo{person}{Yanjun
  Qi}.} \bibinfo{year}{2020}\natexlab{}.
\newblock \showarticletitle{Measuring visual generalization in continuous
  control from pixels}.
\newblock \bibinfo{journal}{\emph{arXiv preprint arXiv:2010.06740}}
  (\bibinfo{year}{2020}).
\newblock


\bibitem[\protect\citeauthoryear{Gronauer}{Gronauer}{2022}]%
        {gronauer2022bullet}
\bibfield{author}{\bibinfo{person}{Sven Gronauer}.}
  \bibinfo{year}{2022}\natexlab{}.
\newblock \showarticletitle{BULLET-SAFETY-GYM: AFRAMEWORK FOR CONSTRAINED
  REINFORCEMENT LEARNING}.
\newblock  (\bibinfo{year}{2022}).
\newblock


\bibitem[\protect\citeauthoryear{Gu and Rigazio}{Gu and Rigazio}{2014}]%
        {gu2014towards}
\bibfield{author}{\bibinfo{person}{Shixiang Gu} {and} \bibinfo{person}{Luca
  Rigazio}.} \bibinfo{year}{2014}\natexlab{}.
\newblock \showarticletitle{Towards deep neural network architectures robust to
  adversarial examples}.
\newblock \bibinfo{journal}{\emph{arXiv preprint arXiv:1412.5068}}
  (\bibinfo{year}{2014}).
\newblock


\bibitem[\protect\citeauthoryear{Gu, Yang, Du, Chen, Walter, Wang, Yang, and
  Knoll}{Gu et~al\mbox{.}}{2022}]%
        {gu2022review}
\bibfield{author}{\bibinfo{person}{Shangding Gu}, \bibinfo{person}{Long Yang},
  \bibinfo{person}{Yali Du}, \bibinfo{person}{Guang Chen},
  \bibinfo{person}{Florian Walter}, \bibinfo{person}{Jun Wang},
  \bibinfo{person}{Yaodong Yang}, {and} \bibinfo{person}{Alois Knoll}.}
  \bibinfo{year}{2022}\natexlab{}.
\newblock \showarticletitle{A Review of Safe Reinforcement Learning: Methods,
  Theory and Applications}.
\newblock \bibinfo{journal}{\emph{arXiv preprint arXiv:2205.10330}}
  (\bibinfo{year}{2022}).
\newblock


\bibitem[\protect\citeauthoryear{Gupta, Savarese, Ganguli, and Fei-Fei}{Gupta
  et~al\mbox{.}}{2021}]%
        {Gupta2021-kq}
\bibfield{author}{\bibinfo{person}{Agrim Gupta}, \bibinfo{person}{Silvio
  Savarese}, \bibinfo{person}{Surya Ganguli}, {and} \bibinfo{person}{Li
  Fei-Fei}.} \bibinfo{year}{2021}\natexlab{}.
\newblock \showarticletitle{Embodied intelligence via learning and evolution}.
\newblock \bibinfo{journal}{\emph{Nature communications}} \bibinfo{volume}{12},
  \bibinfo{number}{1} (\bibinfo{date}{Oct.} \bibinfo{year}{2021}),
  \bibinfo{pages}{5721}.
\newblock


\bibitem[\protect\citeauthoryear{Ha}{Ha}{2019}]%
        {Ha2019-tr}
\bibfield{author}{\bibinfo{person}{David Ha}.} \bibinfo{year}{2019}\natexlab{}.
\newblock \showarticletitle{Reinforcement Learning for Improving Agent Design}.
\newblock \bibinfo{journal}{\emph{Artificial life}} \bibinfo{volume}{25},
  \bibinfo{number}{4} (\bibinfo{date}{Nov.} \bibinfo{year}{2019}),
  \bibinfo{pages}{352--365}.
\newblock


\bibitem[\protect\citeauthoryear{Hallak, Di~Castro, and Mannor}{Hallak
  et~al\mbox{.}}{2015}]%
        {hallak2015contextual}
\bibfield{author}{\bibinfo{person}{Assaf Hallak}, \bibinfo{person}{Dotan
  Di~Castro}, {and} \bibinfo{person}{Shie Mannor}.}
  \bibinfo{year}{2015}\natexlab{}.
\newblock \showarticletitle{Contextual markov decision processes}.
\newblock \bibinfo{journal}{\emph{arXiv preprint arXiv:1502.02259}}
  (\bibinfo{year}{2015}).
\newblock


\bibitem[\protect\citeauthoryear{Hansen and Wang}{Hansen and Wang}{2021}]%
        {hansen2021generalization}
\bibfield{author}{\bibinfo{person}{Nicklas Hansen} {and}
  \bibinfo{person}{Xiaolong Wang}.} \bibinfo{year}{2021}\natexlab{}.
\newblock \showarticletitle{Generalization in reinforcement learning by soft
  data augmentation}. In \bibinfo{booktitle}{\emph{2021 IEEE International
  Conference on Robotics and Automation (ICRA)}}. IEEE,
  \bibinfo{pages}{13611--13617}.
\newblock


\bibitem[\protect\citeauthoryear{Havens, Jiang, and Sarkar}{Havens
  et~al\mbox{.}}{2018}]%
        {havens2018online}
\bibfield{author}{\bibinfo{person}{Aaron Havens}, \bibinfo{person}{Zhanhong
  Jiang}, {and} \bibinfo{person}{Soumik Sarkar}.}
  \bibinfo{year}{2018}\natexlab{}.
\newblock \showarticletitle{Online robust policy learning in the presence of
  unknown adversaries}.
\newblock \bibinfo{journal}{\emph{Advances in neural information processing
  systems}}  \bibinfo{volume}{31} (\bibinfo{year}{2018}).
\newblock


\bibitem[\protect\citeauthoryear{Haydari and Yılmaz}{Haydari and
  Yılmaz}{2022}]%
        {9146378}
\bibfield{author}{\bibinfo{person}{Ammar Haydari} {and} \bibinfo{person}{Yasin
  Yılmaz}.} \bibinfo{year}{2022}\natexlab{}.
\newblock \showarticletitle{Deep Reinforcement Learning for Intelligent
  Transportation Systems: A Survey}.
\newblock \bibinfo{journal}{\emph{IEEE Transactions on Intelligent
  Transportation Systems}} \bibinfo{volume}{23}, \bibinfo{number}{1}
  (\bibinfo{year}{2022}), \bibinfo{pages}{11--32}.
\newblock
\urldef\tempurl%
\url{https://doi.org/10.1109/TITS.2020.3008612}
\showDOI{\tempurl}


\bibitem[\protect\citeauthoryear{Hein and Andriushchenko}{Hein and
  Andriushchenko}{2017}]%
        {hein2017formal}
\bibfield{author}{\bibinfo{person}{Matthias Hein} {and} \bibinfo{person}{Maksym
  Andriushchenko}.} \bibinfo{year}{2017}\natexlab{}.
\newblock \showarticletitle{Formal Guarantees on the Robustness of a Classifier
  against Adversarial Manipulation}. In \bibinfo{booktitle}{\emph{NIPS}}.
\newblock


\bibitem[\protect\citeauthoryear{Hendrycks, Mazeika, Zou, Patel, Zhu, Navarro,
  Song, Li, and Steinhardt}{Hendrycks et~al\mbox{.}}{2021}]%
        {hendrycks2021would}
\bibfield{author}{\bibinfo{person}{Dan Hendrycks}, \bibinfo{person}{Mantas
  Mazeika}, \bibinfo{person}{Andy Zou}, \bibinfo{person}{Sahil Patel},
  \bibinfo{person}{Christine Zhu}, \bibinfo{person}{Jesus Navarro},
  \bibinfo{person}{Dawn Song}, \bibinfo{person}{Bo Li}, {and}
  \bibinfo{person}{Jacob Steinhardt}.} \bibinfo{year}{2021}\natexlab{}.
\newblock \showarticletitle{What would jiminy cricket do? towards agents that
  behave morally}.
\newblock \bibinfo{journal}{\emph{arXiv preprint arXiv:2110.13136}}
  (\bibinfo{year}{2021}).
\newblock


\bibitem[\protect\citeauthoryear{Huang, Xu, Fang, and Zhao}{Huang
  et~al\mbox{.}}{2022}]%
        {huang2022robust}
\bibfield{author}{\bibinfo{person}{Peide Huang}, \bibinfo{person}{Mengdi Xu},
  \bibinfo{person}{Fei Fang}, {and} \bibinfo{person}{Ding Zhao}.}
  \bibinfo{year}{2022}\natexlab{}.
\newblock \showarticletitle{Robust Reinforcement Learning as a Stackelberg Game
  via Adaptively-Regularized Adversarial Training}.
\newblock \bibinfo{journal}{\emph{arXiv preprint arXiv:2202.09514}}
  (\bibinfo{year}{2022}).
\newblock


\bibitem[\protect\citeauthoryear{Huang, Papernot, Goodfellow, Duan, and
  Abbeel}{Huang et~al\mbox{.}}{2017}]%
        {huang2017adversarial}
\bibfield{author}{\bibinfo{person}{Sandy Huang}, \bibinfo{person}{Nicolas
  Papernot}, \bibinfo{person}{Ian Goodfellow}, \bibinfo{person}{Yan Duan},
  {and} \bibinfo{person}{Pieter Abbeel}.} \bibinfo{year}{2017}\natexlab{}.
\newblock \showarticletitle{Adversarial attacks on neural network policies}.
\newblock \bibinfo{journal}{\emph{arXiv preprint arXiv:1702.02284}}
  (\bibinfo{year}{2017}).
\newblock


\bibitem[\protect\citeauthoryear{Huang and Zhu}{Huang and Zhu}{2019}]%
        {huang2019deceptive}
\bibfield{author}{\bibinfo{person}{Yunhan Huang} {and} \bibinfo{person}{Quanyan
  Zhu}.} \bibinfo{year}{2019}\natexlab{}.
\newblock \showarticletitle{Deceptive reinforcement learning under adversarial
  manipulations on cost signals}. In \bibinfo{booktitle}{\emph{International
  Conference on Decision and Game Theory for Security}}. Springer,
  \bibinfo{pages}{217--237}.
\newblock


\bibitem[\protect\citeauthoryear{Huang, Arief, Lam, and Zhao}{Huang
  et~al\mbox{.}}{2019}]%
        {huang2019evaluation}
\bibfield{author}{\bibinfo{person}{Zhiyuan Huang}, \bibinfo{person}{Mansur
  Arief}, \bibinfo{person}{Henry Lam}, {and} \bibinfo{person}{Ding Zhao}.}
  \bibinfo{year}{2019}\natexlab{}.
\newblock \showarticletitle{Evaluation uncertainty in data-driven self-driving
  testing}. In \bibinfo{booktitle}{\emph{2019 IEEE Intelligent Transportation
  Systems Conference (ITSC)}}. IEEE, \bibinfo{pages}{1902--1907}.
\newblock


\bibitem[\protect\citeauthoryear{Ilahi, Usama, Qadir, Janjua, Al-Fuqaha, Hoang,
  and Niyato}{Ilahi et~al\mbox{.}}{2020}]%
        {ilahi2020challenges}
\bibfield{author}{\bibinfo{person}{Inaam Ilahi}, \bibinfo{person}{Muhammad
  Usama}, \bibinfo{person}{Junaid Qadir}, \bibinfo{person}{Muhammad~Umar
  Janjua}, \bibinfo{person}{Ala Al-Fuqaha}, \bibinfo{person}{Dinh~Thai Hoang},
  {and} \bibinfo{person}{Dusit Niyato}.} \bibinfo{year}{2020}\natexlab{}.
\newblock \showarticletitle{Challenges and countermeasures for adversarial
  attacks on deep reinforcement learning}.
\newblock \bibinfo{journal}{\emph{arXiv preprint arXiv:2001.09684}}
  (\bibinfo{year}{2020}).
\newblock


\bibitem[\protect\citeauthoryear{Ilahi, Usama, Qadir, Janjua, Al-Fuqaha, Hoang,
  and Niyato}{Ilahi et~al\mbox{.}}{2021}]%
        {ilahi2021challenges}
\bibfield{author}{\bibinfo{person}{Inaam Ilahi}, \bibinfo{person}{Muhammad
  Usama}, \bibinfo{person}{Junaid Qadir}, \bibinfo{person}{Muhammad~Umar
  Janjua}, \bibinfo{person}{Ala Al-Fuqaha}, \bibinfo{person}{Dinh~Thai Hoang},
  {and} \bibinfo{person}{Dusit Niyato}.} \bibinfo{year}{2021}\natexlab{}.
\newblock \showarticletitle{Challenges and countermeasures for adversarial
  attacks on deep reinforcement learning}.
\newblock \bibinfo{journal}{\emph{IEEE Transactions on Artificial
  Intelligence}} \bibinfo{volume}{3}, \bibinfo{number}{2}
  (\bibinfo{year}{2021}), \bibinfo{pages}{90--109}.
\newblock


\bibitem[\protect\citeauthoryear{Jaafra, Laurent, Deruyver, and Naceur}{Jaafra
  et~al\mbox{.}}{2019}]%
        {jaafra2019robust}
\bibfield{author}{\bibinfo{person}{Yesmina Jaafra}, \bibinfo{person}{Jean~Luc
  Laurent}, \bibinfo{person}{Aline Deruyver}, {and}
  \bibinfo{person}{Mohamed~Saber Naceur}.} \bibinfo{year}{2019}\natexlab{}.
\newblock \showarticletitle{Robust reinforcement learning for autonomous
  driving}.
\newblock  (\bibinfo{year}{2019}).
\newblock


\bibitem[\protect\citeauthoryear{Jabbari, Joseph, Kearns, Morgenstern, and
  Roth}{Jabbari et~al\mbox{.}}{2017}]%
        {jabbari2017fairness}
\bibfield{author}{\bibinfo{person}{Shahin Jabbari}, \bibinfo{person}{Matthew
  Joseph}, \bibinfo{person}{Michael Kearns}, \bibinfo{person}{Jamie
  Morgenstern}, {and} \bibinfo{person}{Aaron Roth}.}
  \bibinfo{year}{2017}\natexlab{}.
\newblock \showarticletitle{Fairness in reinforcement learning}. In
  \bibinfo{booktitle}{\emph{International conference on machine learning}}.
  PMLR, \bibinfo{pages}{1617--1626}.
\newblock


\bibitem[\protect\citeauthoryear{Jakobi}{Jakobi}{1997}]%
        {jakobi1997evolutionary}
\bibfield{author}{\bibinfo{person}{Nick Jakobi}.}
  \bibinfo{year}{1997}\natexlab{}.
\newblock \showarticletitle{Evolutionary robotics and the radical
  envelope-of-noise hypothesis}.
\newblock \bibinfo{journal}{\emph{Adaptive behavior}} \bibinfo{volume}{6},
  \bibinfo{number}{2} (\bibinfo{year}{1997}), \bibinfo{pages}{325--368}.
\newblock


\bibitem[\protect\citeauthoryear{Jaques, Shen, Ghandeharioun, Ferguson,
  Lapedriza, Jones, Gu, and Picard}{Jaques et~al\mbox{.}}{2020}]%
        {jaques2020human}
\bibfield{author}{\bibinfo{person}{Natasha Jaques},
  \bibinfo{person}{Judy~Hanwen Shen}, \bibinfo{person}{Asma Ghandeharioun},
  \bibinfo{person}{Craig Ferguson}, \bibinfo{person}{Agata Lapedriza},
  \bibinfo{person}{Noah Jones}, \bibinfo{person}{Shixiang~Shane Gu}, {and}
  \bibinfo{person}{Rosalind Picard}.} \bibinfo{year}{2020}\natexlab{}.
\newblock \showarticletitle{Human-centric dialog training via offline
  reinforcement learning}.
\newblock \bibinfo{journal}{\emph{arXiv preprint arXiv:2010.05848}}
  (\bibinfo{year}{2020}).
\newblock


\bibitem[\protect\citeauthoryear{Ji, Yang, Gu, Chen, Lin, and Yang}{Ji
  et~al\mbox{.}}{[n.d.]}]%
        {jibenchmarking}
\bibfield{author}{\bibinfo{person}{Jiamg Ji}, \bibinfo{person}{Long Yang},
  \bibinfo{person}{Shangding Gu}, \bibinfo{person}{Yuanpei Chen},
  \bibinfo{person}{Zhouchen Lin}, {and} \bibinfo{person}{Yaodong Yang}.}
  \bibinfo{year}{[n.d.]}\natexlab{}.
\newblock \showarticletitle{Benchmarking Safe Policy Optimization for
  Constrained Reinforcement Learning}.
\newblock  (\bibinfo{year}{[n.\,d.]}).
\newblock


\bibitem[\protect\citeauthoryear{Jiang, Grefenstette, and
  Rockt{\"a}schel}{Jiang et~al\mbox{.}}{2021}]%
        {jiang2021prioritized}
\bibfield{author}{\bibinfo{person}{Minqi Jiang}, \bibinfo{person}{Edward
  Grefenstette}, {and} \bibinfo{person}{Tim Rockt{\"a}schel}.}
  \bibinfo{year}{2021}\natexlab{}.
\newblock \showarticletitle{Prioritized level replay}. In
  \bibinfo{booktitle}{\emph{International Conference on Machine Learning}}.
  PMLR, \bibinfo{pages}{4940--4950}.
\newblock


\bibitem[\protect\citeauthoryear{Kaelbling}{Kaelbling}{1993}]%
        {Kaelbling93learningto}
\bibfield{author}{\bibinfo{person}{Leslie~Pack Kaelbling}.}
  \bibinfo{year}{1993}\natexlab{}.
\newblock \showarticletitle{Learning to Achieve Goals}. In
  \bibinfo{booktitle}{\emph{IN PROC. OF IJCAI-93}}. \bibinfo{publisher}{Morgan
  Kaufmann}, \bibinfo{pages}{1094--1098}.
\newblock


\bibitem[\protect\citeauthoryear{Kahneman and Tversky}{Kahneman and
  Tversky}{2013}]%
        {kahneman2013prospect}
\bibfield{author}{\bibinfo{person}{Daniel Kahneman} {and} \bibinfo{person}{Amos
  Tversky}.} \bibinfo{year}{2013}\natexlab{}.
\newblock \showarticletitle{Prospect theory: An analysis of decision under
  risk}.
\newblock In \bibinfo{booktitle}{\emph{Handbook of the fundamentals of
  financial decision making: Part I}}. \bibinfo{publisher}{World Scientific},
  \bibinfo{pages}{99--127}.
\newblock


\bibitem[\protect\citeauthoryear{Kaplanis, Shanahan, and Clopath}{Kaplanis
  et~al\mbox{.}}{2018}]%
        {kaplanis2018continual}
\bibfield{author}{\bibinfo{person}{Christos Kaplanis}, \bibinfo{person}{Murray
  Shanahan}, {and} \bibinfo{person}{Claudia Clopath}.}
  \bibinfo{year}{2018}\natexlab{}.
\newblock \showarticletitle{Continual reinforcement learning with complex
  synapses}. In \bibinfo{booktitle}{\emph{International Conference on Machine
  Learning}}. PMLR, \bibinfo{pages}{2497--2506}.
\newblock


\bibitem[\protect\citeauthoryear{Kaspar, Osorio, and Bock}{Kaspar
  et~al\mbox{.}}{2020}]%
        {kaspar2020sim2real}
\bibfield{author}{\bibinfo{person}{Manuel Kaspar}, \bibinfo{person}{Juan
  D~Mu{\~n}oz Osorio}, {and} \bibinfo{person}{J{\"u}rgen Bock}.}
  \bibinfo{year}{2020}\natexlab{}.
\newblock \showarticletitle{Sim2real transfer for reinforcement learning
  without dynamics randomization}. In \bibinfo{booktitle}{\emph{2020 IEEE/RSJ
  International Conference on Intelligent Robots and Systems (IROS)}}. IEEE,
  \bibinfo{pages}{4383--4388}.
\newblock


\bibitem[\protect\citeauthoryear{Ke, Bilaniuk, Goyal, Bauer, Larochelle,
  Sch{\"o}lkopf, Mozer, Pal, and Bengio}{Ke et~al\mbox{.}}{2019}]%
        {ke2019learning}
\bibfield{author}{\bibinfo{person}{Nan~Rosemary Ke}, \bibinfo{person}{Olexa
  Bilaniuk}, \bibinfo{person}{Anirudh Goyal}, \bibinfo{person}{Stefan Bauer},
  \bibinfo{person}{Hugo Larochelle}, \bibinfo{person}{Bernhard Sch{\"o}lkopf},
  \bibinfo{person}{Michael~C Mozer}, \bibinfo{person}{Chris Pal}, {and}
  \bibinfo{person}{Yoshua Bengio}.} \bibinfo{year}{2019}\natexlab{}.
\newblock \showarticletitle{Learning neural causal models from unknown
  interventions}.
\newblock \bibinfo{journal}{\emph{arXiv preprint arXiv:1910.01075}}
  (\bibinfo{year}{2019}).
\newblock


\bibitem[\protect\citeauthoryear{Kirk, Zhang, Grefenstette, and
  Rockt{\"a}schel}{Kirk et~al\mbox{.}}{2021}]%
        {kirk2021survey}
\bibfield{author}{\bibinfo{person}{Robert Kirk}, \bibinfo{person}{Amy Zhang},
  \bibinfo{person}{Edward Grefenstette}, {and} \bibinfo{person}{Tim
  Rockt{\"a}schel}.} \bibinfo{year}{2021}\natexlab{}.
\newblock \showarticletitle{A survey of generalisation in deep reinforcement
  learning}.
\newblock \bibinfo{journal}{\emph{arXiv preprint arXiv:2111.09794}}
  (\bibinfo{year}{2021}).
\newblock


\bibitem[\protect\citeauthoryear{Klima, Bloembergen, Kaisers, and Tuyls}{Klima
  et~al\mbox{.}}{2019}]%
        {klima2019robust}
\bibfield{author}{\bibinfo{person}{Richard Klima}, \bibinfo{person}{Daan
  Bloembergen}, \bibinfo{person}{Michael Kaisers}, {and} \bibinfo{person}{Karl
  Tuyls}.} \bibinfo{year}{2019}\natexlab{}.
\newblock \showarticletitle{Robust temporal difference learning for critical
  domains}.
\newblock \bibinfo{journal}{\emph{arXiv preprint arXiv:1901.08021}}
  (\bibinfo{year}{2019}).
\newblock


\bibitem[\protect\citeauthoryear{Koller, Berkenkamp, Turchetta, and
  Krause}{Koller et~al\mbox{.}}{2018}]%
        {koller2018learning}
\bibfield{author}{\bibinfo{person}{Torsten Koller}, \bibinfo{person}{Felix
  Berkenkamp}, \bibinfo{person}{Matteo Turchetta}, {and}
  \bibinfo{person}{Andreas Krause}.} \bibinfo{year}{2018}\natexlab{}.
\newblock \showarticletitle{Learning-based model predictive control for safe
  exploration}. In \bibinfo{booktitle}{\emph{2018 IEEE Conference on Decision
  and Control (CDC)}}. IEEE, \bibinfo{pages}{6059--6066}.
\newblock


\bibitem[\protect\citeauthoryear{Kos and Song}{Kos and Song}{2017}]%
        {kos2017delving}
\bibfield{author}{\bibinfo{person}{Jernej Kos} {and} \bibinfo{person}{Dawn
  Song}.} \bibinfo{year}{2017}\natexlab{}.
\newblock \showarticletitle{Delving into adversarial attacks on deep policies}.
\newblock \bibinfo{journal}{\emph{arXiv preprint arXiv:1705.06452}}
  (\bibinfo{year}{2017}).
\newblock


\bibitem[\protect\citeauthoryear{Kumar, Levine, and Feizi}{Kumar
  et~al\mbox{.}}{2021}]%
        {kumar2021policy}
\bibfield{author}{\bibinfo{person}{Aounon Kumar}, \bibinfo{person}{Alexander
  Levine}, {and} \bibinfo{person}{Soheil Feizi}.}
  \bibinfo{year}{2021}\natexlab{}.
\newblock \showarticletitle{Policy smoothing for provably robust reinforcement
  learning}.
\newblock \bibinfo{journal}{\emph{arXiv preprint arXiv:2106.11420}}
  (\bibinfo{year}{2021}).
\newblock


\bibitem[\protect\citeauthoryear{Kumar, Kumar, Levine, and Finn}{Kumar
  et~al\mbox{.}}{2020}]%
        {kumar2020one}
\bibfield{author}{\bibinfo{person}{Saurabh Kumar}, \bibinfo{person}{Aviral
  Kumar}, \bibinfo{person}{Sergey Levine}, {and} \bibinfo{person}{Chelsea
  Finn}.} \bibinfo{year}{2020}\natexlab{}.
\newblock \showarticletitle{One solution is not all you need: Few-shot
  extrapolation via structured maxent rl}.
\newblock \bibinfo{journal}{\emph{Advances in Neural Information Processing
  Systems}}  \bibinfo{volume}{33} (\bibinfo{year}{2020}),
  \bibinfo{pages}{8198--8210}.
\newblock


\bibitem[\protect\citeauthoryear{Lecarpentier and Rachelson}{Lecarpentier and
  Rachelson}{2019}]%
        {lecarpentier2019non}
\bibfield{author}{\bibinfo{person}{Erwan Lecarpentier} {and}
  \bibinfo{person}{Emmanuel Rachelson}.} \bibinfo{year}{2019}\natexlab{}.
\newblock \showarticletitle{Non-Stationary Markov Decision Processes, a
  Worst-Case Approach using Model-Based Reinforcement Learning, Extended
  version}.
\newblock \bibinfo{journal}{\emph{arXiv preprint arXiv:1904.10090}}
  (\bibinfo{year}{2019}).
\newblock


\bibitem[\protect\citeauthoryear{Lee and See}{Lee and See}{2004}]%
        {lee2004trust}
\bibfield{author}{\bibinfo{person}{John~D Lee} {and} \bibinfo{person}{Katrina~A
  See}.} \bibinfo{year}{2004}\natexlab{}.
\newblock \showarticletitle{Trust in automation: Designing for appropriate
  reliance}.
\newblock \bibinfo{journal}{\emph{Human factors}} \bibinfo{volume}{46},
  \bibinfo{number}{1} (\bibinfo{year}{2004}), \bibinfo{pages}{50--80}.
\newblock


\bibitem[\protect\citeauthoryear{Leike, Martic, Krakovna, Ortega, Everitt,
  Lefrancq, Orseau, and Legg}{Leike et~al\mbox{.}}{2017}]%
        {leike2017ai}
\bibfield{author}{\bibinfo{person}{Jan Leike}, \bibinfo{person}{Miljan Martic},
  \bibinfo{person}{Victoria Krakovna}, \bibinfo{person}{Pedro~A Ortega},
  \bibinfo{person}{Tom Everitt}, \bibinfo{person}{Andrew Lefrancq},
  \bibinfo{person}{Laurent Orseau}, {and} \bibinfo{person}{Shane Legg}.}
  \bibinfo{year}{2017}\natexlab{}.
\newblock \showarticletitle{AI safety gridworlds}.
\newblock \bibinfo{journal}{\emph{arXiv preprint arXiv:1711.09883}}
  (\bibinfo{year}{2017}).
\newblock


\bibitem[\protect\citeauthoryear{Leurent}{Leurent}{2018}]%
        {highway-env}
\bibfield{author}{\bibinfo{person}{Edouard Leurent}.}
  \bibinfo{year}{2018}\natexlab{}.
\newblock \bibinfo{title}{An Environment for Autonomous Driving
  Decision-Making}.
\newblock
  \bibinfo{howpublished}{\url{https://github.com/eleurent/highway-env}}.
\newblock


\bibitem[\protect\citeauthoryear{Lewis, Sycara, and Walker}{Lewis
  et~al\mbox{.}}{2018}]%
        {lewis2018role}
\bibfield{author}{\bibinfo{person}{Michael Lewis}, \bibinfo{person}{Katia
  Sycara}, {and} \bibinfo{person}{Phillip Walker}.}
  \bibinfo{year}{2018}\natexlab{}.
\newblock \showarticletitle{The role of trust in human-robot interaction}.
\newblock In \bibinfo{booktitle}{\emph{Foundations of trusted autonomy}}.
  \bibinfo{publisher}{Springer, Cham}, \bibinfo{pages}{135--159}.
\newblock


\bibitem[\protect\citeauthoryear{Li and Belta}{Li and Belta}{2019}]%
        {li2019temporal}
\bibfield{author}{\bibinfo{person}{Xiao Li} {and} \bibinfo{person}{Calin
  Belta}.} \bibinfo{year}{2019}\natexlab{}.
\newblock \showarticletitle{Temporal logic guided safe reinforcement learning
  using control barrier functions}.
\newblock \bibinfo{journal}{\emph{arXiv preprint arXiv:1903.09885}}
  (\bibinfo{year}{2019}).
\newblock


\bibitem[\protect\citeauthoryear{Liang, Que, and Modiano}{Liang
  et~al\mbox{.}}{2018}]%
        {liang2018accelerated}
\bibfield{author}{\bibinfo{person}{Qingkai Liang}, \bibinfo{person}{Fanyu Que},
  {and} \bibinfo{person}{Eytan Modiano}.} \bibinfo{year}{2018}\natexlab{}.
\newblock \showarticletitle{Accelerated primal-dual policy optimization for
  safe reinforcement learning}.
\newblock \bibinfo{journal}{\emph{arXiv preprint arXiv:1802.06480}}
  (\bibinfo{year}{2018}).
\newblock


\bibitem[\protect\citeauthoryear{Lin, Hong, Liao, Shih, Liu, and Sun}{Lin
  et~al\mbox{.}}{2017a}]%
        {lin2017tactics}
\bibfield{author}{\bibinfo{person}{Yen-Chen Lin}, \bibinfo{person}{Zhang-Wei
  Hong}, \bibinfo{person}{Yuan-Hong Liao}, \bibinfo{person}{Meng-Li Shih},
  \bibinfo{person}{Ming-Yu Liu}, {and} \bibinfo{person}{Min Sun}.}
  \bibinfo{year}{2017}\natexlab{a}.
\newblock \showarticletitle{Tactics of adversarial attack on deep reinforcement
  learning agents}.
\newblock \bibinfo{journal}{\emph{arXiv preprint arXiv:1703.06748}}
  (\bibinfo{year}{2017}).
\newblock


\bibitem[\protect\citeauthoryear{Lin, Liu, Sun, and Huang}{Lin
  et~al\mbox{.}}{2017b}]%
        {lin2017detecting}
\bibfield{author}{\bibinfo{person}{Yen-Chen Lin}, \bibinfo{person}{Ming-Yu
  Liu}, \bibinfo{person}{Min Sun}, {and} \bibinfo{person}{Jia-Bin Huang}.}
  \bibinfo{year}{2017}\natexlab{b}.
\newblock \showarticletitle{Detecting adversarial attacks on neural network
  policies with visual foresight}.
\newblock \bibinfo{journal}{\emph{arXiv preprint arXiv:1710.00814}}
  (\bibinfo{year}{2017}).
\newblock


\bibitem[\protect\citeauthoryear{Lindgren, Kocaoglu, Dimakis, and
  Vishwanath}{Lindgren et~al\mbox{.}}{2018}]%
        {lindgren2018experimental}
\bibfield{author}{\bibinfo{person}{Erik Lindgren}, \bibinfo{person}{Murat
  Kocaoglu}, \bibinfo{person}{Alexandros~G Dimakis}, {and}
  \bibinfo{person}{Sriram Vishwanath}.} \bibinfo{year}{2018}\natexlab{}.
\newblock \showarticletitle{Experimental design for cost-aware learning of
  causal graphs}.
\newblock \bibinfo{journal}{\emph{Advances in Neural Information Processing
  Systems}}  \bibinfo{volume}{31} (\bibinfo{year}{2018}).
\newblock


\bibitem[\protect\citeauthoryear{Liu, Liu, Tang, Liao, Chen, and Heng}{Liu
  et~al\mbox{.}}{2020a}]%
        {liu2020balancing}
\bibfield{author}{\bibinfo{person}{Weiwen Liu}, \bibinfo{person}{Feng Liu},
  \bibinfo{person}{Ruiming Tang}, \bibinfo{person}{Ben Liao},
  \bibinfo{person}{Guangyong Chen}, {and} \bibinfo{person}{Pheng~Ann Heng}.}
  \bibinfo{year}{2020}\natexlab{a}.
\newblock \showarticletitle{Balancing between accuracy and fairness for
  interactive recommendation with reinforcement learning}. In
  \bibinfo{booktitle}{\emph{Pacific-asia conference on knowledge discovery and
  data mining}}. Springer, \bibinfo{pages}{155--167}.
\newblock


\bibitem[\protect\citeauthoryear{Liu, Mehdipour, and Belta}{Liu
  et~al\mbox{.}}{2021}]%
        {liu2021recurrent}
\bibfield{author}{\bibinfo{person}{Wenliang Liu}, \bibinfo{person}{Noushin
  Mehdipour}, {and} \bibinfo{person}{Calin Belta}.}
  \bibinfo{year}{2021}\natexlab{}.
\newblock \showarticletitle{Recurrent neural network controllers for signal
  temporal logic specifications subject to safety constraints}.
\newblock \bibinfo{journal}{\emph{IEEE Control Systems Letters}}
  (\bibinfo{year}{2021}).
\newblock


\bibitem[\protect\citeauthoryear{Liu, Deng, Choo, and Yang}{Liu
  et~al\mbox{.}}{2019}]%
        {liu2019privacy}
\bibfield{author}{\bibinfo{person}{Ximeng Liu}, \bibinfo{person}{Robert~H
  Deng}, \bibinfo{person}{Kim-Kwang~Raymond Choo}, {and} \bibinfo{person}{Yang
  Yang}.} \bibinfo{year}{2019}\natexlab{}.
\newblock \showarticletitle{Privacy-preserving reinforcement learning design
  for patient-centric dynamic treatment regimes}.
\newblock \bibinfo{journal}{\emph{IEEE Transactions on Emerging Topics in
  Computing}} \bibinfo{volume}{9}, \bibinfo{number}{1} (\bibinfo{year}{2019}),
  \bibinfo{pages}{456--470}.
\newblock


\bibitem[\protect\citeauthoryear{Liu, Cen, Isenbaev, Liu, Wu, Li, and Zhao}{Liu
  et~al\mbox{.}}{2022a}]%
        {liu2022constrained}
\bibfield{author}{\bibinfo{person}{Zuxin Liu}, \bibinfo{person}{Zhepeng Cen},
  \bibinfo{person}{Vladislav Isenbaev}, \bibinfo{person}{Wei Liu},
  \bibinfo{person}{Steven Wu}, \bibinfo{person}{Bo Li}, {and}
  \bibinfo{person}{Ding Zhao}.} \bibinfo{year}{2022}\natexlab{a}.
\newblock \showarticletitle{Constrained variational policy optimization for
  safe reinforcement learning}. In \bibinfo{booktitle}{\emph{International
  Conference on Machine Learning}}. PMLR, \bibinfo{pages}{13644--13668}.
\newblock


\bibitem[\protect\citeauthoryear{Liu, Guo, Cen, Zhang, Tan, Li, and Zhao}{Liu
  et~al\mbox{.}}{2022b}]%
        {liu2022robustness}
\bibfield{author}{\bibinfo{person}{Zuxin Liu}, \bibinfo{person}{Zijian Guo},
  \bibinfo{person}{Zhepeng Cen}, \bibinfo{person}{Huan Zhang},
  \bibinfo{person}{Jie Tan}, \bibinfo{person}{Bo Li}, {and}
  \bibinfo{person}{Ding Zhao}.} \bibinfo{year}{2022}\natexlab{b}.
\newblock \showarticletitle{On the Robustness of Safe Reinforcement Learning
  under Observational Perturbations}.
\newblock \bibinfo{journal}{\emph{arXiv preprint arXiv:2205.14691}}
  (\bibinfo{year}{2022}).
\newblock


\bibitem[\protect\citeauthoryear{Liu, Zhou, Chen, Zhong, Hebert, and Zhao}{Liu
  et~al\mbox{.}}{2020b}]%
        {liu2020constrained}
\bibfield{author}{\bibinfo{person}{Zuxin Liu}, \bibinfo{person}{Hongyi Zhou},
  \bibinfo{person}{Baiming Chen}, \bibinfo{person}{Sicheng Zhong},
  \bibinfo{person}{Martial Hebert}, {and} \bibinfo{person}{Ding Zhao}.}
  \bibinfo{year}{2020}\natexlab{b}.
\newblock \showarticletitle{Constrained Model-based Reinforcement Learning with
  Robust Cross-Entropy Method}.
\newblock \bibinfo{journal}{\emph{arXiv preprint arXiv:2010.07968}}
  (\bibinfo{year}{2020}).
\newblock


\bibitem[\protect\citeauthoryear{L{\"u}tjens, Everett, and How}{L{\"u}tjens
  et~al\mbox{.}}{2020}]%
        {lutjens2020certified}
\bibfield{author}{\bibinfo{person}{Bj{\"o}rn L{\"u}tjens},
  \bibinfo{person}{Michael Everett}, {and} \bibinfo{person}{Jonathan~P How}.}
  \bibinfo{year}{2020}\natexlab{}.
\newblock \showarticletitle{Certified adversarial robustness for deep
  reinforcement learning}. In \bibinfo{booktitle}{\emph{Conference on Robot
  Learning}}. PMLR, \bibinfo{pages}{1328--1337}.
\newblock


\bibitem[\protect\citeauthoryear{Ma, Wang, Zhang, Wang, Zou, and Yang}{Ma
  et~al\mbox{.}}{2019a}]%
        {ma2019differentially}
\bibfield{author}{\bibinfo{person}{Pingchuan Ma}, \bibinfo{person}{Zhiqiang
  Wang}, \bibinfo{person}{Le Zhang}, \bibinfo{person}{Ruming Wang},
  \bibinfo{person}{Xiaoxiang Zou}, {and} \bibinfo{person}{Tao Yang}.}
  \bibinfo{year}{2019}\natexlab{a}.
\newblock \showarticletitle{Differentially private reinforcement learning}. In
  \bibinfo{booktitle}{\emph{International Conference on Information and
  Communications Security}}. Springer, \bibinfo{pages}{668--683}.
\newblock


\bibitem[\protect\citeauthoryear{Ma, Zhang, Sun, and Zhu}{Ma
  et~al\mbox{.}}{2019b}]%
        {ma2019policy}
\bibfield{author}{\bibinfo{person}{Yuzhe Ma}, \bibinfo{person}{Xuezhou Zhang},
  \bibinfo{person}{Wen Sun}, {and} \bibinfo{person}{Jerry Zhu}.}
  \bibinfo{year}{2019}\natexlab{b}.
\newblock \showarticletitle{Policy Poisoning in Batch Reinforcement Learning
  and Control}. In \bibinfo{booktitle}{\emph{Advances in Neural Information
  Processing Systems}}, \bibfield{editor}{\bibinfo{person}{H.~Wallach},
  \bibinfo{person}{H.~Larochelle}, \bibinfo{person}{A.~Beygelzimer},
  \bibinfo{person}{F.~d\textquotesingle Alch\'{e}-Buc},
  \bibinfo{person}{E.~Fox}, {and} \bibinfo{person}{R.~Garnett}} (Eds.),
  Vol.~\bibinfo{volume}{32}. \bibinfo{publisher}{Curran Associates, Inc.}
\newblock
\urldef\tempurl%
\url{https://proceedings.neurips.cc/paper/2019/file/315f006f691ef2e689125614ea22cc61-Paper.pdf}
\showURL{%
\tempurl}


\bibitem[\protect\citeauthoryear{Madry, Makelov, Schmidt, Tsipras, and
  Vladu}{Madry et~al\mbox{.}}{2018}]%
        {madry2018towards}
\bibfield{author}{\bibinfo{person}{Aleksander Madry},
  \bibinfo{person}{Aleksandar Makelov}, \bibinfo{person}{Ludwig Schmidt},
  \bibinfo{person}{Dimitris Tsipras}, {and} \bibinfo{person}{Adrian Vladu}.}
  \bibinfo{year}{2018}\natexlab{}.
\newblock \showarticletitle{Towards Deep Learning Models Resistant to
  Adversarial Attacks}. In \bibinfo{booktitle}{\emph{International Conference
  on Learning Representations}}.
\newblock
\urldef\tempurl%
\url{https://openreview.net/forum?id=rJzIBfZAb}
\showURL{%
\tempurl}


\bibitem[\protect\citeauthoryear{Mandlekar, Zhu, Garg, Fei-Fei, and
  Savarese}{Mandlekar et~al\mbox{.}}{2017}]%
        {mandlekar2017adversarially}
\bibfield{author}{\bibinfo{person}{Ajay Mandlekar}, \bibinfo{person}{Yuke Zhu},
  \bibinfo{person}{Animesh Garg}, \bibinfo{person}{Li Fei-Fei}, {and}
  \bibinfo{person}{Silvio Savarese}.} \bibinfo{year}{2017}\natexlab{}.
\newblock \showarticletitle{Adversarially robust policy learning: Active
  construction of physically-plausible perturbations}. In
  \bibinfo{booktitle}{\emph{2017 IEEE/RSJ International Conference on
  Intelligent Robots and Systems (IROS)}}. IEEE, \bibinfo{pages}{3932--3939}.
\newblock


\bibitem[\protect\citeauthoryear{Mankowitz, Levine, Jeong, Shi, Kay,
  Abdolmaleki, Springenberg, Mann, Hester, and Riedmiller}{Mankowitz
  et~al\mbox{.}}{2019}]%
        {mankowitz2019robust}
\bibfield{author}{\bibinfo{person}{Daniel~J Mankowitz}, \bibinfo{person}{Nir
  Levine}, \bibinfo{person}{Rae Jeong}, \bibinfo{person}{Yuanyuan Shi},
  \bibinfo{person}{Jackie Kay}, \bibinfo{person}{Abbas Abdolmaleki},
  \bibinfo{person}{Jost~Tobias Springenberg}, \bibinfo{person}{Timothy Mann},
  \bibinfo{person}{Todd Hester}, {and} \bibinfo{person}{Martin Riedmiller}.}
  \bibinfo{year}{2019}\natexlab{}.
\newblock \showarticletitle{Robust reinforcement learning for continuous
  control with model misspecification}.
\newblock \bibinfo{journal}{\emph{arXiv preprint arXiv:1906.07516}}
  (\bibinfo{year}{2019}).
\newblock


\bibitem[\protect\citeauthoryear{Mannor, Mebel, and Xu}{Mannor
  et~al\mbox{.}}{2016}]%
        {mannor2016robust}
\bibfield{author}{\bibinfo{person}{Shie Mannor}, \bibinfo{person}{Ofir Mebel},
  {and} \bibinfo{person}{Huan Xu}.} \bibinfo{year}{2016}\natexlab{}.
\newblock \showarticletitle{Robust MDPs with k-rectangular uncertainty}.
\newblock \bibinfo{journal}{\emph{Mathematics of Operations Research}}
  \bibinfo{volume}{41}, \bibinfo{number}{4} (\bibinfo{year}{2016}),
  \bibinfo{pages}{1484--1509}.
\newblock


\bibitem[\protect\citeauthoryear{Mehta, Deleu, Raparthy, Pal, and Paull}{Mehta
  et~al\mbox{.}}{2020a}]%
        {mehta2020curriculum}
\bibfield{author}{\bibinfo{person}{Bhairav Mehta}, \bibinfo{person}{Tristan
  Deleu}, \bibinfo{person}{Sharath~Chandra Raparthy}, \bibinfo{person}{Chris~J
  Pal}, {and} \bibinfo{person}{Liam Paull}.} \bibinfo{year}{2020}\natexlab{a}.
\newblock \showarticletitle{Curriculum in gradient-based meta-reinforcement
  learning}.
\newblock \bibinfo{journal}{\emph{arXiv preprint arXiv:2002.07956}}
  (\bibinfo{year}{2020}).
\newblock


\bibitem[\protect\citeauthoryear{Mehta, Diaz, Golemo, Pal, and Paull}{Mehta
  et~al\mbox{.}}{2020b}]%
        {mehta2020active}
\bibfield{author}{\bibinfo{person}{Bhairav Mehta}, \bibinfo{person}{Manfred
  Diaz}, \bibinfo{person}{Florian Golemo}, \bibinfo{person}{Christopher~J Pal},
  {and} \bibinfo{person}{Liam Paull}.} \bibinfo{year}{2020}\natexlab{b}.
\newblock \showarticletitle{Active domain randomization}. In
  \bibinfo{booktitle}{\emph{Conference on Robot Learning}}. PMLR,
  \bibinfo{pages}{1162--1176}.
\newblock


\bibitem[\protect\citeauthoryear{Mendez, Hussing, Gummadi, and Eaton}{Mendez
  et~al\mbox{.}}{2022}]%
        {mendez2022composuite}
\bibfield{author}{\bibinfo{person}{Jorge~A Mendez}, \bibinfo{person}{Marcel
  Hussing}, \bibinfo{person}{Meghna Gummadi}, {and} \bibinfo{person}{Eric
  Eaton}.} \bibinfo{year}{2022}\natexlab{}.
\newblock \showarticletitle{CompoSuite: A Compositional Reinforcement Learning
  Benchmark}.
\newblock \bibinfo{journal}{\emph{arXiv preprint arXiv:2207.04136}}
  (\bibinfo{year}{2022}).
\newblock


\bibitem[\protect\citeauthoryear{Miriyev and Kova{\v{c}}}{Miriyev and
  Kova{\v{c}}}{2020}]%
        {miriyev2020skills}
\bibfield{author}{\bibinfo{person}{Aslan Miriyev} {and} \bibinfo{person}{Mirko
  Kova{\v{c}}}.} \bibinfo{year}{2020}\natexlab{}.
\newblock \showarticletitle{Skills for physical artificial intelligence}.
\newblock \bibinfo{journal}{\emph{Nature Machine Intelligence}}
  \bibinfo{volume}{2}, \bibinfo{number}{11} (\bibinfo{year}{2020}),
  \bibinfo{pages}{658--660}.
\newblock


\bibitem[\protect\citeauthoryear{Moos, Hansel, Abdulsamad, Stark, Clever, and
  Peters}{Moos et~al\mbox{.}}{2022}]%
        {moos2022robust}
\bibfield{author}{\bibinfo{person}{Janosch Moos}, \bibinfo{person}{Kay Hansel},
  \bibinfo{person}{Hany Abdulsamad}, \bibinfo{person}{Svenja Stark},
  \bibinfo{person}{Debora Clever}, {and} \bibinfo{person}{Jan Peters}.}
  \bibinfo{year}{2022}\natexlab{}.
\newblock \showarticletitle{Robust Reinforcement Learning: A Review of
  Foundations and Recent Advances}.
\newblock \bibinfo{journal}{\emph{Machine Learning and Knowledge Extraction}}
  \bibinfo{volume}{4}, \bibinfo{number}{1} (\bibinfo{year}{2022}),
  \bibinfo{pages}{276--315}.
\newblock


\bibitem[\protect\citeauthoryear{Mordatch, Lowrey, and Todorov}{Mordatch
  et~al\mbox{.}}{2015}]%
        {mordatch2015ensemble}
\bibfield{author}{\bibinfo{person}{Igor Mordatch}, \bibinfo{person}{Kendall
  Lowrey}, {and} \bibinfo{person}{Emanuel Todorov}.}
  \bibinfo{year}{2015}\natexlab{}.
\newblock \showarticletitle{Ensemble-cio: Full-body dynamic motion planning
  that transfers to physical humanoids}. In \bibinfo{booktitle}{\emph{2015
  IEEE/RSJ International Conference on Intelligent Robots and Systems (IROS)}}.
  IEEE, \bibinfo{pages}{5307--5314}.
\newblock


\bibitem[\protect\citeauthoryear{Morgenstern and Von~Neumann}{Morgenstern and
  Von~Neumann}{1953}]%
        {morgenstern1953theory}
\bibfield{author}{\bibinfo{person}{Oskar Morgenstern} {and}
  \bibinfo{person}{John Von~Neumann}.} \bibinfo{year}{1953}\natexlab{}.
\newblock \bibinfo{booktitle}{\emph{Theory of games and economic behavior}}.
\newblock \bibinfo{publisher}{Princeton university press}.
\newblock


\bibitem[\protect\citeauthoryear{Muratore, Eilers, Gienger, and
  Peters}{Muratore et~al\mbox{.}}{2020}]%
        {muratore2020bayesian}
\bibfield{author}{\bibinfo{person}{Fabio Muratore}, \bibinfo{person}{Christian
  Eilers}, \bibinfo{person}{Michael Gienger}, {and} \bibinfo{person}{Jan
  Peters}.} \bibinfo{year}{2020}\natexlab{}.
\newblock \showarticletitle{Bayesian domain randomization for sim-to-real
  transfer}.
\newblock \bibinfo{journal}{\emph{arXiv e-prints}} (\bibinfo{year}{2020}),
  \bibinfo{pages}{arXiv--2003}.
\newblock


\bibitem[\protect\citeauthoryear{Nagabandi, Clavera, Liu, Fearing, Abbeel,
  Levine, and Finn}{Nagabandi et~al\mbox{.}}{2018a}]%
        {nagabandi2018learning}
\bibfield{author}{\bibinfo{person}{Anusha Nagabandi}, \bibinfo{person}{Ignasi
  Clavera}, \bibinfo{person}{Simin Liu}, \bibinfo{person}{Ronald~S Fearing},
  \bibinfo{person}{Pieter Abbeel}, \bibinfo{person}{Sergey Levine}, {and}
  \bibinfo{person}{Chelsea Finn}.} \bibinfo{year}{2018}\natexlab{a}.
\newblock \showarticletitle{Learning to adapt in dynamic, real-world
  environments through meta-reinforcement learning}.
\newblock \bibinfo{journal}{\emph{arXiv preprint arXiv:1803.11347}}
  (\bibinfo{year}{2018}).
\newblock


\bibitem[\protect\citeauthoryear{Nagabandi, Finn, and Levine}{Nagabandi
  et~al\mbox{.}}{2018b}]%
        {nagabandi2018deep}
\bibfield{author}{\bibinfo{person}{Anusha Nagabandi}, \bibinfo{person}{Chelsea
  Finn}, {and} \bibinfo{person}{Sergey Levine}.}
  \bibinfo{year}{2018}\natexlab{b}.
\newblock \showarticletitle{Deep online learning via meta-learning: Continual
  adaptation for model-based rl}.
\newblock \bibinfo{journal}{\emph{arXiv preprint arXiv:1812.07671}}
  (\bibinfo{year}{2018}).
\newblock


\bibitem[\protect\citeauthoryear{Nakao, Jiang, and Shen}{Nakao
  et~al\mbox{.}}{2021}]%
        {nakao2021distributionally}
\bibfield{author}{\bibinfo{person}{Hideaki Nakao}, \bibinfo{person}{Ruiwei
  Jiang}, {and} \bibinfo{person}{Siqian Shen}.}
  \bibinfo{year}{2021}\natexlab{}.
\newblock \showarticletitle{Distributionally Robust Partially Observable Markov
  Decision Process with Moment-Based Ambiguity}.
\newblock \bibinfo{journal}{\emph{SIAM Journal on Optimization}}
  \bibinfo{volume}{31}, \bibinfo{number}{1} (\bibinfo{year}{2021}),
  \bibinfo{pages}{461--488}.
\newblock


\bibitem[\protect\citeauthoryear{Nam, Walker, Lewis, and Sycara}{Nam
  et~al\mbox{.}}{2017}]%
        {nam2017predicting}
\bibfield{author}{\bibinfo{person}{Changjoo Nam}, \bibinfo{person}{Phillip
  Walker}, \bibinfo{person}{Michael Lewis}, {and} \bibinfo{person}{Katia
  Sycara}.} \bibinfo{year}{2017}\natexlab{}.
\newblock \showarticletitle{Predicting trust in human control of swarms via
  inverse reinforcement learning}. In \bibinfo{booktitle}{\emph{2017 26th ieee
  international symposium on robot and human interactive communication
  (ro-man)}}. IEEE, \bibinfo{pages}{528--533}.
\newblock


\bibitem[\protect\citeauthoryear{Nilim and El~Ghaoui}{Nilim and
  El~Ghaoui}{2005}]%
        {nilim2005robust}
\bibfield{author}{\bibinfo{person}{Arnab Nilim} {and} \bibinfo{person}{Laurent
  El~Ghaoui}.} \bibinfo{year}{2005}\natexlab{}.
\newblock \showarticletitle{Robust control of Markov decision processes with
  uncertain transition matrices}.
\newblock \bibinfo{journal}{\emph{Operations Research}} \bibinfo{volume}{53},
  \bibinfo{number}{5} (\bibinfo{year}{2005}), \bibinfo{pages}{780--798}.
\newblock


\bibitem[\protect\citeauthoryear{Oikarinen, Zhang, Megretski, Daniel, and
  Weng}{Oikarinen et~al\mbox{.}}{2021}]%
        {oikarinen2021robust}
\bibfield{author}{\bibinfo{person}{Tuomas Oikarinen}, \bibinfo{person}{Wang
  Zhang}, \bibinfo{person}{Alexandre Megretski}, \bibinfo{person}{Luca Daniel},
  {and} \bibinfo{person}{Tsui-Wei Weng}.} \bibinfo{year}{2021}\natexlab{}.
\newblock \showarticletitle{Robust Deep Reinforcement Learning through
  Adversarial Loss}. In \bibinfo{booktitle}{\emph{Advances in Neural
  Information Processing Systems}},
  \bibfield{editor}{\bibinfo{person}{A.~Beygelzimer},
  \bibinfo{person}{Y.~Dauphin}, \bibinfo{person}{P.~Liang}, {and}
  \bibinfo{person}{J.~Wortman Vaughan}} (Eds.).
\newblock
\urldef\tempurl%
\url{https://openreview.net/forum?id=eaAM_bdW0Q}
\showURL{%
\tempurl}


\bibitem[\protect\citeauthoryear{Ono and Takahashi}{Ono and Takahashi}{2020}]%
        {ono2020locally}
\bibfield{author}{\bibinfo{person}{Hajime Ono} {and} \bibinfo{person}{Tsubasa
  Takahashi}.} \bibinfo{year}{2020}\natexlab{}.
\newblock \showarticletitle{Locally private distributed reinforcement
  learning}.
\newblock \bibinfo{journal}{\emph{arXiv preprint arXiv:2001.11718}}
  (\bibinfo{year}{2020}).
\newblock


\bibitem[\protect\citeauthoryear{Osogami}{Osogami}{2015}]%
        {osogami2015robust}
\bibfield{author}{\bibinfo{person}{Takayuki Osogami}.}
  \bibinfo{year}{2015}\natexlab{}.
\newblock \showarticletitle{Robust partially observable Markov decision
  process}. In \bibinfo{booktitle}{\emph{International Conference on Machine
  Learning}}. PMLR, \bibinfo{pages}{106--115}.
\newblock


\bibitem[\protect\citeauthoryear{Panzer and Bender}{Panzer and Bender}{2022}]%
        {doi:10.1080/00207543.2021.1973138}
\bibfield{author}{\bibinfo{person}{Marcel Panzer} {and}
  \bibinfo{person}{Benedict Bender}.} \bibinfo{year}{2022}\natexlab{}.
\newblock \showarticletitle{Deep reinforcement learning in production systems:
  a systematic literature review}.
\newblock \bibinfo{journal}{\emph{International Journal of Production
  Research}} \bibinfo{volume}{60}, \bibinfo{number}{13} (\bibinfo{year}{2022}),
  \bibinfo{pages}{4316--4341}.
\newblock
\urldef\tempurl%
\url{https://doi.org/10.1080/00207543.2021.1973138}
\showDOI{\tempurl}
\showeprint{https://doi.org/10.1080/00207543.2021.1973138}


\bibitem[\protect\citeauthoryear{Parasuraman and Riley}{Parasuraman and
  Riley}{1997}]%
        {parasuraman1997humans}
\bibfield{author}{\bibinfo{person}{Raja Parasuraman} {and}
  \bibinfo{person}{Victor Riley}.} \bibinfo{year}{1997}\natexlab{}.
\newblock \showarticletitle{Humans and automation: Use, misuse, disuse, abuse}.
\newblock \bibinfo{journal}{\emph{Human factors}} \bibinfo{volume}{39},
  \bibinfo{number}{2} (\bibinfo{year}{1997}), \bibinfo{pages}{230--253}.
\newblock


\bibitem[\protect\citeauthoryear{Paternain, Chamon, Calvo-Fullana, and
  Ribeiro}{Paternain et~al\mbox{.}}{2019}]%
        {paternain2019constrained}
\bibfield{author}{\bibinfo{person}{Santiago Paternain},
  \bibinfo{person}{Luiz~FO Chamon}, \bibinfo{person}{Miguel Calvo-Fullana},
  {and} \bibinfo{person}{Alejandro Ribeiro}.} \bibinfo{year}{2019}\natexlab{}.
\newblock \showarticletitle{Constrained reinforcement learning has zero duality
  gap}.
\newblock \bibinfo{journal}{\emph{arXiv preprint arXiv:1910.13393}}
  (\bibinfo{year}{2019}).
\newblock


\bibitem[\protect\citeauthoryear{Pattanaik, Tang, Liu, Bommannan, and
  Chowdhary}{Pattanaik et~al\mbox{.}}{2018}]%
        {pattanaik2018robust}
\bibfield{author}{\bibinfo{person}{Anay Pattanaik}, \bibinfo{person}{Zhenyi
  Tang}, \bibinfo{person}{Shuijing Liu}, \bibinfo{person}{Gautham Bommannan},
  {and} \bibinfo{person}{Girish Chowdhary}.} \bibinfo{year}{2018}\natexlab{}.
\newblock \showarticletitle{Robust Deep Reinforcement Learning with adversarial
  attacks}. In \bibinfo{booktitle}{\emph{17th International Conference on
  Autonomous Agents and Multiagent Systems, AAMAS 2018}}. International
  Foundation for Autonomous Agents and Multiagent Systems (IFAAMAS),
  \bibinfo{pages}{2040--2042}.
\newblock


\bibitem[\protect\citeauthoryear{Peng, Andrychowicz, Zaremba, and Abbeel}{Peng
  et~al\mbox{.}}{2018}]%
        {peng2018sim}
\bibfield{author}{\bibinfo{person}{Xue~Bin Peng}, \bibinfo{person}{Marcin
  Andrychowicz}, \bibinfo{person}{Wojciech Zaremba}, {and}
  \bibinfo{person}{Pieter Abbeel}.} \bibinfo{year}{2018}\natexlab{}.
\newblock \showarticletitle{Sim-to-real transfer of robotic control with
  dynamics randomization}. In \bibinfo{booktitle}{\emph{2018 IEEE international
  conference on robotics and automation (ICRA)}}. IEEE,
  \bibinfo{pages}{3803--3810}.
\newblock


\bibitem[\protect\citeauthoryear{Peng, Li, Liu, and Zhou}{Peng
  et~al\mbox{.}}{2022}]%
        {peng2022safe}
\bibfield{author}{\bibinfo{person}{Zhenghao Peng}, \bibinfo{person}{Quanyi Li},
  \bibinfo{person}{Chunxiao Liu}, {and} \bibinfo{person}{Bolei Zhou}.}
  \bibinfo{year}{2022}\natexlab{}.
\newblock \showarticletitle{Safe driving via expert guided policy
  optimization}. In \bibinfo{booktitle}{\emph{Conference on Robot Learning}}.
  PMLR, \bibinfo{pages}{1554--1563}.
\newblock


\bibitem[\protect\citeauthoryear{Pham, De~Magistris, and Tachibana}{Pham
  et~al\mbox{.}}{2018}]%
        {pham2018optlayer}
\bibfield{author}{\bibinfo{person}{Tu-Hoa Pham}, \bibinfo{person}{Giovanni
  De~Magistris}, {and} \bibinfo{person}{Ryuki Tachibana}.}
  \bibinfo{year}{2018}\natexlab{}.
\newblock \showarticletitle{Optlayer-practical constrained optimization for
  deep reinforcement learning in the real world}. In
  \bibinfo{booktitle}{\emph{2018 IEEE International Conference on Robotics and
  Automation (ICRA)}}. IEEE, \bibinfo{pages}{6236--6243}.
\newblock


\bibitem[\protect\citeauthoryear{Pinto, Davidson, Sukthankar, and Gupta}{Pinto
  et~al\mbox{.}}{2017}]%
        {pinto2017robust}
\bibfield{author}{\bibinfo{person}{Lerrel Pinto}, \bibinfo{person}{James
  Davidson}, \bibinfo{person}{Rahul Sukthankar}, {and} \bibinfo{person}{Abhinav
  Gupta}.} \bibinfo{year}{2017}\natexlab{}.
\newblock \showarticletitle{Robust adversarial reinforcement learning}. In
  \bibinfo{booktitle}{\emph{International Conference on Machine Learning}}.
  PMLR, \bibinfo{pages}{2817--2826}.
\newblock


\bibitem[\protect\citeauthoryear{Portelas, Colas, Hofmann, and
  Oudeyer}{Portelas et~al\mbox{.}}{2020}]%
        {portelas2020teacher}
\bibfield{author}{\bibinfo{person}{R{\'e}my Portelas},
  \bibinfo{person}{C{\'e}dric Colas}, \bibinfo{person}{Katja Hofmann}, {and}
  \bibinfo{person}{Pierre-Yves Oudeyer}.} \bibinfo{year}{2020}\natexlab{}.
\newblock \showarticletitle{Teacher algorithms for curriculum learning of deep
  rl in continuously parameterized environments}. In
  \bibinfo{booktitle}{\emph{Conference on Robot Learning}}. PMLR,
  \bibinfo{pages}{835--853}.
\newblock


\bibitem[\protect\citeauthoryear{Puranic, Deshmukh, and Nikolaidis}{Puranic
  et~al\mbox{.}}{2021}]%
        {puranic2021learning}
\bibfield{author}{\bibinfo{person}{Aniruddh~Gopinath Puranic},
  \bibinfo{person}{Jyotirmoy Deshmukh}, {and} \bibinfo{person}{Stefanos
  Nikolaidis}.} \bibinfo{year}{2021}\natexlab{}.
\newblock \showarticletitle{Learning from Demonstrations Using Signal Temporal
  Logic in Stochastic and Continuous Domains}.
\newblock \bibinfo{journal}{\emph{IEEE Robotics and Automation Letters}}
  (\bibinfo{year}{2021}).
\newblock


\bibitem[\protect\citeauthoryear{Radulescu, Niv, and Ballard}{Radulescu
  et~al\mbox{.}}{2019}]%
        {radulescu2019holistic}
\bibfield{author}{\bibinfo{person}{Angela Radulescu}, \bibinfo{person}{Yael
  Niv}, {and} \bibinfo{person}{Ian Ballard}.} \bibinfo{year}{2019}\natexlab{}.
\newblock \showarticletitle{Holistic reinforcement learning: the role of
  structure and attention}.
\newblock \bibinfo{journal}{\emph{Trends in cognitive sciences}}
  \bibinfo{volume}{23}, \bibinfo{number}{4} (\bibinfo{year}{2019}),
  \bibinfo{pages}{278--292}.
\newblock


\bibitem[\protect\citeauthoryear{Rahimian and Mehrotra}{Rahimian and
  Mehrotra}{2019}]%
        {rahimian2019distributionally}
\bibfield{author}{\bibinfo{person}{Hamed Rahimian} {and}
  \bibinfo{person}{Sanjay Mehrotra}.} \bibinfo{year}{2019}\natexlab{}.
\newblock \showarticletitle{Distributionally robust optimization: A review}.
\newblock \bibinfo{journal}{\emph{arXiv preprint arXiv:1908.05659}}
  (\bibinfo{year}{2019}).
\newblock


\bibitem[\protect\citeauthoryear{Rajeswaran, Ghotra, Ravindran, and
  Levine}{Rajeswaran et~al\mbox{.}}{2016}]%
        {rajeswaran2016epopt}
\bibfield{author}{\bibinfo{person}{Aravind Rajeswaran},
  \bibinfo{person}{Sarvjeet Ghotra}, \bibinfo{person}{Balaraman Ravindran},
  {and} \bibinfo{person}{Sergey Levine}.} \bibinfo{year}{2016}\natexlab{}.
\newblock \showarticletitle{Epopt: Learning robust neural network policies
  using model ensembles}.
\newblock \bibinfo{journal}{\emph{arXiv preprint arXiv:1610.01283}}
  (\bibinfo{year}{2016}).
\newblock


\bibitem[\protect\citeauthoryear{Rasouli and Saghafian}{Rasouli and
  Saghafian}{2018}]%
        {rasouli2018robust}
\bibfield{author}{\bibinfo{person}{Mohammad Rasouli} {and}
  \bibinfo{person}{Soroush Saghafian}.} \bibinfo{year}{2018}\natexlab{}.
\newblock \showarticletitle{Robust partially observable Markov decision
  processes}.
\newblock  (\bibinfo{year}{2018}).
\newblock


\bibitem[\protect\citeauthoryear{Ratliff and Mazumdar}{Ratliff and
  Mazumdar}{2019}]%
        {ratliff2019inverse}
\bibfield{author}{\bibinfo{person}{Lillian~J Ratliff} {and}
  \bibinfo{person}{Eric Mazumdar}.} \bibinfo{year}{2019}\natexlab{}.
\newblock \showarticletitle{Inverse risk-sensitive reinforcement learning}.
\newblock \bibinfo{journal}{\emph{IEEE Trans. Automat. Control}}
  \bibinfo{volume}{65}, \bibinfo{number}{3} (\bibinfo{year}{2019}),
  \bibinfo{pages}{1256--1263}.
\newblock


\bibitem[\protect\citeauthoryear{Ray, Achiam, and Amodei}{Ray
  et~al\mbox{.}}{2019}]%
        {ray2019benchmarking}
\bibfield{author}{\bibinfo{person}{Alex Ray}, \bibinfo{person}{Joshua Achiam},
  {and} \bibinfo{person}{Dario Amodei}.} \bibinfo{year}{2019}\natexlab{}.
\newblock \showarticletitle{Benchmarking safe exploration in deep reinforcement
  learning}.
\newblock \bibinfo{journal}{\emph{arXiv preprint arXiv:1910.01708}}
  \bibinfo{volume}{7} (\bibinfo{year}{2019}).
\newblock


\bibitem[\protect\citeauthoryear{Reddy, Dragan, and Levine}{Reddy
  et~al\mbox{.}}{2018}]%
        {reddy2018you}
\bibfield{author}{\bibinfo{person}{Siddharth Reddy}, \bibinfo{person}{Anca~D
  Dragan}, {and} \bibinfo{person}{Sergey Levine}.}
  \bibinfo{year}{2018}\natexlab{}.
\newblock \showarticletitle{Where do you think you're going?: Inferring beliefs
  about dynamics from behavior}.
\newblock \bibinfo{journal}{\emph{arXiv preprint arXiv:1805.08010}}
  (\bibinfo{year}{2018}).
\newblock


\bibitem[\protect\citeauthoryear{Richards, Berkenkamp, and Krause}{Richards
  et~al\mbox{.}}{2018}]%
        {richards2018lyapunov}
\bibfield{author}{\bibinfo{person}{Spencer~M Richards}, \bibinfo{person}{Felix
  Berkenkamp}, {and} \bibinfo{person}{Andreas Krause}.}
  \bibinfo{year}{2018}\natexlab{}.
\newblock \showarticletitle{The lyapunov neural network: Adaptive stability
  certification for safe learning of dynamical systems}. In
  \bibinfo{booktitle}{\emph{Conference on Robot Learning}}. PMLR,
  \bibinfo{pages}{466--476}.
\newblock


\bibitem[\protect\citeauthoryear{Romoff, Henderson, Pich{\'e}, Francois-Lavet,
  and Pineau}{Romoff et~al\mbox{.}}{2018}]%
        {romoff2018reward}
\bibfield{author}{\bibinfo{person}{Joshua Romoff}, \bibinfo{person}{Peter
  Henderson}, \bibinfo{person}{Alexandre Pich{\'e}}, \bibinfo{person}{Vincent
  Francois-Lavet}, {and} \bibinfo{person}{Joelle Pineau}.}
  \bibinfo{year}{2018}\natexlab{}.
\newblock \showarticletitle{Reward estimation for variance reduction in deep
  reinforcement learning}.
\newblock \bibinfo{journal}{\emph{arXiv preprint arXiv:1805.03359}}
  (\bibinfo{year}{2018}).
\newblock


\bibitem[\protect\citeauthoryear{Rupprecht, Ibrahim, and Pal}{Rupprecht
  et~al\mbox{.}}{2019}]%
        {rupprecht2019finding}
\bibfield{author}{\bibinfo{person}{Christian Rupprecht}, \bibinfo{person}{Cyril
  Ibrahim}, {and} \bibinfo{person}{Christopher~J Pal}.}
  \bibinfo{year}{2019}\natexlab{}.
\newblock \showarticletitle{Finding and visualizing weaknesses of deep
  reinforcement learning agents}.
\newblock \bibinfo{journal}{\emph{arXiv preprint arXiv:1904.01318}}
  (\bibinfo{year}{2019}).
\newblock


\bibitem[\protect\citeauthoryear{Sadeghi and Levine}{Sadeghi and
  Levine}{2016}]%
        {sadeghi2016cad2rl}
\bibfield{author}{\bibinfo{person}{Fereshteh Sadeghi} {and}
  \bibinfo{person}{Sergey Levine}.} \bibinfo{year}{2016}\natexlab{}.
\newblock \showarticletitle{Cad2rl: Real single-image flight without a single
  real image}.
\newblock \bibinfo{journal}{\emph{arXiv preprint arXiv:1611.04201}}
  (\bibinfo{year}{2016}).
\newblock


\bibitem[\protect\citeauthoryear{Sadeghi, Toshev, Jang, and Levine}{Sadeghi
  et~al\mbox{.}}{2018}]%
        {sadeghi2018sim2real}
\bibfield{author}{\bibinfo{person}{Fereshteh Sadeghi},
  \bibinfo{person}{Alexander Toshev}, \bibinfo{person}{Eric Jang}, {and}
  \bibinfo{person}{Sergey Levine}.} \bibinfo{year}{2018}\natexlab{}.
\newblock \showarticletitle{Sim2real viewpoint invariant visual servoing by
  recurrent control}. In \bibinfo{booktitle}{\emph{Proceedings of the IEEE
  Conference on Computer Vision and Pattern Recognition}}.
  \bibinfo{pages}{4691--4699}.
\newblock


\bibitem[\protect\citeauthoryear{S{\ae}mundsson, Hofmann, and
  Deisenroth}{S{\ae}mundsson et~al\mbox{.}}{2018}]%
        {saemundsson2018meta}
\bibfield{author}{\bibinfo{person}{Steind{\'o}r S{\ae}mundsson},
  \bibinfo{person}{Katja Hofmann}, {and} \bibinfo{person}{Marc~Peter
  Deisenroth}.} \bibinfo{year}{2018}\natexlab{}.
\newblock \showarticletitle{Meta reinforcement learning with latent variable
  gaussian processes}.
\newblock \bibinfo{journal}{\emph{arXiv preprint arXiv:1803.07551}}
  (\bibinfo{year}{2018}).
\newblock


\bibitem[\protect\citeauthoryear{Saghafian}{Saghafian}{2018}]%
        {saghafian2018ambiguous}
\bibfield{author}{\bibinfo{person}{Soroush Saghafian}.}
  \bibinfo{year}{2018}\natexlab{}.
\newblock \showarticletitle{Ambiguous partially observable Markov decision
  processes: Structural results and applications}.
\newblock \bibinfo{journal}{\emph{Journal of Economic Theory}}
  \bibinfo{volume}{178} (\bibinfo{year}{2018}), \bibinfo{pages}{1--35}.
\newblock


\bibitem[\protect\citeauthoryear{Sakuma, Kobayashi, and Wright}{Sakuma
  et~al\mbox{.}}{2008}]%
        {sakuma2008privacy}
\bibfield{author}{\bibinfo{person}{Jun Sakuma}, \bibinfo{person}{Shigenobu
  Kobayashi}, {and} \bibinfo{person}{Rebecca~N Wright}.}
  \bibinfo{year}{2008}\natexlab{}.
\newblock \showarticletitle{Privacy-preserving reinforcement learning}. In
  \bibinfo{booktitle}{\emph{Proceedings of the 25th international conference on
  Machine learning}}. \bibinfo{pages}{864--871}.
\newblock


\bibitem[\protect\citeauthoryear{Schaul, Horgan, Gregor, and Silver}{Schaul
  et~al\mbox{.}}{2015}]%
        {pmlr-v37-schaul15}
\bibfield{author}{\bibinfo{person}{Tom Schaul}, \bibinfo{person}{Daniel
  Horgan}, \bibinfo{person}{Karol Gregor}, {and} \bibinfo{person}{David
  Silver}.} \bibinfo{year}{2015}\natexlab{}.
\newblock \showarticletitle{Universal Value Function Approximators}. In
  \bibinfo{booktitle}{\emph{Proceedings of the 32nd International Conference on
  Machine Learning}} \emph{(\bibinfo{series}{Proceedings of Machine Learning
  Research}, Vol.~\bibinfo{volume}{37})},
  \bibfield{editor}{\bibinfo{person}{Francis Bach} {and} \bibinfo{person}{David
  Blei}} (Eds.). \bibinfo{publisher}{PMLR}, \bibinfo{address}{Lille, France},
  \bibinfo{pages}{1312--1320}.
\newblock
\urldef\tempurl%
\url{https://proceedings.mlr.press/v37/schaul15.html}
\showURL{%
\tempurl}


\bibitem[\protect\citeauthoryear{Scherrer, Bilaniuk, Annadani, Goyal, Schwab,
  Sch{\"o}lkopf, Mozer, Bengio, Bauer, and Ke}{Scherrer et~al\mbox{.}}{2021}]%
        {scherrer2021learning}
\bibfield{author}{\bibinfo{person}{Nino Scherrer}, \bibinfo{person}{Olexa
  Bilaniuk}, \bibinfo{person}{Yashas Annadani}, \bibinfo{person}{Anirudh
  Goyal}, \bibinfo{person}{Patrick Schwab}, \bibinfo{person}{Bernhard
  Sch{\"o}lkopf}, \bibinfo{person}{Michael~C Mozer}, \bibinfo{person}{Yoshua
  Bengio}, \bibinfo{person}{Stefan Bauer}, {and} \bibinfo{person}{Nan~Rosemary
  Ke}.} \bibinfo{year}{2021}\natexlab{}.
\newblock \showarticletitle{Learning neural causal models with active
  interventions}.
\newblock \bibinfo{journal}{\emph{arXiv preprint arXiv:2109.02429}}
  (\bibinfo{year}{2021}).
\newblock


\bibitem[\protect\citeauthoryear{Schulman, Levine, Abbeel, Jordan, and
  Moritz}{Schulman et~al\mbox{.}}{2015}]%
        {schulman2015trust}
\bibfield{author}{\bibinfo{person}{John Schulman}, \bibinfo{person}{Sergey
  Levine}, \bibinfo{person}{Pieter Abbeel}, \bibinfo{person}{Michael Jordan},
  {and} \bibinfo{person}{Philipp Moritz}.} \bibinfo{year}{2015}\natexlab{}.
\newblock \showarticletitle{Trust region policy optimization}. In
  \bibinfo{booktitle}{\emph{International conference on machine learning}}.
  PMLR, \bibinfo{pages}{1889--1897}.
\newblock


\bibitem[\protect\citeauthoryear{Science and
  on~Artificial~Intelligence}{Science and on~Artificial~Intelligence}{2019}]%
        {national2019national}
\bibfield{author}{\bibinfo{person}{National Science} {and}
  \bibinfo{person}{Technology Council (US). Select~Committee on
  Artificial~Intelligence}.} \bibinfo{year}{2019}\natexlab{}.
\newblock \bibinfo{booktitle}{\emph{The national artificial intelligence
  research and development strategic plan: 2019 update}}.
\newblock \bibinfo{publisher}{National Science and Technology Council (US),
  Select Committee on Artificial~…}.
\newblock


\bibitem[\protect\citeauthoryear{Seo, Lee, Clavera~Gilaberte, Kurutach, Shin,
  and Abbeel}{Seo et~al\mbox{.}}{2020}]%
        {seo2020trajectory}
\bibfield{author}{\bibinfo{person}{Younggyo Seo}, \bibinfo{person}{Kimin Lee},
  \bibinfo{person}{Ignasi Clavera~Gilaberte}, \bibinfo{person}{Thanard
  Kurutach}, \bibinfo{person}{Jinwoo Shin}, {and} \bibinfo{person}{Pieter
  Abbeel}.} \bibinfo{year}{2020}\natexlab{}.
\newblock \showarticletitle{Trajectory-wise multiple choice learning for
  dynamics generalization in reinforcement learning}.
\newblock \bibinfo{journal}{\emph{Advances in Neural Information Processing
  Systems}}  \bibinfo{volume}{33} (\bibinfo{year}{2020}),
  \bibinfo{pages}{12968--12979}.
\newblock


\bibitem[\protect\citeauthoryear{Shen, Li, Jiang, Wang, and Zhao}{Shen
  et~al\mbox{.}}{2020}]%
        {shen2020deep}
\bibfield{author}{\bibinfo{person}{Qianli Shen}, \bibinfo{person}{Yan Li},
  \bibinfo{person}{Haoming Jiang}, \bibinfo{person}{Zhaoran Wang}, {and}
  \bibinfo{person}{Tuo Zhao}.} \bibinfo{year}{2020}\natexlab{}.
\newblock \showarticletitle{Deep Reinforcement Learning with Robust and Smooth
  Policy}. In \bibinfo{booktitle}{\emph{International Conference on Machine
  Learning}}. PMLR, \bibinfo{pages}{8707--8718}.
\newblock


\bibitem[\protect\citeauthoryear{Singh, Gupta, and Shroff}{Singh
  et~al\mbox{.}}{2020}]%
        {singh2020learning}
\bibfield{author}{\bibinfo{person}{Rahul Singh}, \bibinfo{person}{Abhishek
  Gupta}, {and} \bibinfo{person}{Ness~B Shroff}.}
  \bibinfo{year}{2020}\natexlab{}.
\newblock \showarticletitle{Learning in Markov decision processes under
  constraints}.
\newblock \bibinfo{journal}{\emph{arXiv preprint arXiv:2002.12435}}
  (\bibinfo{year}{2020}).
\newblock


\bibitem[\protect\citeauthoryear{Sinha, Namkoong, Volpi, and Duchi}{Sinha
  et~al\mbox{.}}{2017}]%
        {sinha2017certifying}
\bibfield{author}{\bibinfo{person}{Aman Sinha}, \bibinfo{person}{Hongseok
  Namkoong}, \bibinfo{person}{Riccardo Volpi}, {and} \bibinfo{person}{John
  Duchi}.} \bibinfo{year}{2017}\natexlab{}.
\newblock \showarticletitle{Certifying some distributional robustness with
  principled adversarial training}.
\newblock \bibinfo{journal}{\emph{arXiv preprint arXiv:1710.10571}}
  (\bibinfo{year}{2017}).
\newblock


\bibitem[\protect\citeauthoryear{Sinha, O’Kelly, Zheng, Mangharam, Duchi, and
  Tedrake}{Sinha et~al\mbox{.}}{2020}]%
        {sinha2020formulazero}
\bibfield{author}{\bibinfo{person}{Aman Sinha}, \bibinfo{person}{Matthew
  O’Kelly}, \bibinfo{person}{Hongrui Zheng}, \bibinfo{person}{Rahul
  Mangharam}, \bibinfo{person}{John Duchi}, {and} \bibinfo{person}{Russ
  Tedrake}.} \bibinfo{year}{2020}\natexlab{}.
\newblock \showarticletitle{Formulazero: Distributionally robust online
  adaptation via offline population synthesis}. In
  \bibinfo{booktitle}{\emph{International Conference on Machine Learning}}.
  PMLR, \bibinfo{pages}{8992--9004}.
\newblock


\bibitem[\protect\citeauthoryear{Smirnova, Dohmatob, and Mary}{Smirnova
  et~al\mbox{.}}{2019}]%
        {smirnova2019distributionally}
\bibfield{author}{\bibinfo{person}{Elena Smirnova}, \bibinfo{person}{Elvis
  Dohmatob}, {and} \bibinfo{person}{J{\'e}r{\'e}mie Mary}.}
  \bibinfo{year}{2019}\natexlab{}.
\newblock \showarticletitle{Distributionally robust reinforcement learning}.
\newblock \bibinfo{journal}{\emph{arXiv preprint arXiv:1902.08708}}
  (\bibinfo{year}{2019}).
\newblock


\bibitem[\protect\citeauthoryear{Sootla, Cowen-Rivers, Jafferjee, Wang, Mguni,
  Wang, and Ammar}{Sootla et~al\mbox{.}}{2022}]%
        {sootla2022saute}
\bibfield{author}{\bibinfo{person}{Aivar Sootla}, \bibinfo{person}{Alexander~I
  Cowen-Rivers}, \bibinfo{person}{Taher Jafferjee}, \bibinfo{person}{Ziyan
  Wang}, \bibinfo{person}{David~H Mguni}, \bibinfo{person}{Jun Wang}, {and}
  \bibinfo{person}{Haitham Ammar}.} \bibinfo{year}{2022}\natexlab{}.
\newblock \showarticletitle{Saut{\'e} RL: Almost surely safe reinforcement
  learning using state augmentation}. In
  \bibinfo{booktitle}{\emph{International Conference on Machine Learning}}.
  PMLR, \bibinfo{pages}{20423--20443}.
\newblock


\bibitem[\protect\citeauthoryear{Srinivasan, Eysenbach, Ha, Tan, and
  Finn}{Srinivasan et~al\mbox{.}}{2020}]%
        {srinivasan2020learning}
\bibfield{author}{\bibinfo{person}{Krishnan Srinivasan},
  \bibinfo{person}{Benjamin Eysenbach}, \bibinfo{person}{Sehoon Ha},
  \bibinfo{person}{Jie Tan}, {and} \bibinfo{person}{Chelsea Finn}.}
  \bibinfo{year}{2020}\natexlab{}.
\newblock \showarticletitle{Learning to be safe: Deep rl with a safety critic}.
\newblock \bibinfo{journal}{\emph{arXiv preprint arXiv:2010.14603}}
  (\bibinfo{year}{2020}).
\newblock


\bibitem[\protect\citeauthoryear{Stahl~II and Wilson}{Stahl~II and
  Wilson}{1994}]%
        {stahl1994experimental}
\bibfield{author}{\bibinfo{person}{Dale~O Stahl~II} {and}
  \bibinfo{person}{Paul~W Wilson}.} \bibinfo{year}{1994}\natexlab{}.
\newblock \showarticletitle{Experimental evidence on players' models of other
  players}.
\newblock \bibinfo{journal}{\emph{Journal of economic behavior \&
  organization}} \bibinfo{volume}{25}, \bibinfo{number}{3}
  (\bibinfo{year}{1994}), \bibinfo{pages}{309--327}.
\newblock


\bibitem[\protect\citeauthoryear{Stooke, Achiam, and Abbeel}{Stooke
  et~al\mbox{.}}{2020}]%
        {stooke2020responsive}
\bibfield{author}{\bibinfo{person}{Adam Stooke}, \bibinfo{person}{Joshua
  Achiam}, {and} \bibinfo{person}{Pieter Abbeel}.}
  \bibinfo{year}{2020}\natexlab{}.
\newblock \showarticletitle{Responsive safety in reinforcement learning by pid
  lagrangian methods}. In \bibinfo{booktitle}{\emph{International Conference on
  Machine Learning}}. PMLR, \bibinfo{pages}{9133--9143}.
\newblock


\bibitem[\protect\citeauthoryear{Sun, Zheng, Liang, and Huang}{Sun
  et~al\mbox{.}}{2022}]%
        {sun2022who}
\bibfield{author}{\bibinfo{person}{Yanchao Sun}, \bibinfo{person}{Ruijie
  Zheng}, \bibinfo{person}{Yongyuan Liang}, {and} \bibinfo{person}{Furong
  Huang}.} \bibinfo{year}{2022}\natexlab{}.
\newblock \showarticletitle{Who Is the Strongest Enemy? Towards Optimal and
  Efficient Evasion Attacks in Deep {RL}}. In
  \bibinfo{booktitle}{\emph{International Conference on Learning
  Representations}}.
\newblock
\urldef\tempurl%
\url{https://openreview.net/forum?id=JM2kFbJvvI}
\showURL{%
\tempurl}


\bibitem[\protect\citeauthoryear{Sutton and Barto}{Sutton and Barto}{1998}]%
        {DBLP:books/lib/SuttonB98}
\bibfield{author}{\bibinfo{person}{Richard~S. Sutton} {and}
  \bibinfo{person}{Andrew~G. Barto}.} \bibinfo{year}{1998}\natexlab{}.
\newblock \bibinfo{booktitle}{\emph{Reinforcement learning - an introduction}}.
\newblock \bibinfo{publisher}{{MIT} Press}.
\newblock
\showISBNx{978-0-262-19398-6}
\urldef\tempurl%
\url{https://www.worldcat.org/oclc/37293240}
\showURL{%
\tempurl}


\bibitem[\protect\citeauthoryear{Szegedy, Zaremba, Sutskever, Bruna, Erhan,
  Goodfellow, and Fergus}{Szegedy et~al\mbox{.}}{2013}]%
        {szegedy2013intriguing}
\bibfield{author}{\bibinfo{person}{Christian Szegedy},
  \bibinfo{person}{Wojciech Zaremba}, \bibinfo{person}{Ilya Sutskever},
  \bibinfo{person}{Joan Bruna}, \bibinfo{person}{Dumitru Erhan},
  \bibinfo{person}{Ian Goodfellow}, {and} \bibinfo{person}{Rob Fergus}.}
  \bibinfo{year}{2013}\natexlab{}.
\newblock \showarticletitle{Intriguing properties of neural networks}.
\newblock \bibinfo{journal}{\emph{arXiv preprint arXiv:1312.6199}}
  (\bibinfo{year}{2013}).
\newblock


\bibitem[\protect\citeauthoryear{Tabrez and Hayes}{Tabrez and Hayes}{2019}]%
        {tabrez2019improving}
\bibfield{author}{\bibinfo{person}{Aaquib Tabrez} {and}
  \bibinfo{person}{Bradley Hayes}.} \bibinfo{year}{2019}\natexlab{}.
\newblock \showarticletitle{Improving human-robot interaction through
  explainable reinforcement learning}. In \bibinfo{booktitle}{\emph{2019 14th
  ACM/IEEE International Conference on Human-Robot Interaction (HRI)}}. IEEE,
  \bibinfo{pages}{751--753}.
\newblock


\bibitem[\protect\citeauthoryear{Tamar, Glassner, and Mannor}{Tamar
  et~al\mbox{.}}{2015}]%
        {tamar2015optimizing}
\bibfield{author}{\bibinfo{person}{Aviv Tamar}, \bibinfo{person}{Yonatan
  Glassner}, {and} \bibinfo{person}{Shie Mannor}.}
  \bibinfo{year}{2015}\natexlab{}.
\newblock \showarticletitle{Optimizing the CVaR via sampling}. In
  \bibinfo{booktitle}{\emph{Proceedings of the AAAI Conference on Artificial
  Intelligence}}, Vol.~\bibinfo{volume}{29}.
\newblock


\bibitem[\protect\citeauthoryear{Tan, Zhang, Coumans, Iscen, Bai, Hafner,
  Bohez, and Vanhoucke}{Tan et~al\mbox{.}}{2018}]%
        {tan2018sim}
\bibfield{author}{\bibinfo{person}{Jie Tan}, \bibinfo{person}{Tingnan Zhang},
  \bibinfo{person}{Erwin Coumans}, \bibinfo{person}{Atil Iscen},
  \bibinfo{person}{Yunfei Bai}, \bibinfo{person}{Danijar Hafner},
  \bibinfo{person}{Steven Bohez}, {and} \bibinfo{person}{Vincent Vanhoucke}.}
  \bibinfo{year}{2018}\natexlab{}.
\newblock \showarticletitle{Sim-to-real: Learning agile locomotion for
  quadruped robots}.
\newblock \bibinfo{journal}{\emph{arXiv preprint arXiv:1804.10332}}
  (\bibinfo{year}{2018}).
\newblock


\bibitem[\protect\citeauthoryear{Tessler, Efroni, and Mannor}{Tessler
  et~al\mbox{.}}{2019}]%
        {tessler2019action}
\bibfield{author}{\bibinfo{person}{Chen Tessler}, \bibinfo{person}{Yonathan
  Efroni}, {and} \bibinfo{person}{Shie Mannor}.}
  \bibinfo{year}{2019}\natexlab{}.
\newblock \showarticletitle{Action robust reinforcement learning and
  applications in continuous control}. In
  \bibinfo{booktitle}{\emph{International Conference on Machine Learning}}.
  PMLR, \bibinfo{pages}{6215--6224}.
\newblock


\bibitem[\protect\citeauthoryear{Tessler, Mankowitz, and Mannor}{Tessler
  et~al\mbox{.}}{2018}]%
        {tessler2018reward}
\bibfield{author}{\bibinfo{person}{Chen Tessler}, \bibinfo{person}{Daniel~J
  Mankowitz}, {and} \bibinfo{person}{Shie Mannor}.}
  \bibinfo{year}{2018}\natexlab{}.
\newblock \showarticletitle{Reward constrained policy optimization}.
\newblock \bibinfo{journal}{\emph{arXiv preprint arXiv:1805.11074}}
  (\bibinfo{year}{2018}).
\newblock


\bibitem[\protect\citeauthoryear{Thananjeyan, Balakrishna, Nair, Luo,
  Srinivasan, Hwang, Gonzalez, Ibarz, Finn, and Goldberg}{Thananjeyan
  et~al\mbox{.}}{2021}]%
        {thananjeyan2021recovery}
\bibfield{author}{\bibinfo{person}{Brijen Thananjeyan}, \bibinfo{person}{Ashwin
  Balakrishna}, \bibinfo{person}{Suraj Nair}, \bibinfo{person}{Michael Luo},
  \bibinfo{person}{Krishnan Srinivasan}, \bibinfo{person}{Minho Hwang},
  \bibinfo{person}{Joseph~E Gonzalez}, \bibinfo{person}{Julian Ibarz},
  \bibinfo{person}{Chelsea Finn}, {and} \bibinfo{person}{Ken Goldberg}.}
  \bibinfo{year}{2021}\natexlab{}.
\newblock \showarticletitle{Recovery rl: Safe reinforcement learning with
  learned recovery zones}.
\newblock \bibinfo{journal}{\emph{IEEE Robotics and Automation Letters}}
  \bibinfo{volume}{6}, \bibinfo{number}{3} (\bibinfo{year}{2021}),
  \bibinfo{pages}{4915--4922}.
\newblock


\bibitem[\protect\citeauthoryear{Thananjeyan, Balakrishna, Rosolia, Li,
  McAllister, Gonzalez, Levine, Borrelli, and Goldberg}{Thananjeyan
  et~al\mbox{.}}{2020}]%
        {thananjeyan2020safety}
\bibfield{author}{\bibinfo{person}{Brijen Thananjeyan}, \bibinfo{person}{Ashwin
  Balakrishna}, \bibinfo{person}{Ugo Rosolia}, \bibinfo{person}{Felix Li},
  \bibinfo{person}{Rowan McAllister}, \bibinfo{person}{Joseph~E Gonzalez},
  \bibinfo{person}{Sergey Levine}, \bibinfo{person}{Francesco Borrelli}, {and}
  \bibinfo{person}{Ken Goldberg}.} \bibinfo{year}{2020}\natexlab{}.
\newblock \showarticletitle{Safety augmented value estimation from
  demonstrations (saved): Safe deep model-based rl for sparse cost robotic
  tasks}.
\newblock \bibinfo{journal}{\emph{IEEE Robotics and Automation Letters}}
  \bibinfo{volume}{5}, \bibinfo{number}{2} (\bibinfo{year}{2020}),
  \bibinfo{pages}{3612--3619}.
\newblock


\bibitem[\protect\citeauthoryear{Thrun and Mitchell}{Thrun and
  Mitchell}{1995}]%
        {thrun1995lifelong}
\bibfield{author}{\bibinfo{person}{Sebastian Thrun} {and}
  \bibinfo{person}{Tom~M Mitchell}.} \bibinfo{year}{1995}\natexlab{}.
\newblock \showarticletitle{Lifelong robot learning}.
\newblock \bibinfo{journal}{\emph{Robotics and autonomous systems}}
  \bibinfo{volume}{15}, \bibinfo{number}{1-2} (\bibinfo{year}{1995}),
  \bibinfo{pages}{25--46}.
\newblock


\bibitem[\protect\citeauthoryear{Tobin, Fong, Ray, Schneider, Zaremba, and
  Abbeel}{Tobin et~al\mbox{.}}{2017}]%
        {tobin2017domain}
\bibfield{author}{\bibinfo{person}{Josh Tobin}, \bibinfo{person}{Rachel Fong},
  \bibinfo{person}{Alex Ray}, \bibinfo{person}{Jonas Schneider},
  \bibinfo{person}{Wojciech Zaremba}, {and} \bibinfo{person}{Pieter Abbeel}.}
  \bibinfo{year}{2017}\natexlab{}.
\newblock \showarticletitle{Domain randomization for transferring deep neural
  networks from simulation to the real world}. In
  \bibinfo{booktitle}{\emph{2017 IEEE/RSJ international conference on
  intelligent robots and systems (IROS)}}. IEEE, \bibinfo{pages}{23--30}.
\newblock


\bibitem[\protect\citeauthoryear{Todorov, Erez, and Tassa}{Todorov
  et~al\mbox{.}}{2012}]%
        {todorov2012mujoco}
\bibfield{author}{\bibinfo{person}{Emanuel Todorov}, \bibinfo{person}{Tom
  Erez}, {and} \bibinfo{person}{Yuval Tassa}.} \bibinfo{year}{2012}\natexlab{}.
\newblock \showarticletitle{Mujoco: A physics engine for model-based control}.
  In \bibinfo{booktitle}{\emph{2012 IEEE/RSJ International Conference on
  Intelligent Robots and Systems}}. IEEE, \bibinfo{pages}{5026--5033}.
\newblock


\bibitem[\protect\citeauthoryear{Tretschk, Oh, and Fritz}{Tretschk
  et~al\mbox{.}}{2018}]%
        {tretschk2018sequential}
\bibfield{author}{\bibinfo{person}{Edgar Tretschk}, \bibinfo{person}{Seong~Joon
  Oh}, {and} \bibinfo{person}{Mario Fritz}.} \bibinfo{year}{2018}\natexlab{}.
\newblock \showarticletitle{Sequential attacks on agents for long-term
  adversarial goals}.
\newblock \bibinfo{journal}{\emph{arXiv preprint arXiv:1805.12487}}
  (\bibinfo{year}{2018}).
\newblock


\bibitem[\protect\citeauthoryear{Turchetta, Kolobov, Shah, Krause, and
  Agarwal}{Turchetta et~al\mbox{.}}{2020}]%
        {turchetta2020safe}
\bibfield{author}{\bibinfo{person}{Matteo Turchetta}, \bibinfo{person}{Andrey
  Kolobov}, \bibinfo{person}{Shital Shah}, \bibinfo{person}{Andreas Krause},
  {and} \bibinfo{person}{Alekh Agarwal}.} \bibinfo{year}{2020}\natexlab{}.
\newblock \showarticletitle{Safe reinforcement learning via curriculum
  induction}.
\newblock \bibinfo{journal}{\emph{arXiv preprint arXiv:2006.12136}}
  (\bibinfo{year}{2020}).
\newblock


\bibitem[\protect\citeauthoryear{van~der Pol, Worrall, van Hoof, Oliehoek, and
  Welling}{van~der Pol et~al\mbox{.}}{2020}]%
        {van2020mdp}
\bibfield{author}{\bibinfo{person}{Elise van~der Pol}, \bibinfo{person}{Daniel
  Worrall}, \bibinfo{person}{Herke van Hoof}, \bibinfo{person}{Frans Oliehoek},
  {and} \bibinfo{person}{Max Welling}.} \bibinfo{year}{2020}\natexlab{}.
\newblock \showarticletitle{MDP homomorphic networks: Group symmetries in
  reinforcement learning}.
\newblock \bibinfo{journal}{\emph{Advances in Neural Information Processing
  Systems}}  \bibinfo{volume}{33} (\bibinfo{year}{2020}),
  \bibinfo{pages}{4199--4210}.
\newblock


\bibitem[\protect\citeauthoryear{Vithayathil~Varghese and
  Mahmoud}{Vithayathil~Varghese and Mahmoud}{2020}]%
        {vithayathil2020survey}
\bibfield{author}{\bibinfo{person}{Nelson Vithayathil~Varghese} {and}
  \bibinfo{person}{Qusay~H Mahmoud}.} \bibinfo{year}{2020}\natexlab{}.
\newblock \showarticletitle{A survey of multi-task deep reinforcement
  learning}.
\newblock \bibinfo{journal}{\emph{Electronics}} \bibinfo{volume}{9},
  \bibinfo{number}{9} (\bibinfo{year}{2020}), \bibinfo{pages}{1363}.
\newblock


\bibitem[\protect\citeauthoryear{Wang, Liu, and Li}{Wang et~al\mbox{.}}{2020}]%
        {wang2020reinforcement}
\bibfield{author}{\bibinfo{person}{Jingkang Wang}, \bibinfo{person}{Yang Liu},
  {and} \bibinfo{person}{Bo Li}.} \bibinfo{year}{2020}\natexlab{}.
\newblock \showarticletitle{Reinforcement learning with perturbed rewards}. In
  \bibinfo{booktitle}{\emph{Proceedings of the AAAI Conference on Artificial
  Intelligence}}, Vol.~\bibinfo{volume}{34}. \bibinfo{pages}{6202--6209}.
\newblock


\bibitem[\protect\citeauthoryear{Wang, Weng, and Daniel}{Wang
  et~al\mbox{.}}{2019}]%
        {wang2019verification}
\bibfield{author}{\bibinfo{person}{Yuh-Shyang Wang}, \bibinfo{person}{Tsui-Wei
  Weng}, {and} \bibinfo{person}{Luca Daniel}.} \bibinfo{year}{2019}\natexlab{}.
\newblock \showarticletitle{Verification of neural network control policy under
  persistent adversarial perturbation}.
\newblock \bibinfo{journal}{\emph{arXiv preprint arXiv:1908.06353}}
  (\bibinfo{year}{2019}).
\newblock


\bibitem[\protect\citeauthoryear{Wawrzy{\'n}ski}{Wawrzy{\'n}ski}{2009}]%
        {wawrzynski2009real}
\bibfield{author}{\bibinfo{person}{Pawe{\l} Wawrzy{\'n}ski}.}
  \bibinfo{year}{2009}\natexlab{}.
\newblock \showarticletitle{Real-time reinforcement learning by sequential
  actor--critics and experience replay}.
\newblock \bibinfo{journal}{\emph{Neural networks}} \bibinfo{volume}{22},
  \bibinfo{number}{10} (\bibinfo{year}{2009}), \bibinfo{pages}{1484--1497}.
\newblock


\bibitem[\protect\citeauthoryear{Wells and Bednarz}{Wells and Bednarz}{2021}]%
        {wells2021explainable}
\bibfield{author}{\bibinfo{person}{Lindsay Wells} {and} \bibinfo{person}{Tomasz
  Bednarz}.} \bibinfo{year}{2021}\natexlab{}.
\newblock \showarticletitle{Explainable ai and reinforcement learning—a
  systematic review of current approaches and trends}.
\newblock \bibinfo{journal}{\emph{Frontiers in artificial intelligence}}
  \bibinfo{volume}{4} (\bibinfo{year}{2021}), \bibinfo{pages}{550030}.
\newblock


\bibitem[\protect\citeauthoryear{Weng, Zhang, Chen, Song, Hsieh, Daniel,
  Boning, and Dhillon}{Weng et~al\mbox{.}}{2018}]%
        {weng2018towards}
\bibfield{author}{\bibinfo{person}{Lily Weng}, \bibinfo{person}{Huan Zhang},
  \bibinfo{person}{Hongge Chen}, \bibinfo{person}{Zhao Song},
  \bibinfo{person}{Cho-Jui Hsieh}, \bibinfo{person}{Luca Daniel},
  \bibinfo{person}{Duane Boning}, {and} \bibinfo{person}{Inderjit Dhillon}.}
  \bibinfo{year}{2018}\natexlab{}.
\newblock \showarticletitle{Towards fast computation of certified robustness
  for relu networks}. In \bibinfo{booktitle}{\emph{International Conference on
  Machine Learning}}. PMLR, \bibinfo{pages}{5276--5285}.
\newblock


\bibitem[\protect\citeauthoryear{Wiesemann, Kuhn, and Rustem}{Wiesemann
  et~al\mbox{.}}{2013}]%
        {wiesemann2013robust}
\bibfield{author}{\bibinfo{person}{Wolfram Wiesemann}, \bibinfo{person}{Daniel
  Kuhn}, {and} \bibinfo{person}{Ber{\c{c}} Rustem}.}
  \bibinfo{year}{2013}\natexlab{}.
\newblock \showarticletitle{Robust Markov decision processes}.
\newblock \bibinfo{journal}{\emph{Mathematics of Operations Research}}
  \bibinfo{volume}{38}, \bibinfo{number}{1} (\bibinfo{year}{2013}),
  \bibinfo{pages}{153--183}.
\newblock


\bibitem[\protect\citeauthoryear{Wilson, Fern, Ray, and Tadepalli}{Wilson
  et~al\mbox{.}}{2007}]%
        {wilson2007multi}
\bibfield{author}{\bibinfo{person}{Aaron Wilson}, \bibinfo{person}{Alan Fern},
  \bibinfo{person}{Soumya Ray}, {and} \bibinfo{person}{Prasad Tadepalli}.}
  \bibinfo{year}{2007}\natexlab{}.
\newblock \showarticletitle{Multi-task reinforcement learning: a hierarchical
  bayesian approach}. In \bibinfo{booktitle}{\emph{Proceedings of the 24th
  international conference on Machine learning}}. \bibinfo{pages}{1015--1022}.
\newblock


\bibitem[\protect\citeauthoryear{Wu, Li, Huang, Vorobeychik, Zhao, and Li}{Wu
  et~al\mbox{.}}{2021}]%
        {wu2021crop}
\bibfield{author}{\bibinfo{person}{Fan Wu}, \bibinfo{person}{Linyi Li},
  \bibinfo{person}{Zijian Huang}, \bibinfo{person}{Yevgeniy Vorobeychik},
  \bibinfo{person}{Ding Zhao}, {and} \bibinfo{person}{Bo Li}.}
  \bibinfo{year}{2021}\natexlab{}.
\newblock \bibinfo{journal}{\emph{arXiv preprint arXiv:2106.09292}}
  (\bibinfo{year}{2021}).
\newblock


\bibitem[\protect\citeauthoryear{Wu, Li, Huang, Vorobeychik, Zhao, and Li}{Wu
  et~al\mbox{.}}{2022a}]%
        {wu2022crop}
\bibfield{author}{\bibinfo{person}{Fan Wu}, \bibinfo{person}{Linyi Li},
  \bibinfo{person}{Zijian Huang}, \bibinfo{person}{Yevgeniy Vorobeychik},
  \bibinfo{person}{Ding Zhao}, {and} \bibinfo{person}{Bo Li}.}
  \bibinfo{year}{2022}\natexlab{a}.
\newblock \showarticletitle{CROP: Certifying Robust Policies for Reinforcement
  Learning through Functional Smoothing}. In
  \bibinfo{booktitle}{\emph{International Conference on Learning
  Representations}}.
\newblock


\bibitem[\protect\citeauthoryear{Wu, Li, Xu, Zhang, Kailkhura, Kenthapadi,
  Zhao, and Li}{Wu et~al\mbox{.}}{2022b}]%
        {wu2022copa}
\bibfield{author}{\bibinfo{person}{Fan Wu}, \bibinfo{person}{Linyi Li},
  \bibinfo{person}{Chejian Xu}, \bibinfo{person}{Huan Zhang},
  \bibinfo{person}{Bhavya Kailkhura}, \bibinfo{person}{Krishnaram Kenthapadi},
  \bibinfo{person}{Ding Zhao}, {and} \bibinfo{person}{Bo Li}.}
  \bibinfo{year}{2022}\natexlab{b}.
\newblock \showarticletitle{COPA: Certifying Robust Policies for Offline
  Reinforcement Learning against Poisoning Attacks}.
\newblock \bibinfo{journal}{\emph{arXiv preprint arXiv:2203.08398}}
  (\bibinfo{year}{2022}).
\newblock


\bibitem[\protect\citeauthoryear{Xu, Ding, Lyu, Liu, Wang, He, Hu, Zhao, and
  Li}{Xu et~al\mbox{.}}{2022}]%
        {xu2022safebench}
\bibfield{author}{\bibinfo{person}{Chejian Xu}, \bibinfo{person}{Wenhao Ding},
  \bibinfo{person}{Weijie Lyu}, \bibinfo{person}{Zuxin Liu},
  \bibinfo{person}{Shuai Wang}, \bibinfo{person}{Yihan He},
  \bibinfo{person}{Hanjiang Hu}, \bibinfo{person}{Ding Zhao}, {and}
  \bibinfo{person}{Bo Li}.} \bibinfo{year}{2022}\natexlab{}.
\newblock \showarticletitle{SafeBench: A Benchmarking Platform for Safety
  Evaluation of Autonomous Vehicles}.
\newblock \bibinfo{journal}{\emph{arXiv preprint arXiv:2206.09682}}
  (\bibinfo{year}{2022}).
\newblock


\bibitem[\protect\citeauthoryear{Xu and Mannor}{Xu and Mannor}{2006}]%
        {xu2006robustness}
\bibfield{author}{\bibinfo{person}{Huan Xu} {and} \bibinfo{person}{Shie
  Mannor}.} \bibinfo{year}{2006}\natexlab{}.
\newblock \showarticletitle{The robustness-performance tradeoff in Markov
  decision processes}.
\newblock \bibinfo{journal}{\emph{Advances in Neural Information Processing
  Systems}}  \bibinfo{volume}{19} (\bibinfo{year}{2006}).
\newblock


\bibitem[\protect\citeauthoryear{Xu and Mannor}{Xu and Mannor}{2010}]%
        {xu2010distributionally}
\bibfield{author}{\bibinfo{person}{Huan Xu} {and} \bibinfo{person}{Shie
  Mannor}.} \bibinfo{year}{2010}\natexlab{}.
\newblock \showarticletitle{Distributionally Robust Markov Decision
  Processes.}. In \bibinfo{booktitle}{\emph{NIPS}}.
  \bibinfo{pages}{2505--2513}.
\newblock


\bibitem[\protect\citeauthoryear{Xu and Mannor}{Xu and Mannor}{2012}]%
        {xu2012robustness}
\bibfield{author}{\bibinfo{person}{Huan Xu} {and} \bibinfo{person}{Shie
  Mannor}.} \bibinfo{year}{2012}\natexlab{}.
\newblock \showarticletitle{Robustness and generalization}.
\newblock \bibinfo{journal}{\emph{Machine learning}} \bibinfo{volume}{86},
  \bibinfo{number}{3} (\bibinfo{year}{2012}), \bibinfo{pages}{391--423}.
\newblock


\bibitem[\protect\citeauthoryear{Xu, Ding, Zhu, Liu, Chen, and Zhao}{Xu
  et~al\mbox{.}}{2020}]%
        {xu2020task}
\bibfield{author}{\bibinfo{person}{Mengdi Xu}, \bibinfo{person}{Wenhao Ding},
  \bibinfo{person}{Jiacheng Zhu}, \bibinfo{person}{Zuxin Liu},
  \bibinfo{person}{Baiming Chen}, {and} \bibinfo{person}{Ding Zhao}.}
  \bibinfo{year}{2020}\natexlab{}.
\newblock \showarticletitle{Task-agnostic online reinforcement learning with an
  infinite mixture of gaussian processes}.
\newblock \bibinfo{journal}{\emph{arXiv preprint arXiv:2006.11441}}
  (\bibinfo{year}{2020}).
\newblock


\bibitem[\protect\citeauthoryear{Xu, Huang, Li, Zhu, Qi, Oguchi, Huang, Lam,
  and Zhao}{Xu et~al\mbox{.}}{2021}]%
        {xu2021accelerated}
\bibfield{author}{\bibinfo{person}{Mengdi Xu}, \bibinfo{person}{Peide Huang},
  \bibinfo{person}{Fengpei Li}, \bibinfo{person}{Jiacheng Zhu},
  \bibinfo{person}{Xuewei Qi}, \bibinfo{person}{Kentaro Oguchi},
  \bibinfo{person}{Zhiyuan Huang}, \bibinfo{person}{Henry Lam}, {and}
  \bibinfo{person}{Ding Zhao}.} \bibinfo{year}{2021}\natexlab{}.
\newblock \showarticletitle{Accelerated Policy Evaluation: Learning Adversarial
  Environments with Adaptive Importance Sampling}.
\newblock \bibinfo{journal}{\emph{arXiv preprint arXiv:2106.10566}}
  (\bibinfo{year}{2021}).
\newblock


\bibitem[\protect\citeauthoryear{Xue, Pinto, Gamage, Nikonova, Zhang, and
  Renz}{Xue et~al\mbox{.}}{2021}]%
        {xue2021phy}
\bibfield{author}{\bibinfo{person}{Cheng Xue}, \bibinfo{person}{Vimukthini
  Pinto}, \bibinfo{person}{Chathura Gamage}, \bibinfo{person}{Ekaterina
  Nikonova}, \bibinfo{person}{Peng Zhang}, {and} \bibinfo{person}{Jochen
  Renz}.} \bibinfo{year}{2021}\natexlab{}.
\newblock \showarticletitle{Phy-q: A benchmark for physical reasoning}.
\newblock \bibinfo{journal}{\emph{arXiv preprint arXiv:2108.13696}}
  (\bibinfo{year}{2021}).
\newblock


\bibitem[\protect\citeauthoryear{Yan, Guo, and Zhang}{Yan
  et~al\mbox{.}}{2018}]%
        {yan2018deep}
\bibfield{author}{\bibinfo{person}{Ziang Yan}, \bibinfo{person}{Yiwen Guo},
  {and} \bibinfo{person}{Changshui Zhang}.} \bibinfo{year}{2018}\natexlab{}.
\newblock \showarticletitle{Deep Defense: Training DNNs with Improved
  Adversarial Robustness}.
\newblock \bibinfo{journal}{\emph{Advances in Neural Information Processing
  Systems}}  \bibinfo{volume}{31} (\bibinfo{year}{2018}),
  \bibinfo{pages}{419--428}.
\newblock


\bibitem[\protect\citeauthoryear{Yang}{Yang}{2017}]%
        {yang2017convex}
\bibfield{author}{\bibinfo{person}{Insoon Yang}.}
  \bibinfo{year}{2017}\natexlab{}.
\newblock \showarticletitle{A convex optimization approach to distributionally
  robust Markov decision processes with Wasserstein distance}.
\newblock \bibinfo{journal}{\emph{IEEE control systems letters}}
  \bibinfo{volume}{1}, \bibinfo{number}{1} (\bibinfo{year}{2017}),
  \bibinfo{pages}{164--169}.
\newblock


\bibitem[\protect\citeauthoryear{Yang, Sim{\~a}o, Tindemans, and Spaan}{Yang
  et~al\mbox{.}}{2021b}]%
        {yang2021wcsac}
\bibfield{author}{\bibinfo{person}{Qisong Yang}, \bibinfo{person}{Thiago~D
  Sim{\~a}o}, \bibinfo{person}{Simon~H Tindemans}, {and}
  \bibinfo{person}{Matthijs~TJ Spaan}.} \bibinfo{year}{2021}\natexlab{b}.
\newblock \showarticletitle{WCSAC: Worst-Case Soft Actor Critic for
  Safety-Constrained Reinforcement Learning.}. In
  \bibinfo{booktitle}{\emph{AAAI}}. \bibinfo{pages}{10639--10646}.
\newblock


\bibitem[\protect\citeauthoryear{Yang, Rosca, Narasimhan, and Ramadge}{Yang
  et~al\mbox{.}}{2020}]%
        {yang2020projection}
\bibfield{author}{\bibinfo{person}{Tsung-Yen Yang}, \bibinfo{person}{Justinian
  Rosca}, \bibinfo{person}{Karthik Narasimhan}, {and} \bibinfo{person}{Peter~J
  Ramadge}.} \bibinfo{year}{2020}\natexlab{}.
\newblock \showarticletitle{Projection-based constrained policy optimization}.
\newblock \bibinfo{journal}{\emph{arXiv preprint arXiv:2010.03152}}
  (\bibinfo{year}{2020}).
\newblock


\bibitem[\protect\citeauthoryear{Yang, Rosca, Narasimhan, and Ramadge}{Yang
  et~al\mbox{.}}{2021a}]%
        {yang2021accelerating}
\bibfield{author}{\bibinfo{person}{Tsung-Yen Yang}, \bibinfo{person}{Justinian
  Rosca}, \bibinfo{person}{Karthik Narasimhan}, {and} \bibinfo{person}{Peter~J
  Ramadge}.} \bibinfo{year}{2021}\natexlab{a}.
\newblock \showarticletitle{Accelerating Safe Reinforcement Learning with
  Constraint-mismatched Baseline Policies}. In
  \bibinfo{booktitle}{\emph{International Conference on Machine Learning}}.
  PMLR, \bibinfo{pages}{11795--11807}.
\newblock


\bibitem[\protect\citeauthoryear{Yu, Liu, Nemati, and Yin}{Yu
  et~al\mbox{.}}{2021}]%
        {10.1145/3477600}
\bibfield{author}{\bibinfo{person}{Chao Yu}, \bibinfo{person}{Jiming Liu},
  \bibinfo{person}{Shamim Nemati}, {and} \bibinfo{person}{Guosheng Yin}.}
  \bibinfo{year}{2021}\natexlab{}.
\newblock \showarticletitle{Reinforcement Learning in Healthcare: A Survey}.
\newblock \bibinfo{journal}{\emph{ACM Comput. Surv.}} \bibinfo{volume}{55},
  \bibinfo{number}{1}, Article \bibinfo{articleno}{5} (\bibinfo{date}{nov}
  \bibinfo{year}{2021}), \bibinfo{numpages}{36}~pages.
\newblock
\showISSN{0360-0300}
\urldef\tempurl%
\url{https://doi.org/10.1145/3477600}
\showDOI{\tempurl}


\bibitem[\protect\citeauthoryear{Yu, Xu, and Zhang}{Yu et~al\mbox{.}}{2022}]%
        {yu2022towards}
\bibfield{author}{\bibinfo{person}{Haonan Yu}, \bibinfo{person}{Wei Xu}, {and}
  \bibinfo{person}{Haichao Zhang}.} \bibinfo{year}{2022}\natexlab{}.
\newblock \showarticletitle{Towards Safe Reinforcement Learning with a Safety
  Editor Policy}.
\newblock \bibinfo{journal}{\emph{arXiv preprint arXiv:2201.12427}}
  (\bibinfo{year}{2022}).
\newblock


\bibitem[\protect\citeauthoryear{Yu, Yang, Kolar, and Wang}{Yu
  et~al\mbox{.}}{2019}]%
        {yu2019convergent}
\bibfield{author}{\bibinfo{person}{Ming Yu}, \bibinfo{person}{Zhuoran Yang},
  \bibinfo{person}{Mladen Kolar}, {and} \bibinfo{person}{Zhaoran Wang}.}
  \bibinfo{year}{2019}\natexlab{}.
\newblock \showarticletitle{Convergent policy optimization for safe
  reinforcement learning}.
\newblock \bibinfo{journal}{\emph{arXiv preprint arXiv:1910.12156}}
  (\bibinfo{year}{2019}).
\newblock


\bibitem[\protect\citeauthoryear{Yu and Xu}{Yu and Xu}{2015}]%
        {yu2015distributionally}
\bibfield{author}{\bibinfo{person}{Pengqian Yu} {and} \bibinfo{person}{Huan
  Xu}.} \bibinfo{year}{2015}\natexlab{}.
\newblock \showarticletitle{Distributionally robust counterpart in Markov
  decision processes}.
\newblock \bibinfo{journal}{\emph{IEEE Trans. Automat. Control}}
  \bibinfo{volume}{61}, \bibinfo{number}{9} (\bibinfo{year}{2015}),
  \bibinfo{pages}{2538--2543}.
\newblock


\bibitem[\protect\citeauthoryear{Yu, Tan, Liu, and Turk}{Yu
  et~al\mbox{.}}{2017}]%
        {yu2017preparing}
\bibfield{author}{\bibinfo{person}{Wenhao Yu}, \bibinfo{person}{Jie Tan},
  \bibinfo{person}{C~Karen Liu}, {and} \bibinfo{person}{Greg Turk}.}
  \bibinfo{year}{2017}\natexlab{}.
\newblock \showarticletitle{Preparing for the unknown: Learning a universal
  policy with online system identification}.
\newblock \bibinfo{journal}{\emph{arXiv preprint arXiv:1702.02453}}
  (\bibinfo{year}{2017}).
\newblock


\bibitem[\protect\citeauthoryear{Yuan, Song, Luo, Sun, and Kitani}{Yuan
  et~al\mbox{.}}{2021b}]%
        {yuan2021transform2act}
\bibfield{author}{\bibinfo{person}{Ye Yuan}, \bibinfo{person}{Yuda Song},
  \bibinfo{person}{Zhengyi Luo}, \bibinfo{person}{Wen Sun}, {and}
  \bibinfo{person}{Kris Kitani}.} \bibinfo{year}{2021}\natexlab{b}.
\newblock \showarticletitle{Transform2Act: Learning a Transform-and-Control
  Policy for Efficient Agent Design}.
\newblock \bibinfo{journal}{\emph{arXiv preprint arXiv:2110.03659}}
  (\bibinfo{year}{2021}).
\newblock


\bibitem[\protect\citeauthoryear{Yuan, Hall, Zhou, Brunke, Greeff, Panerati,
  and Schoellig}{Yuan et~al\mbox{.}}{2021a}]%
        {yuan2021safe}
\bibfield{author}{\bibinfo{person}{Zhaocong Yuan}, \bibinfo{person}{Adam~W
  Hall}, \bibinfo{person}{Siqi Zhou}, \bibinfo{person}{Lukas Brunke},
  \bibinfo{person}{Melissa Greeff}, \bibinfo{person}{Jacopo Panerati}, {and}
  \bibinfo{person}{Angela~P Schoellig}.} \bibinfo{year}{2021}\natexlab{a}.
\newblock \showarticletitle{safe-control-gym: a Unified Benchmark Suite for
  Safe Learning-based Control and Reinforcement Learning}.
\newblock \bibinfo{journal}{\emph{arXiv preprint arXiv:2109.06325}}
  (\bibinfo{year}{2021}).
\newblock


\bibitem[\protect\citeauthoryear{Zhang, Lyle, Sodhani, Filos, Kwiatkowska,
  Pineau, Gal, and Precup}{Zhang et~al\mbox{.}}{2020b}]%
        {zhang2020invariant}
\bibfield{author}{\bibinfo{person}{Amy Zhang}, \bibinfo{person}{Clare Lyle},
  \bibinfo{person}{Shagun Sodhani}, \bibinfo{person}{Angelos Filos},
  \bibinfo{person}{Marta Kwiatkowska}, \bibinfo{person}{Joelle Pineau},
  \bibinfo{person}{Yarin Gal}, {and} \bibinfo{person}{Doina Precup}.}
  \bibinfo{year}{2020}\natexlab{b}.
\newblock \showarticletitle{Invariant causal prediction for block mdps}. In
  \bibinfo{booktitle}{\emph{International Conference on Machine Learning}}.
  PMLR, \bibinfo{pages}{11214--11224}.
\newblock


\bibitem[\protect\citeauthoryear{Zhang, McAllister, Calandra, Gal, and
  Levine}{Zhang et~al\mbox{.}}{2020d}]%
        {zhang2020learning}
\bibfield{author}{\bibinfo{person}{Amy Zhang}, \bibinfo{person}{Rowan
  McAllister}, \bibinfo{person}{Roberto Calandra}, \bibinfo{person}{Yarin Gal},
  {and} \bibinfo{person}{Sergey Levine}.} \bibinfo{year}{2020}\natexlab{d}.
\newblock \showarticletitle{Learning invariant representations for
  reinforcement learning without reconstruction}.
\newblock \bibinfo{journal}{\emph{arXiv preprint arXiv:2006.10742}}
  (\bibinfo{year}{2020}).
\newblock


\bibitem[\protect\citeauthoryear{Zhang, Chen, Boning, and Hsieh}{Zhang
  et~al\mbox{.}}{2021}]%
        {zhang2021atla}
\bibfield{author}{\bibinfo{person}{Huan Zhang}, \bibinfo{person}{Hongge Chen},
  \bibinfo{person}{Duane~S Boning}, {and} \bibinfo{person}{Cho-Jui Hsieh}.}
  \bibinfo{year}{2021}\natexlab{}.
\newblock \showarticletitle{Robust Reinforcement Learning on State Observations
  with Learned Optimal Adversary}. In \bibinfo{booktitle}{\emph{International
  Conference on Learning Representations}}.
\newblock
\urldef\tempurl%
\url{https://openreview.net/forum?id=sCZbhBvqQaU}
\showURL{%
\tempurl}


\bibitem[\protect\citeauthoryear{Zhang, Chen, Xiao, Li, Liu, Boning, and
  Hsieh}{Zhang et~al\mbox{.}}{2020a}]%
        {zhang2021robust}
\bibfield{author}{\bibinfo{person}{Huan Zhang}, \bibinfo{person}{Hongge Chen},
  \bibinfo{person}{Chaowei Xiao}, \bibinfo{person}{Bo Li},
  \bibinfo{person}{Mingyan Liu}, \bibinfo{person}{Duane Boning}, {and}
  \bibinfo{person}{Cho-Jui Hsieh}.} \bibinfo{year}{2020}\natexlab{a}.
\newblock \showarticletitle{Robust Deep Reinforcement Learning against
  Adversarial Perturbations on State Observations}. In
  \bibinfo{booktitle}{\emph{Advances in Neural Information Processing
  Systems}}, \bibfield{editor}{\bibinfo{person}{H.~Larochelle},
  \bibinfo{person}{M.~Ranzato}, \bibinfo{person}{R.~Hadsell},
  \bibinfo{person}{M.~F. Balcan}, {and} \bibinfo{person}{H.~Lin}} (Eds.),
  Vol.~\bibinfo{volume}{33}. \bibinfo{publisher}{Curran Associates, Inc.},
  \bibinfo{pages}{21024--21037}.
\newblock
\urldef\tempurl%
\url{https://proceedings.neurips.cc/paper/2020/file/f0eb6568ea114ba6e293f903c34d7488-Paper.pdf}
\showURL{%
\tempurl}


\bibitem[\protect\citeauthoryear{Zhang and Parkes}{Zhang and Parkes}{2008}]%
        {zhang2008value}
\bibfield{author}{\bibinfo{person}{Haoqi Zhang} {and} \bibinfo{person}{David~C
  Parkes}.} \bibinfo{year}{2008}\natexlab{}.
\newblock \showarticletitle{Value-Based Policy Teaching with Active Indirect
  Elicitation.}. In \bibinfo{booktitle}{\emph{AAAI}}, Vol.~\bibinfo{volume}{8}.
  \bibinfo{pages}{208--214}.
\newblock


\bibitem[\protect\citeauthoryear{Zhang, Ma, Singla, and Zhu}{Zhang
  et~al\mbox{.}}{2020c}]%
        {zhang2020adaptive}
\bibfield{author}{\bibinfo{person}{Xuezhou Zhang}, \bibinfo{person}{Yuzhe Ma},
  \bibinfo{person}{Adish Singla}, {and} \bibinfo{person}{Xiaojin Zhu}.}
  \bibinfo{year}{2020}\natexlab{c}.
\newblock \showarticletitle{Adaptive reward-poisoning attacks against
  reinforcement learning}. In \bibinfo{booktitle}{\emph{International
  Conference on Machine Learning}}. PMLR, \bibinfo{pages}{11225--11234}.
\newblock


\bibitem[\protect\citeauthoryear{Zhang, Vuong, and Ross}{Zhang
  et~al\mbox{.}}{2020e}]%
        {zhang2020first}
\bibfield{author}{\bibinfo{person}{Yiming Zhang}, \bibinfo{person}{Quan Vuong},
  {and} \bibinfo{person}{Keith Ross}.} \bibinfo{year}{2020}\natexlab{e}.
\newblock \showarticletitle{First order constrained optimization in policy
  space}.
\newblock \bibinfo{journal}{\emph{Advances in Neural Information Processing
  Systems}} (\bibinfo{year}{2020}).
\newblock


\bibitem[\protect\citeauthoryear{Zhao, Queralta, and Westerlund}{Zhao
  et~al\mbox{.}}{2020}]%
        {zhao2020sim}
\bibfield{author}{\bibinfo{person}{Wenshuai Zhao},
  \bibinfo{person}{Jorge~Pe{\~n}a Queralta}, {and} \bibinfo{person}{Tomi
  Westerlund}.} \bibinfo{year}{2020}\natexlab{}.
\newblock \showarticletitle{Sim-to-real transfer in deep reinforcement learning
  for robotics: a survey}. In \bibinfo{booktitle}{\emph{2020 IEEE Symposium
  Series on Computational Intelligence (SSCI)}}. IEEE,
  \bibinfo{pages}{737--744}.
\newblock


\bibitem[\protect\citeauthoryear{Zhu, Wong, Mandlekar, and
  Mart{\'\i}n-Mart{\'\i}n}{Zhu et~al\mbox{.}}{2020}]%
        {zhu2020robosuite}
\bibfield{author}{\bibinfo{person}{Yuke Zhu}, \bibinfo{person}{Josiah Wong},
  \bibinfo{person}{Ajay Mandlekar}, {and} \bibinfo{person}{Roberto
  Mart{\'\i}n-Mart{\'\i}n}.} \bibinfo{year}{2020}\natexlab{}.
\newblock \showarticletitle{robosuite: A modular simulation framework and
  benchmark for robot learning}.
\newblock \bibinfo{journal}{\emph{arXiv preprint arXiv:2009.12293}}
  (\bibinfo{year}{2020}).
\newblock


\bibitem[\protect\citeauthoryear{Ziebart, Maas, Bagnell, Dey,
  et~al\mbox{.}}{Ziebart et~al\mbox{.}}{2008}]%
        {ziebart2008maximum}
\bibfield{author}{\bibinfo{person}{Brian~D Ziebart}, \bibinfo{person}{Andrew~L
  Maas}, \bibinfo{person}{J~Andrew Bagnell}, \bibinfo{person}{Anind~K Dey},
  {et~al\mbox{.}}} \bibinfo{year}{2008}\natexlab{}.
\newblock \showarticletitle{Maximum entropy inverse reinforcement learning.}.
  In \bibinfo{booktitle}{\emph{Aaai}}, Vol.~\bibinfo{volume}{8}. Chicago, IL,
  USA, \bibinfo{pages}{1433--1438}.
\newblock


\end{thebibliography}
\bibliographystyle{ACM-Reference-Format}

\end{document}
\endinput